%% file: main.tex
\documentclass[runningheads]{llncs}
\usepackage{eccv} 
\usepackage{eccvabbrv}
\usepackage{bbding}
\usepackage{booktabs}
\usepackage{graphicx}	
\usepackage{amsmath}	
\usepackage{amssymb}	
\usepackage{microtype}
\usepackage{xcolor, colortbl}
\usepackage{stfloats}
\usepackage{enumitem}
\usepackage{tabularx}
\usepackage{multirow}
\usepackage{wrapfig}
\usepackage{xspace}
\usepackage{url}
\usepackage{comment}
\usepackage{subcaption}
\usepackage[accsupp]{axessibility} 
\usepackage{pifont}
\usepackage{utfsym}
\usepackage{makecell}
\definecolor{Firebrick}{RGB}{178,34,34}
\definecolor{tabfirst}{rgb}{1.0, 0.86, 0.86} 
\definecolor{tabsecond}{rgb}{0.65, 0.92, 0.87}
\definecolor{tabthird}{rgb}{0.79, 0.92, 1.0}

\usepackage{fontawesome5} 
\usepackage{xcolor,colortbl}
\definecolor{GoogleBlue}{HTML}{1A73E8}
\usepackage[colorlinks=true,
            linkcolor=red,   
            citecolor=GoogleBlue,  
            urlcolor=GoogleBlue]{hyperref}
\ifdefined\trackedversion
  \newcommand{\revadd}[1]{{\color{blue}#1}}
  \newcommand{\revdel}[1]{{\color{gray}#1}\enspace}
  \newcommand{\revaddcolor}{\color{blue}}
\else
  \newcommand{\revadd}[1]{#1}
  \newcommand{\revdel}[1]{}
  \newcommand{\revaddcolor}{}
\fi

\begin{document}
\title{One Scene, Two Depths: Probing Geometric Ambiguity in Monocular Foundation Models}
\titlerunning{One Scene, Two Depths}
\author{
Xiaohao Xu\inst{1} \and Feng Xue\inst{1} \and Xiang Li\inst{2}
\and Haowei Li\inst{1} \and Shusheng Yang\inst{3}
\and Tianyi Zhang\inst{2} \and Matthew Johnson-Roberson\inst{2,4}
\and Xiaonan Huang\inst{1}
}

\institute{University of Michigan, Ann Arbor, MI, USA\\ \email{\{xiaohaox, fengxe, haoweili, xiaonanh\}@umich.edu} \and Carnegie Mellon University, Pittsburgh, PA, USA\\ \email{\{xl6, tianyiz4\}@andrew.cmu.edu, }  \and New York University, New York, NY, USA\\ \email{shusheng.yang@nyu.edu} \and Vanderbilt University, Nashville, TN, USA\\ \email{matthew.johnson-roberson@vanderbilt.edu} \\ \quad 
\\
GitHub Repo:  \href{https://github.com/Xiaohao-Xu/Ambiguity-in-Space}{https://github.com/Xiaohao-Xu/Ambiguity-in-Space}}
\authorrunning{Xu et al.}

\maketitle

\begin{abstract}
  \input{sec/0_abstract}
\end{abstract}
\input{sec/1_intro}

\input{sec/2_related_work}
\input{sec/3_method}
\input{sec/4_experiment}

\input{sec/5_conclusion}

\bibliographystyle{splncs04}
\bibliography{main}

\clearpage

\appendix
\input{sec/X_appendix}

\end{document}

%% file: sec/0_abstract.tex
A faithful 3D world representation should account for layered geometry, where a single camera ray may contain multiple visible and geometrically valid surfaces. Monocular depth estimation, however, reduces this structure to one scalar depth per pixel. Transparent scenes make this ambiguity measurable: the same ray can pass through foreground glass and observe the background, turning the supervised target into a convention of annotation, data, and training rather than a scene-intrinsic truth. A learned predictor exposes this convention as its depth-layer preference.
We introduce \texttt{MultiDepth-3k} (\texttt{MD-3k}), a sparse two-layer ordinal benchmark for measuring depth-layer preference and multi-layer spatial relationship accuracy (ML-SRA). On \texttt{MD-3k}, leading depth foundation models exhibit diverse layer preferences under standard RGB input, showing that the same layered geometry can be resolved differently across models. We further find that Laplacian Visual Prompting (LVP), a training-free spectral input transformation, can substantially change the reported layer for certain frozen models. The strongest RGB/LVP pair, DAv2-L, reaches 75.5\% ML-SRA. These results suggest that depth foundation models may express complementary geometric hypotheses that standard RGB inference leaves unexpressed. We invite the community to rethink depth supervision and evaluation through an ambiguity-aware lens, where multiple valid 3D interpretations are treated as geometric structure to be measured, preserved, and expressed.

%% file: sec/1_intro.tex
\section{Introduction}
\label{sec:intro}

\begin{figure}[t]
    \centering
    \setlength{\abovecaptionskip}{0.2cm}
    \includegraphics[width=\textwidth]{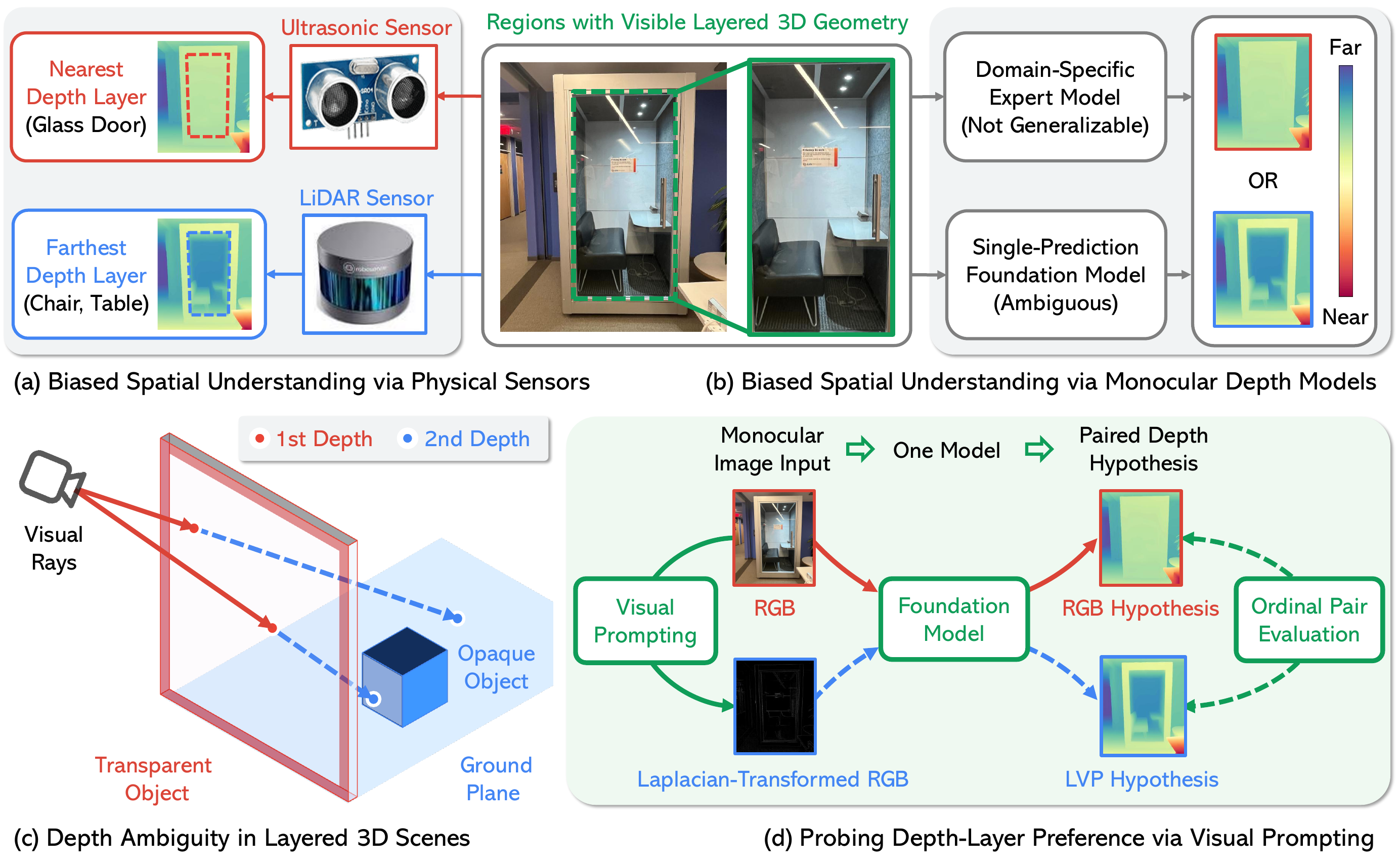}
   \caption{\textbf{Rethinking geometric ambiguity for 3D spatial understanding.}
    \revdel{(\textbf{a}) A transparent scene can support multiple visible layers
    along one ray, but sensors and renderers record one layer according to their
    own conventions.
    (\textbf{b}) Single-layer supervision collapses this geometric ambiguity
    into one recorded target.}
    \revadd{(\textbf{a}) Ambiguous layered scenes can contain multiple visible
    surfaces along one ray, while single-depth supervision records only one
    biased layer.
    (\textbf{b}) This collapse turns multi-layer geometry into a dataset-shaped
    scalar target.}
    (\textbf{c}) A single line of sight intersects multiple surfaces in
    transparent scenes.
    (\textbf{d}) Laplacian Visual Prompting (LVP) \revdel{modulates a receptive frozen model's depth-layer preference and can yield a complementary ordinal hypothesis without retraining; the response is not guaranteed per image and remains a benchmark-level behavior}\revadd{can surprisingly modulate the predicted layer of certain frozen models without retraining, revealing complementary ordinal behavior at the benchmark level}.}
    \label{fig:teaser} 
\end{figure}

Depth maps are a compact interface between images and 3D reasoning: each pixel is assigned a distance and can be lifted into a point for reconstruction, navigation, and scene understanding. This interface quietly assumes that each visual ray has one surface to report. The assumption is usually acceptable in opaque scenes, but it becomes fragile when visibility is layered. Modern depth foundation models~\cite{midas,depth_anything,yang2024depth} inherit this interface: despite broad pretraining, they predominantly follow the \textbf{single-depth prediction} paradigm, mapping an image to one depth value per pixel. Transparent scenes expose the tension clearly: one ray can carry evidence from both a transparent foreground and the scene behind it (Fig.~\ref{fig:teaser}c). 
Collapsing this multi-layer geometry into one target turns the depth label into a \emph{layer convention} shaped by annotation, dataset construction, and single-depth training, rather than a unique property of the image or scene.


\begin{figure}[t]
\centering
\setlength{\abovecaptionskip}{0.2cm}
\includegraphics[width=\textwidth]{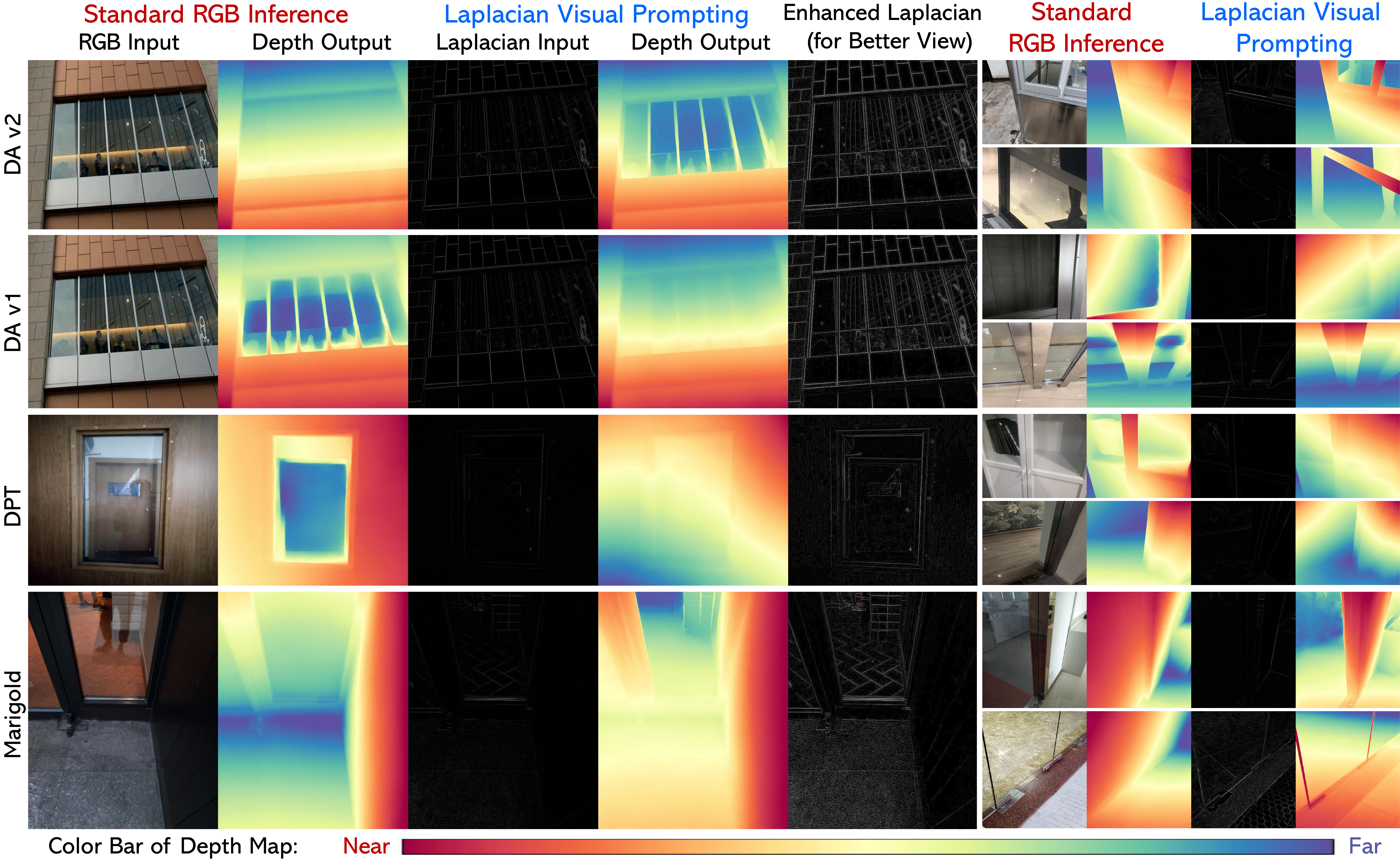}
\caption{\revdel{\textbf{Model-dependent depth-layer modulation.} Standard RGB input yields each model's default depth layer prediction convention (\textbf{Cols.~2 and 7}); LVP can change the expressed preference of receptive frozen models and produce a candidate complementary ordinal hypothesis in ambiguous regions (\textbf{Cols.~4 and 9}). Weak, absent, or undesired shifts are informative boundary conditions for the probe rather than proof of universal switching~\cite{depth_anything,yang2024depth,dpt,marigold}.}\revadd{\textbf{Model-dependent depth-layer modulation.} Standard RGB input reveals each model's default depth layer preference (\textbf{Cols.~2 and 7}). Laplacian Visual Prompting can change the reported layer for receptive frozen models~\cite{depth_anything,yang2024depth,dpt,marigold} and produce a candidate complementary depth hypothesis in ambiguous regions (\textbf{Cols.~4 and 9}).}}
\label{fig:hidden_depth_1}\vspace{0mm}
\end{figure}

\revdel{Once geometric ambiguity is forced into the single-depth paradigm, the layer convention is formed through mixed supervision bias. Sensor-derived labels inherit hardware behavior: ultrasound may return the proximal glass, while LiDAR may pass through, miss, or emphasize the distal scene (Fig.~\ref{fig:teaser}a). Synthetic labels inherit renderer conventions for ray termination and alpha compositing, as in Hypersim~\cite{hypersim}. Thus, a model trained on this mixture does not learn a tie-free geometry; it learns a \textbf{depth-layer preference}, namely the layer it tends to report when one ray is forced into one scalar target by supervision.

The same convention also appears in evaluation. A standard single-layer metric asks whether a prediction matches the recorded surface, and can penalize another visible, physically valid layer. Therefore, under geometric ambiguity, the sharper question is not simply \emph{which prediction matches the label}, but \textbf{which layer the model chooses, whether the benchmark can measure that choice, and whether the choice can be probed without retraining a frozen model}.

This motivates a benchmark that measures layer choice without assuming dense multi-layer metric ground truth. We introduce \textbf{MultiDepth-3k} (\texttt{MD-3k}), which uses sparse ordinal relations for the transparent foreground and the visible background. Instead of asking for an ill-posed dense target, \texttt{MD-3k} asks a direct behavioral question: \textbf{does an output satisfy the foreground spatial relation or the background spatial relation?}

With this measurement in place, we ask whether a frozen model can express a different answer when queried through a different input representation. We propose Laplacian Visual Prompting (LVP) as this probe. LVP is a deterministic high-frequency transform that can change the expressed preference of receptive models, without changing their weights or guaranteeing a desired layer per image (Fig.~\ref{fig:teaser}d). Figure~\ref{fig:hidden_depth_1} qualitatively illustrates this behavior: RGB exposes the model's default layer convention, whereas LVP can yield a candidate complementary ordinal hypothesis in receptive frozen models.}\revadd{This convention is further shaped by the mixture of supervision. Sensor-derived labels may emphasize different physical layers: ultrasound can return the proximal glass, while LiDAR may pass through, miss, or emphasize the distal scene (Fig.~\ref{fig:teaser}a). Synthetic labels also encode renderer choices for ray termination and alpha compositing, as in Hypersim~\cite{hypersim}. When such sources are mixed in single-depth training, the resulting supervision collapses layered geometry into a dataset-shaped scalar target: as illustrated in Fig.~\ref{fig:teaser}b, either a domain-specific transparent-depth pipeline or a single-prediction foundation model must report one layer, even though the scene admits multiple valid depths. Thus, a model does not learn layer-free geometry; it learns a \textbf{depth-layer preference}, the layer it tends to report when one ray is forced into one target.

The same convention affects evaluation. A standard single-layer metric rewards agreement with the recorded surface and can penalize another visible, physically valid layer. Thus, under layered ambiguity, the key question is not only \emph{which prediction matches the label}, but which valid layer a model reports and how dataset bias shapes that behavior. To this end, we introduce \texttt{MultiDepth-3k} (\texttt{MD-3k}), a sparse ordinal benchmark for measuring depth-layer preference and multi-layer spatial relationship accuracy without dense metric ground truth. \texttt{MD-3k} annotates the transparent foreground and visible background with paired ordinal spatial relations, allowing us to evaluate \emph{{whether a biased single-layer depth output satisfies the foreground or the background spatial relation}}.

Once the default model-specific depth-layer preference under RGB image inputs is measurable, we ask a second question: {\emph{{can a frozen model's predicted layer be modulated by changing only the input representation?} }}This leads to {Laplacian Visual Prompting} (LVP), a simple and deterministic high-frequency input-space transform. Surprisingly, LVP can strongly change the predicted layer for certain frozen models, revealing candidate complementary ordinal behavior without retraining (Fig.~\ref{fig:teaser}d). Figure~\ref{fig:hidden_depth_1} qualitatively illustrates this modulation.}

\revdel{Using \texttt{MD-3k}, we study two questions. First, under standard RGB input, which layer does each foundation model prefer when a transparent scene admits more than one valid depth? Second, when the same frozen model is queried with LVP, does the RGB/LVP pair cover complementary ordinal relations beyond what any duplicated single output can satisfy? We find that current leading depth foundation models exhibit diverse layer preferences under ambiguity. LVP further exposes model-dependent changes in depth prediction: it produces strong depth-layer modulation for DPT~\cite{dpt}, Depth Pro~\cite{bochkovskii2024depth}, and several DAv2 models~\cite{yang2024depth}, while leaving other models closer to their RGB convention. Among these, DAv2-L reaches 75.5\% Multi-Layer Spatial Relationship Accuracy (ML-SRA) on \texttt{MD-3k}, above the strict 56.4\% duplicated single-hypothesis ceiling. This shows that a frozen model's reported geometry can depend on the convention used to query and evaluate it, without implying a discrete hidden layer or per-image control.}\revadd{Using \texttt{MD-3k}, we first find that leading depth foundation models exhibit diverse layer preferences under standard RGB input. Beyond this default behavior, LVP reveals a surprising input-dependent modulation: it strongly changes the predicted layer for DPT~\cite{dpt}, Depth Pro~\cite{bochkovskii2024depth}, and several DAv2 models~\cite{yang2024depth}, while other models remain closer to their RGB preference. The strongest case is DAv2-L, whose RGB/LVP pair reaches 75.5\% Multi-Layer Spatial Relationship Accuracy (ML-SRA) on \texttt{MD-3k}, above the strict 56.4\% duplicated single-hypothesis ceiling. These observations make the central message concrete: even with fixed weights and a single-output head, a depth foundation model can express different valid ordinal relations when the input representation changes.}

\vspace{1mm}
\noindent\textbf{In summary, this framing leads to three contributions:}
\begin{itemize}

\item We frame transparent-scene depth as a controlled case of \textbf{ambiguous layered 3D representation}, where multiple visible and geometrically valid depths may coexist along a ray, but a single-output depth model must choose one. We characterize this choice as a \textbf{model-intrinsic depth-layer preference}. To the best of our knowledge, we provide the first systematic characterization of this preference across diverse monocular depth foundation models.

\item We introduce \textbf{\texttt{MD-3k}}, a real-world transparent-scene benchmark that makes depth-layer preference measurable. \texttt{MD-3k} provides sparse ordinal labels for two valid ray-wise depth layers, the transparent foreground surface and the visible background, enabling \textbf{per-layer and multi-layer evaluation}.

\item We identify a surprising \textbf{input-dependent modulation} of depth-layer preference. Laplacian Visual Prompting serves as a training-free spectral probe that can expose complementary ordinal behavior in certain frozen models, while remaining model-dependent.

\end{itemize}

%% file: sec/2_related_work.tex
\section{Related Work}
\label{sec:related_work}

\noindent\textbf{Monocular depth estimation.}
Our work builds upon the generalization capabilities of modern {Monocular Depth Estimation (MDE)} foundation models. The field
has evolved from domain-specific architectures trained on datasets such as
KITTI~\cite{kitti} and
NYUv2~\cite{nyud,eigen2014depth,adabins} to general-purpose
systems trained on large-scale mixed data. Current state-of-the-art
models achieve robust zero-shot performance through diverse supervision
strategies, including mixing heterogeneous datasets~\cite{midas,dpt,midasv31,
metric3d}, distilling generative priors from diffusion
models~\cite{sd,marigold,depthfm,geowizard}, or leveraging large-scale
pseudo-labeling~\cite{depth_anything,yang2024depth}. Despite their
robust representations, these models are architecturally constrained to produce a {{single depth value per pixel}}. This design choice requires models to resolve complex scene
geometries into a single map,
often resulting in systematic layer preferences under ambiguity.
Our work investigates whether such layer preferences are influenced by input frequency content, and whether spectral input transformations can modulate which depth hypothesis a frozen model produces.

\vspace{1mm}
\noindent\textbf{Depth estimation for ambiguous scenes.}
Recovering 3D scene geometry in the presence of transparent or specular surfaces
is a persistent challenge. Early approaches formulated this as a
{{completion problem}}, employing specialized layers to
infer missing depth values~\cite{sajjan2020cleargrasp,zhu2021rgbd} or
to regress background depth via transparency-aware
losses~\cite{chen2023tode,fang2022transcg}. While effective in
controlled settings, these methods do not explicitly represent
multiple geometric interpretations within a single model output.
Recently, the field has moved toward {{multi-layer
inference}}. Wen et al.~\cite{wen2025layereddepth} introduced a
multi-layer synthetic dataset and fine-tuned depth
models to predict separate geometric layers.
These approaches occupy a fundamentally different research axis from ours: they ask how accurately a retrained model can predict multiple layers, whereas we ask \revdel{what geometric understanding is already latent in frozen single-output models}\revadd{how a frozen single-output model's expressed depth-layer preference changes under controlled input modulation}. Our goal is diagnostic: we probe frequency-conditioned layer preferences in frozen models, rather than maximizing multi-layer prediction accuracy. \revdel{As such, direct performance comparison with supervised multi-layer methods is orthogonal to our contribution.}

\vspace{1mm}
\noindent\textbf{Visual prompting for model adaptation.}
We leverage Visual Prompting (VP)~\cite{Bahng_2022_NeurIPS,
bai2024sequential} to operate within the frozen model's input space.
Inspired by prompting in NLP~\cite{brown2020language}, VP has been
adapted for vision-language alignment~\cite{singha2023ad,wasim2023vita,
khattak2022maple,wang2024vilt}, domain
adaptation~\cite{chen2021adversarial,neekhara2022cross}, and
adversarial robustness~\cite{chen2022visual}. Most existing VP methods
focus on learnable prompts such as optimized pixel patches or input tokens. A distinct sub-category is {{input-space
prompting}}~\cite{tsai2020transfer}, where the input signal is
transformed to steer model behavior without any parameter optimization. Our work
extends this paradigm to 3D geometry.
We apply non-learnable spectral input-space prompting to analyze and modulate depth-layer preference in foundation models, demonstrating that simple spectral prompting can alter the depth hypothesis produced by a frozen single-output estimator.

%% file: sec/3_method.tex
\section{Methodology}
\label{sec:method}

We define a framework for measuring depth-layer preference and testing whether this preference can be modulated in frozen single-output depth models. The framework has three parts. First, we define the single-output setting and the two-layer ordinal representation used for ambiguous scenes. Second, we formulate depth-layer preference for a single prediction and paired-hypothesis complementarity for two predictions. Third, we apply Laplacian Visual Prompting (LVP) as a training-free input transformation and compare its outputs with standard RGB outputs under the same ordinal evaluation.

\subsection{Preliminaries: From Single-Layer Depth to Layered Geometry}
\label{sec:prelim}

\noindent\textbf{The single-layer constraint.}
We consider a pre-trained monocular depth model
$f_\theta : \mathbb{R}^{H\times W\times 3} \to \mathbb{R}^{H\times W}$,
parameterized by frozen model weights $\theta$, that produces a single depth estimate
$\hat{\mathcal{D}} = f_\theta(\mathcal{I})$ per input image $\mathcal{I}$.
Regardless of architecture, all models evaluated in this work emit a scalar depth value per pixel at inference time. For opaque scenes this representation is usually well posed. In ambiguous layered scenes, however, one ray can contain evidence for multiple visible surfaces. We use transparency as a clean two-layer case and denote the valid interpretations as $\mathcal{D}^{(1)}$ for the transparent foreground and $\mathcal{D}^{(2)}$ for the visible background. A single scalar output cannot express both layers, so it must report one layer convention rather than an intrinsically unique depth.

\vspace{1mm}
\noindent\textbf{Multi-layer geometry via sparse ordering.}
We represent the two-layer scene by an ordered pair
$(\mathcal{D}^{(1)}, \mathcal{D}^{(2)})$ such that
$\forall\,\mathbf{x},\; \mathcal{D}^{(1)}(\mathbf{x}) \le
\mathcal{D}^{(2)}(\mathbf{x})$, where $\mathbf{x}$ denotes a spatial location on the image plane. Reliable dense multi-layer depth is fundamentally difficult to collect in real layered scenes. Physical sensors generally return one physical layer, or fail, penetrate, or reflect in material-dependent ways, rather than capturing the complete stack of visible surfaces along a ray. A dense metric label is therefore not sensor-neutral for transparency. Following DIW~\cite{diw} and \texttt{DA-2K}~\cite{yang2024depth}, we formulate correctness through sparse ordinal constraints.

Let $\mathcal{P} = \{(\mathbf{u}_m, \mathbf{v}_m)\}_{m=1}^M$ be a set of sparse point pairs sampled from ambiguous regions. We sample one pair per image and manually annotate its ordinal relation for both the foreground and background layers. The spatial order of ground truth for layer $k \in \{1, 2\}$ is
$y_m^{(k)} = \operatorname{sign}\bigl(\mathcal{D}^{(k)}(\mathbf{u}_m) -
\mathcal{D}^{(k)}(\mathbf{v}_m)\bigr)$.

A predicted depth map $\hat{\mathcal{D}}$ is considered valid for layer $k$ if its relative depth ordering matches the corresponding ordinal label. For clarity, we write $\hat{\mathcal{D}} \equiv y_m^{(k)}$ as shorthand for
$\operatorname{sign}\!\bigl(\hat{\mathcal{D}}(\mathbf{u}_m) -
\hat{\mathcal{D}}(\mathbf{v}_m)\bigr) = y_m^{(k)}$ in the following paragraphs.

\subsection{Problem Formulation: Layer Preference and Paired Hypotheses}
\label{sec:formulation}

\noindent\textbf{Single-output depth-layer preference.}
The sparse ordinal formulation lets us measure the layer convention expressed by a single-output model. Depth-layer preference asks which valid layer a model tends to report under ambiguity. We define $\alpha(f_\theta)$ as the expected difference in sparse ordinal correctness between the background and foreground layers for a frozen model:
\begin{equation}
\label{eq:bias}
\alpha(f_\theta) = \mathbb{E}_{m \sim \mathcal{P}}\Bigl[
  \mathbb{I}(\hat{\mathcal{D}} \equiv y_m^{(2)}) - \mathbb{I}(\hat{\mathcal{D}} \equiv y_m^{(1)})
\Bigr],
\end{equation}
where $\alpha > 0$ indicates background preference and $\alpha < 0$ indicates foreground preference. The absolute value of $\alpha$ denotes preference strength.

\vspace{1mm}
\noindent\textbf{Paired-hypothesis complementarity.}
A single depth map can report only one layer convention. To evaluate whether two depth outputs provide complementary evidence, we form a candidate depth pair $\mathbf{H} = \{\hat{\mathcal{D}}_A, \hat{\mathcal{D}}_B\}$ and jointly evaluate its two maps against both annotated depth layers. We define success as the existence of a dataset-level permutation $\pi^\star$ that maps hypotheses to layers and maximizes the joint satisfaction of the sparse constraints:
\begin{equation}
    \pi^\star = \underset{\pi \in S_2}{\arg\max}\; \sum_{m=1}^M
    \mathbb{I}\!\left(\bigwedge_{k=1}^{2}
    \hat{\mathcal{D}}_{\pi(k)} \equiv y_m^{(k)}\right).
    \label{eq:objective}
\end{equation}
Benchmark labels calibrate a single dataset-level assignment for each candidate
pair. The assignment is fixed across all images, so paired evaluation measures
candidate-pair complementarity after dataset-level label matching.

\subsection{Scope of Ambiguity}
\label{sec:scope}

\noindent\textbf{Radiometric superposition vs.\ occlusion.}
We focus on \textit{transparency} as a primary mode of ambiguity. Unlike \textbf{occlusion}, where the background signal is physically blocked, transparency results in \textbf{radiometric superposition}: photons from both the foreground surface and the background contribute to the observed pixel intensity. Consequently, visual cues from multiple surfaces coexist in the input signal $\mathcal{I}$ in superimposed form. Our method does not attempt to hallucinate missing geometry. Instead, it probes how frozen models respond when frequency components of this signal are selectively emphasized in the image.

\subsection{\texttt{MultiDepth-3k}: Sparse Ordinal Benchmark for Ambiguity}
\label{sec:benchmark}

\begin{figure}[t]
    \centering
    \setlength{\abovecaptionskip}{0.2cm}
    \includegraphics[width=\textwidth]{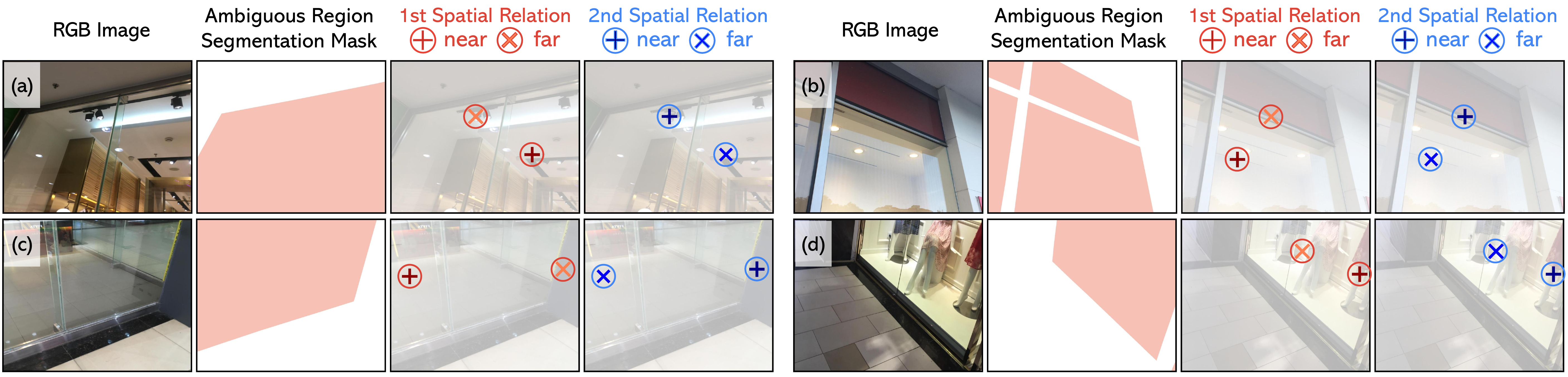}
\caption{\textbf{\texttt{MD-3k} benchmark.} Examples showing ambiguous region masks and sparse point pairs with multi-layer spatial labels. Spatial relationships of sparse point pairs reverse across layers in (\textbf{a})--(\textbf{c}) and remain consistent in (\textbf{d}).}
    \label{fig:benchmark}
\end{figure}

Building on the sparse ordinal formulation, \texttt{MultiDepth-3k} (\texttt{MD-3k}) provides a real-world benchmark for measuring layer choice in transparent scenes.
It consists of 3,161 RGB images sourced from the GDD dataset~\cite{mei2020don},
with one annotated point pair per image. Because physical sensing cannot reliably recover a dense, sensor-neutral stack of visible layers, we use sparse ordinal relations rather than dense metric multi-layer depth, following DIW~\cite{diw} and \texttt{DA-2K}~\cite{yang2024depth}, as shown in Fig.~\ref{fig:benchmark}. Each pair has labels for both layers. Masks and labels were cross-checked by multiple annotators in multiple review rounds before evaluation.

\begin{wrapfigure}{r}{0.5\textwidth}
    \centering\vspace{-22pt}
    \setlength{\abovecaptionskip}{0.2cm}
    \includegraphics[width=0.5\textwidth]{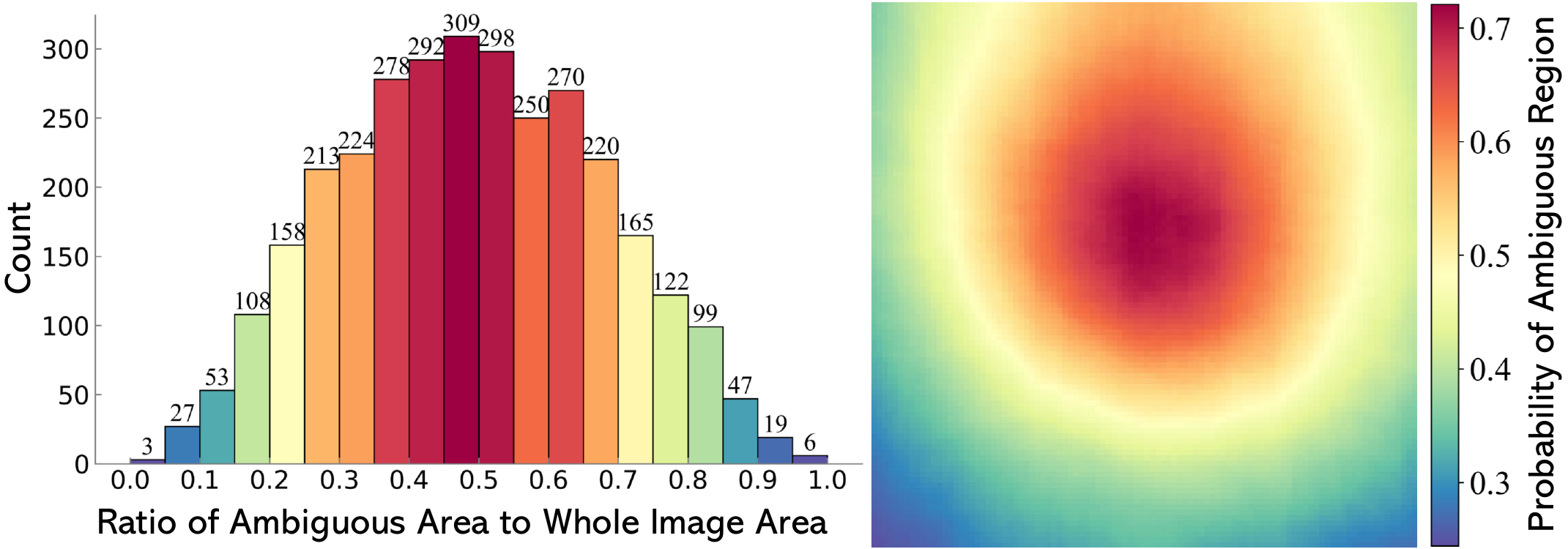}
    \caption{\textbf{Statistics.}
    \textbf{Left}: distribution of ambiguous area ratio per image.
    \textbf{Right}: 2D spatial heatmap of ambiguous regions over benchmark images in \texttt{MD-3k}, shown in normalized image coordinates.}
    \label{fig:benchmark-distribution}\vspace{-20pt}
\end{wrapfigure}

\vspace{1mm}
\noindent\textbf{Benchmark statistics.}
As shown in Fig.~\ref{fig:benchmark-distribution}, the dataset captures diverse ambiguity ratios. We partition the dataset into two subsets. In the \textbf{\textit{Same}} subset (1,783 pairs), relative depth ordering is consistent across layers. In the \textbf{\textit{Reverse}} subset (1,378 pairs), the two layers impose conflicting orders. Consequently, a single-output model cannot satisfy both layers on the \textit{Reverse} subset by duplicating one map, which is exactly the case where multiple hypotheses become necessary. Because all images are drawn from GDD~\cite{mei2020don}, broader validation across capture domains, transparent materials, and object categories remains necessary before treating the benchmark as general transparent-scene coverage rather than a focused diagnostic for measuring layer choice under transparency.

\subsection{Evaluation Metrics}
\label{sec:metrics}

Given these two-layer ordinal labels, we report three quantities. Together, they separate single-output layer preference from paired-output complementarity.

\vspace{1mm}
\noindent\textbf{Spatial Relationship Accuracy (SRA).}
For a given layer $i \in \{1, 2\}$, $\mathrm{SRA}(i)$ measures the fraction of point pairs for which the predicted ordering is consistent with the ground truth:
\label{eq:sra}
$\mathrm{SRA}(i) = \frac{1}{|\mathcal{P}|} \sum_{m=1}^M \mathbb{I}\Bigl(\hat{\mathcal{D}} \equiv y_m^{(i)}\Bigr)$.

\vspace{1mm}
\noindent\textbf{Depth-Layer Preference ($\alpha$).}
We diagnose the direction of a model's bias using the empirical estimator of Eq.~\eqref{eq:bias}:
\label{eq:dlp}
$\alpha(f_\theta) = \mathrm{SRA}(2) - \mathrm{SRA}(1)$.
A positive $\alpha$ indicates background preference. A negative $\alpha$ indicates foreground preference.


\noindent\textbf{Multi-Layer Spatial Relationship Accuracy (ML-SRA).}
ML-SRA measures the fraction of point pairs for which the relative depth ordering
is correctly predicted for both depth layers:
$
\mathrm{ML\mbox{-}SRA}
=
\frac{1}{|\mathcal{P}|}
\sum_{m=1}^{M}
\mathbb{I}
\left(
\bigwedge_{k=1}^{2}
\hat{\mathcal{D}}_{\pi^\star(k)} \equiv y_m^{(k)}
\right)$.
The benchmark-level assignment $\pi^\star$ from Eq.~\eqref{eq:objective} is
fixed for every image. In our RGB/LVP evaluation (Sec.~\ref{ssec:recovery_results}), we use a deterministic
calibration rule: the RGB output is assigned to the layer for which its
benchmark-level SRA is higher, and the LVP output is assigned to the
complementary layer. 


\subsection{Laplacian Visual Prompting: Querying the Expressed Convention}
\label{sec:lvp_method}

\begin{wrapfigure}{r}{0.48\textwidth}
    \centering
    \vspace{-20pt}
    \ifdefined\trackedversion
      \includegraphics[width=0.46\textwidth]{figs/lap_method4_tracked.pdf}
    \else
      \includegraphics[width=0.48\textwidth]{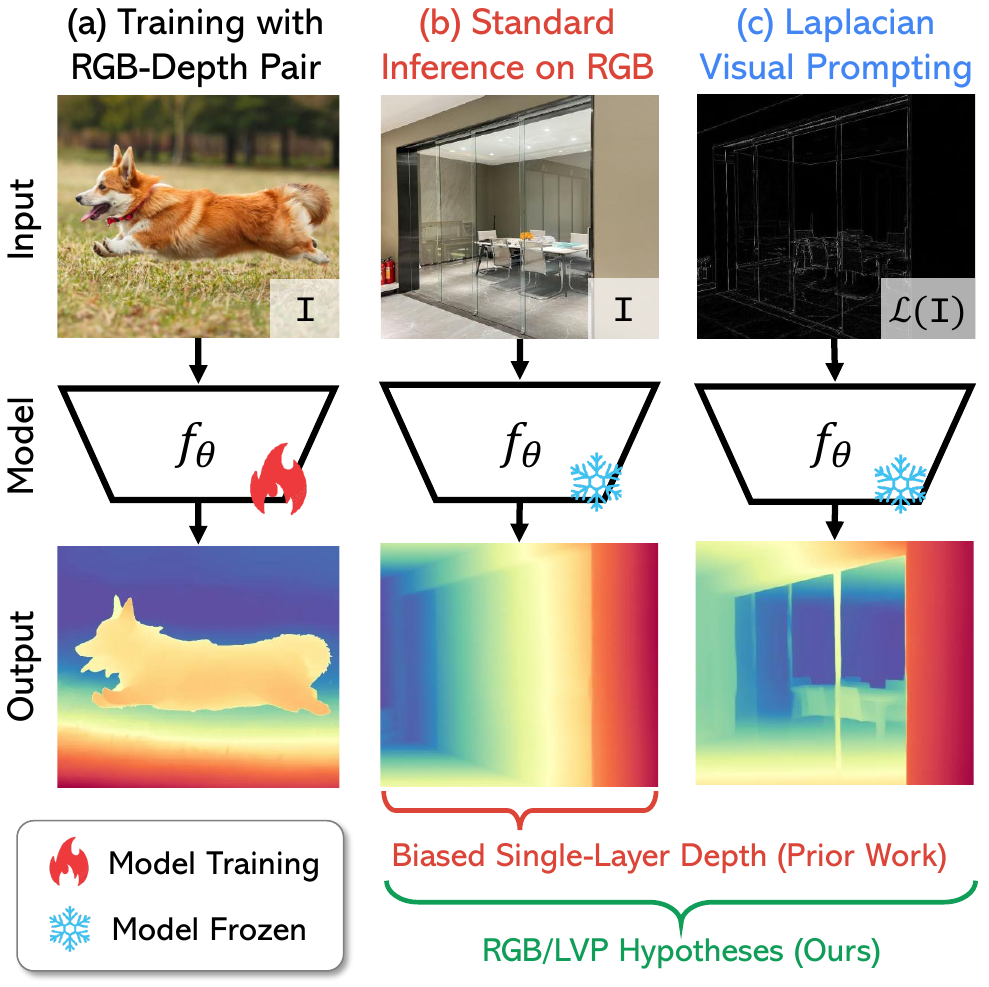}
    \fi\vspace{-5mm}
    \caption{\textbf{The Laplacian Visual Prompting (LVP) method.}
    (\textbf{a}) Standard model training couples RGB to single-layer depth.
    (\textbf{b}) At inference, the standard RGB input yields a single depth estimate, which is biased for ambiguous scenes.
    (\textbf{c}) LVP transforms the input via per-channel floating-point convolution with the Laplacian kernel, followed by min--max mapping back to the image-input value range, producing a candidate alternative output hypothesis from the same frozen depth model.}
    \label{fig:method}
   \vspace{-40pt}
\end{wrapfigure}

\vspace{2mm}
After defining how layer choice is measured, we use LVP as the complementary query: can the same frozen model express a different convention under a different input representation? As illustrated in Fig.~\ref{fig:method}a--b, standard depth foundation model training couples RGB images to single-layer supervision, so RGB inference returns one preferred/biased depth layer under ambiguity. LVP constructs a Laplacian-transformed RGB image from the same image and uses it as an alternative input representation. As shown in Fig.~\ref{fig:method}c, this can alter the expressed depth-layer preference in architecturally sensitive models without modifying frozen model weights $\theta$.

\vspace{1mm}
\noindent\textbf{Spectral reweighting.}
We derive the discrete operator from the continuous Laplacian $\Delta \mathcal{I} = \nabla^2 \mathcal{I}$.
Using a second-order finite difference approximation, we obtain the discrete Laplacian kernel $\mathcal{K}_{\mathcal{L}}$:
{\small
\begin{equation}
    \mathcal{K}_{\mathcal{L}} =
    \begin{bmatrix} 0 \quad& 1 \quad& 0 \\ 1 \quad & -4 \quad& 1 \\ 0 \quad& 1 \quad& 0
    \end{bmatrix}.
    \label{eq:kernel}
\end{equation}
}
We apply this convolution channel-wise in floating point to the input RGB image $\mathcal{I}$, yielding signed residuals
$\mathcal{R}_{\mathrm{raw},c} = \mathcal{I}_c * \mathcal{K}_{\mathcal{L}}$ for each color channel $c$.

\vspace{2mm}
\noindent\textbf{Normalization and inference.}
The raw Laplacian response is a signed floating-point residual and is not directly a valid image input. We therefore map it back to the value range of the model's image-input representation before applying the model-specific preprocessing pipeline. Let this image-input range be $[a,b]$. For example, $[a,b]=[0,1]$ for floating-point RGB images, or equivalently $[a,b]=[0,255]$ before conversion by an image processor. We define
{\small
\begin{equation}
    \mathcal{L}(\mathcal{I})_c(\mathbf{x}) =
    a + (b-a)
    \frac{
    \mathcal{R}_{\mathrm{raw},c}(\mathbf{x}) -
    \min_{\mathbf{x}'} \mathcal{R}_{\mathrm{raw},c}(\mathbf{x}')
    }
    {
    \max_{\mathbf{x}'} \mathcal{R}_{\mathrm{raw},c}(\mathbf{x}') -
    \min_{\mathbf{x}'} \mathcal{R}_{\mathrm{raw},c}(\mathbf{x}') + \epsilon
    },
 \label{eq:norm}
\end{equation}
}
where the minimum and maximum are computed spatially for each channel $c$, and $\epsilon$ is a small constant (e.g., $\epsilon = 10^{-8}$) for numerical stability.
This step is an image-space value-range mapping. It does not replace or modify the model-specific resizing, rescaling, or normalization used for standard RGB inference. In our implementation, the image processor expects a standard 8-bit RGB image, so we map the LVP residual image back to the \texttt{uint8} value range $[0,255]$ and pass the resulting image through the same processor used for the original RGB input.


Given a frozen model $f_\theta$, we obtain the LVP-conditioned depth hypothesis:
\begin{equation}
    \mathcal{D}_{\mathrm{LVP}} = f_\theta\!\left(\mathcal{L}(\mathcal{I})\right).
    \label{eq:lvp_depth}
\end{equation}
For ML-SRA, $(\mathcal{D}_{\mathrm{RGB}}, \mathcal{D}_{\mathrm{LVP}})$ is treated as an unordered candidate pair. Benchmark labels select one global permutation $\pi^\star$ for each model's output pair, and that assignment is fixed for all images. No per-instance oracle is used. Thus, ML-SRA measures pair complementarity after dataset-level label matching, not automatic layer control. Deployment would require labeled calibration or an external semantic or uncertainty-based selector.

%% file: sec/4_experiment.tex
\section{Experiments}
\label{sec:experiments}

\revdel{\noindent\textbf{Research question.} The experiments follow the same logic as the method: ambiguity first, measurement second, and probing third. We first ask which layer each frozen model selects under the standard RGB input. We then test whether an RGB/LVP candidate pair jointly satisfies the two layer-specific ordinal relations in \texttt{MD-3k}, beyond what a duplicated single output can satisfy. Finally, we examine when this behavior appears, when it fails, and what it suggests for qualitative downstream use. Accordingly, we report (i)~RGB depth-layer preference, (ii)~ML-SRA for RGB/LVP candidate pairs, (iii)~scale and training-distribution effects, (iv)~prompt-design boundary conditions, and (v)~qualitative uses of paired hypotheses.}\revadd{\noindent\textbf{Research question.} We first use \texttt{MD-3k} to measure the default depth-layer preference of leading models under standard RGB image input. We then ask the question: can a training-free input transformation modulate that preference in a frozen model? We evaluate this through RGB/LVP candidate pairs, report ML-SRA against the two layer-specific ordinal relations, and analyze when the modulation appears or fails. Accordingly, we report (i)~depth-layer preference, (ii)~ML-SRA for RGB/LVP candidate pairs, (iii)~scale and training-distribution effects, (iv)~prompt-design ablation studies, and (v)~uses of paired hypotheses.}

\vspace{2mm}
\noindent\textbf{Experimental setup.}
We evaluate a diverse suite of pre-trained monocular depth models in a
strictly training-free manner. The benchmarked models include the Depth Anything series (DAv1/v2-\{S,B,L\}) along with the domain-specialized Indoor/Outdoor variants of DAv2 (DAv2-I/O-\{S,B,L\})~\cite{depth_anything,yang2024depth}. We also
evaluate discriminative architectures
(DPT~\cite{dpt}, ZoeDepth~\cite{zoedepth}), generative models
(Marigold~\cite{marigold}, GeoWizard~\cite{geowizard}), and metric
estimators (Depth Pro~\cite{bochkovskii2024depth}, UniK3D~\cite{piccinelli2025unik3d}, UniDepth-v2~\cite{piccinelli2025unidepthv2}).
Our evaluations are mainly performed on \texttt{MD-3k}.

\subsection{Depth-Layer Preference Analysis}
\label{ssec:bias_diagnosis}

\begin{table*}[t]
    \centering
    \setlength{\belowcaptionskip}{0.2cm}
       \caption{\textbf{Per-layer Spatial Relationship Accuracy (SRA) [\%] on \texttt{MD-3k} and reference SRA on \texttt{DA-2K}.} On \texttt{MD-3k}, \textcolor[HTML]{DB4437}{SRA(1)} and \textcolor[HTML]{4285F4}{SRA(2)} measure agreement between a model's single-layer depth prediction and the transparent foreground layer or visible background layer, respectively. The \textit{Reverse} subset contains conflicting foreground/background ordinal relations, while the \textit{Same} subset contains consistent relations and is therefore reported as SRA(1/2). \texttt{DA-2K} is included as a non-ambiguous reference. Bold entries mark the better RGB or LVP input for the corresponding foreground/background SRA comparison on the \textit{Overall} and \textit{Reverse} subsets.} \label{tab:per_layer_sra_main}
    \resizebox{\textwidth}{!}{
        \centering
        \renewcommand{\arraystretch}{1.30}
        \setlength\tabcolsep{1mm}
        
        \begin{tabular}{l|cccc|cccc|cc|cc}
            \hline \hline
            \multirow{3}{*}{\textbf{Method}} & \multicolumn{4}{c|}{\textbf{(a) \texttt{MD-3k} (\textit{Overall})}} & \multicolumn{4}{c|}{\textbf{(b) \texttt{MD-3k} (\textit{Reverse})}} & \multicolumn{2}{c|}{\textbf{(c) \texttt{MD-3k} (\textit{Same})}} & \multicolumn{2}{c}{\textbf{(d)} \texttt{\textbf{DA-2K}}} \\ \cline{2-13}
             & \multicolumn{2}{c}{RGB Input} & \multicolumn{2}{c|}{LVP Input} & \multicolumn{2}{c}{RGB Input} & \multicolumn{2}{c|}{LVP Input} & RGB & LVP & RGB & LVP \\ \cline{2-13}
             &  \textcolor[HTML]{DB4437}{SRA(1)} & \textcolor[HTML]{4285F4}{SRA(2)} &  \textcolor[HTML]{DB4437}{SRA(1)} & \textcolor[HTML]{4285F4}{SRA(2)} &  \textcolor[HTML]{DB4437}{SRA(1)} & \textcolor[HTML]{4285F4}{SRA(2)} &  \textcolor[HTML]{DB4437}{SRA(1)} & \textcolor[HTML]{4285F4}{SRA(2)} & SRA(1/2) & SRA(1/2) & SRA & SRA \\
            \hline
           \rowcolor[HTML]{F1F3F4}  Random & 50.0 & 50.0 & 50.0 & 50.0 & 50.0 & 50.0 & 50.0 & 50.0 & 50.0 & 50.0 & 50.0 & 50.0 \\ \hline
            Marigold & 70.7 & \textcolor[HTML]{4285F4}{\textbf{79.9}} & \textcolor[HTML]{DB4437}{\textbf{70.8}}  & 77.4 & 39.6 & \textcolor[HTML]{4285F4}{\textbf{60.4}} & \textcolor[HTML]{DB4437}{\textbf{42.3}}  & 57.7 & 95.0 & 92.6 & 88.9 & 78.9 \\
         \rowcolor[HTML]{F1F3F4}   GeoWizard & 68.3 & \textcolor[HTML]{4285F4}{\textbf{83.8}} & \textcolor[HTML]{DB4437}{\textbf{70.3}}  & 78.7 & 32.3 & \textcolor[HTML]{4285F4}{\textbf{67.7}} & \textcolor[HTML]{DB4437}{\textbf{40.3}}  & 59.7 & 96.2 & 93.4 & 90.3 & 85.0 \\
            ZoeDepth & 58.7 & \textcolor[HTML]{4285F4}{\textbf{92.6}} & \textcolor[HTML]{DB4437}{\textbf{74.6}}  & 70.9 & 12.0 & \textcolor[HTML]{4285F4}{\textbf{88.0}} & \textcolor[HTML]{DB4437}{\textbf{54.3}}  & 45.7 & 94.9 & 90.4 & 86.7 & 77.2 \\
        \rowcolor[HTML]{F1F3F4}    DPT & 58.0 & \textcolor[HTML]{4285F4}{\textbf{91.9}} & \textcolor[HTML]{DB4437}{\textbf{75.7}}  & 71.1 & 10.2 & \textcolor[HTML]{4285F4}{\textbf{89.8}} & \textcolor[HTML]{DB4437}{\textbf{55.2}}  & 44.8 & 94.8 & 91.5 & 83.2 & 72.2 \\ \hline 
        \textcolor{black}{Depth Pro} &\textcolor[HTML]{DB4437}{\textbf{84.9}}  &69.6   & 73.5   & \textcolor[HTML]{4285F4}{\textbf{76.8}}  &\textcolor[HTML]{DB4437}{\textbf{67.6}}   &32.4  &46.2   &\textcolor[HTML]{4285F4}{\textbf{53.8}}   &98.3   &94.6  &95.8   &84.1   \\
     \rowcolor[HTML]{F1F3F4} \textcolor{black}{UniDepth-v2-L}&  \textcolor[HTML]{DB4437}{\textbf{66.8}}  &\textcolor[HTML]{4285F4}{\textbf{86.6}}   &65.5    & 85.4  &\textcolor[HTML]{DB4437}{\textbf{27.3}}   &72.7  &27.1   &\textcolor[HTML]{4285F4}{\textbf{72.9}}   &97.4   &95.1  &95.4   &92.2   \\
        \textcolor{black}{UniK3D-L} &64.3  &\textcolor[HTML]{4285F4}{\textbf{88.9}}   &\textcolor[HTML]{DB4437}{\textbf{67.1}}    &84.3   &21.8    &\textcolor[HTML]{4285F4}{\textbf{78.2}}   &\textcolor[HTML]{DB4437}{\textbf{30.3}}   &69.7   &97.2  &95.6   &93.1  &85.8  \\

        \hline
            DAv1-S & 56.8 & \textcolor[HTML]{4285F4}{\textbf{95.3}} & \textcolor[HTML]{DB4437}{\textbf{60.6}}  & 85.4 & 5.8 & \textcolor[HTML]{4285F4}{\textbf{94.2}} & \textcolor[HTML]{DB4437}{\textbf{21.5}}  & 78.5 & 96.1 & 90.7 & 88.7 & 83.0 \\
       \rowcolor[HTML]{F1F3F4}     DAv1-B & 56.8 & \textcolor[HTML]{4285F4}{\textbf{95.4}} & \textcolor[HTML]{DB4437}{\textbf{58.8}}  & 89.7 & 5.8 & \textcolor[HTML]{4285F4}{\textbf{94.2}} & \textcolor[HTML]{DB4437}{\textbf{14.6}}  & 85.4 & 96.3 & 93.0 & 89.9 & 86.2 \\
            DAv1-L & 56.6 & \textcolor[HTML]{4285F4}{\textbf{96.0}} & \textcolor[HTML]{DB4437}{\textbf{59.2}}  & 90.9 & 4.8 & \textcolor[HTML]{4285F4}{\textbf{95.2}} & \textcolor[HTML]{DB4437}{\textbf{13.6}}  & 86.4 & 96.6 & 94.3 & 89.5 & 88.5 \\ \hline
            DAv2-O-S & 71.6 & \textcolor[HTML]{4285F4}{\textbf{77.3}} & \textcolor[HTML]{DB4437}{\textbf{81.4}}  & 60.1 & 43.4 & \textcolor[HTML]{4285F4}{\textbf{56.6}} & \textcolor[HTML]{DB4437}{\textbf{74.4}}  & 25.6 & 93.3 & 86.8 & 82.3 & 60.9 \\
        \rowcolor[HTML]{F1F3F4}    DAv2-O-B & 70.3 & \textcolor[HTML]{4285F4}{\textbf{80.3}} & \textcolor[HTML]{DB4437}{\textbf{77.2}}  & 65.8 & 38.5 & \textcolor[HTML]{4285F4}{\textbf{61.5}} & \textcolor[HTML]{DB4437}{\textbf{63.1}}  & 36.9 & 94.8 & 88.2 & 89.7 & 70.4 \\
            DAv2-O-L & 70.4 & \textcolor[HTML]{4285F4}{\textbf{82.4}} & \textcolor[HTML]{DB4437}{\textbf{71.5}}  & 79.5 & 36.2 & \textcolor[HTML]{4285F4}{\textbf{63.8}} & \textcolor[HTML]{DB4437}{\textbf{40.8}}  & 59.2 & 96.7 & 95.2 & 93.7 & 83.8 \\ \hline
            DAv2-I-S & \textcolor[HTML]{DB4437}{\textbf{80.4}} & 71.2 & {{70.0}} & \textcolor[HTML]{4285F4}{\textbf{74.3}}  & \textcolor[HTML]{DB4437}{\textbf{60.6}} & 39.4 & {{45.1}} & \textcolor[HTML]{4285F4}{\textbf{54.9}}  & 95.8 & 89.3 & 88.5 & 79.2 \\
       \rowcolor[HTML]{F1F3F4}     DAv2-I-B & \textcolor[HTML]{DB4437}{\textbf{83.1}} & 69.7 & {{72.6}} & \textcolor[HTML]{4285F4}{\textbf{76.1}}  & \textcolor[HTML]{DB4437}{\textbf{65.3}} & 34.7 & {{46.0}} & \textcolor[HTML]{4285F4}{\textbf{54.0}}  & 96.8 & 93.1 & 91.8 & 81.1 \\
            DAv2-I-L & \textcolor[HTML]{DB4437}{\textbf{85.2}} & 68.1 & {{68.1}} & \textcolor[HTML]{4285F4}{\textbf{82.6}}  & \textcolor[HTML]{DB4437}{\textbf{69.6}} & 30.4 & {{33.5}} & \textcolor[HTML]{4285F4}{\textbf{66.5}}  & 97.3 & 95.0 & 94.8 & 88.4 \\ \hline
            DAv2-S & \textcolor[HTML]{DB4437}{\textbf{78.0}} & 76.2 & {{61.5}} & \textcolor[HTML]{4285F4}{\textbf{85.8}}  & \textcolor[HTML]{DB4437}{\textbf{52.0}} & 48.0 & {{22.1}} & \textcolor[HTML]{4285F4}{\textbf{77.9}}  & 98.0 & 91.9 & 95.1 & 86.6 \\
      \rowcolor[HTML]{F1F3F4}      DAv2-B & \textcolor[HTML]{DB4437}{\textbf{82.4}} & 72.3 & {{60.5}} & \textcolor[HTML]{4285F4}{\textbf{88.6}}  & \textcolor[HTML]{DB4437}{\textbf{61.7}} & 38.3 & {{17.7}} & \textcolor[HTML]{4285F4}{\textbf{82.3}}  & 98.5 & 93.5 & 96.7 & 89.5 \\
            DAv2-L & \textcolor[HTML]{DB4437}{\textbf{84.0}} & 70.6 & {{60.2}} & \textcolor[HTML]{4285F4}{\textbf{89.9}}  & \textcolor[HTML]{DB4437}{\textbf{65.3}} & 34.7 & {{15.9}} & \textcolor[HTML]{4285F4}{\textbf{84.1}}  & 98.5 & 94.4 & 96.9 & 91.5 \\
            \hline \hline
        \end{tabular}
    }

\end{table*}

\noindent\textbf{Per-layer ordinal behavior.}
Before aggregating RGB/LVP outputs into paired depth hypotheses, we first report the single-output depth layer behavior under RGB and LVP inputs directly. Table~\ref{tab:per_layer_sra_main} presents per-layer Spatial Relationship Accuracy (SRA) with respect to the transparent foreground layer and the visible background layer on \texttt{MD-3k}, together with \texttt{DA-2K} results as a non-ambiguous reference. On the \textit{Same} subset of \texttt{MD-3k}, where the two layers induce consistent ordinal spatial relations, SRA is generally on par with \texttt{DA-2K}, a standard benchmark of mostly non-ambiguous scenes. This confirms that \texttt{MD-3k} does not merely introduce a harder ordinal task; its distinctive challenge lies in the \textit{Reverse} subset, where the two valid layers impose conflicting spatial relations.


\begin{figure}[t]
    \centering
    \setlength{\abovecaptionskip}{0.2cm}
    \includegraphics[width=\linewidth]{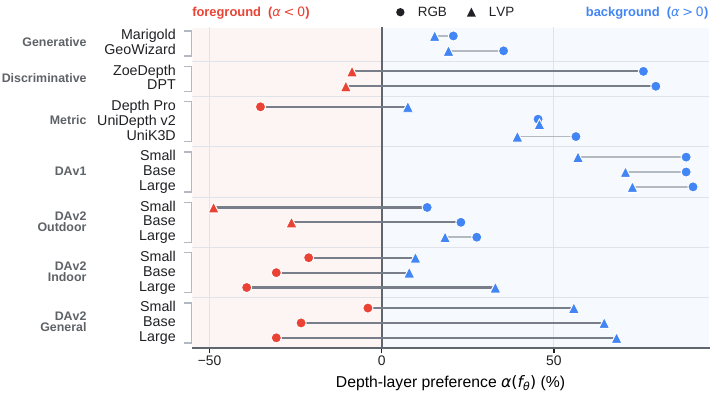}
    \caption{\revdel{\textbf{Model-specific depth-layer preference and LVP response.}
    Each row connects the RGB circle to the LVP diamond on the \textit{Reverse}
    subset of \texttt{MD-3k}. Shape identifies the input, whereas marker fill
    identifies the preferred layer: red denotes the transparent foreground
    ($\alpha<0$) and blue denotes the background ($\alpha>0$). A cross-zero
    segment therefore marks a change in preferred layer. The heterogeneous
    endpoints and shifts show that both baseline preference and LVP response are
    model-dependent. Light red/blue fields reinforce the sign semantics; the
    foreground side is cropped to the observed range to avoid unused plot area.}
    \revadd{\textbf{Model-dependent depth-layer preference.}
    On \texttt{MD-3k} \textit{Reverse}, each row links RGB (circle) and LVP
    (triangle). Fill indicates the preferred layer (red: foreground; blue:
    background), and crossing $\alpha=0$ indicates a layer change. The varied
    endpoints and shifts expose model-specific RGB priors and LVP responses.}}
    \label{fig:depth_layer_bias}
   \vspace{-1mm}
\end{figure}

\vspace{1mm}
\noindent\textbf{Heterogeneity of RGB bias.}
Table~\ref{tab:per_layer_sra_main} reports the raw per-layer ordinal agreement, from which the depth-layer preference is computed as $\alpha=\mathrm{SRA}(2)-\mathrm{SRA}(1)$. The RGB circles in Fig.~\ref{fig:depth_layer_bias} visualize this derived preference on the \textit{Reverse} subset. Depth foundation models exhibit strong but inconsistent layer preferences under standard RGB image input. The
\textit{Depth Anything} family shows this split: general-purpose DAv2
(DAv2-S/B/L) and indoor-tuned DAv2-I variants favor the first layer
(transparent foreground, $\alpha < 0$), whereas outdoor-tuned DAv2-O variants and
DAv1 favor the second layer (background, $\alpha > 0$), similar to generative
models such as Marigold. \revdel{This heterogeneity confirms that ambiguity resolution is driven by training distribution priors (e.g., indoor vs.\ outdoor tuning) rather than a universal geometric rule. This is a key finding that enables practitioners to select models whose bias aligns with their application requirements.}\revadd{This pattern matches the supervision story above: mixed training data and domain bias affect which surface a model reports under transparency. In particular, indoor-tuned variants more often select the proximal surface, whereas outdoor-tuned variants more often select the distal scene.}

\vspace{1mm}
\noindent\textbf{Surprising LVP modulation.}
The same per-layer SRA values in Table~\ref{tab:per_layer_sra_main} also allow us to compute how LVP changes the derived preference $\alpha=\mathrm{SRA}(2)-\mathrm{SRA}(1)$. The LVP triangles and RGB-to-LVP segments in Fig.~\ref{fig:depth_layer_bias} visualize this preference shift on the \textit{Reverse} subset. The effect is strongly model-specific and, for several models, unexpectedly large: general-purpose DAv2, DPT, ZoeDepth, and Depth Pro shift substantially, whereas DAv1 and several generative estimators respond weakly. For example, on the \textit{Reverse} subset, DAv2-L changes from foreground-biased RGB behavior to background-biased LVP behavior: SRA(1) drops from 65.3\% to 15.9\%, while SRA(2) rises from 34.7\% to 84.1\%. This contrast suggests that LVP is not a universal layer switch, but a probe of spectral receptivity: it reveals which frozen backbones can express alternative layer behavior under the same input-level intervention.

\subsection{\revdel{Multi-Layer Recovery Performance}\revadd{Multi-Layer Ordinal Performance}}
\label{ssec:recovery_results}

\begin{figure*}[t]
    \centering
    \begin{minipage}[t]{0.345\textwidth}
        \captionsetup{type=table}
        \caption{\textbf{ML-SRA [\%] on \texttt{MD-3k}.}
        ${\dagger}$ The strict 56.4\% ceiling applies to a duplicated single-map pair under ML-SRA and follows from the benchmark partition ratio.}
        \vspace{1mm}
        \renewcommand{\arraystretch}{1.10}
        \label{tab:ml_sra_table}
        \resizebox{1.0\linewidth}{!}{%
            \begin{tabular}{l|c|cc}
             \hline \hline
            \textbf{Method} & \textbf{\textit{Overall}} &
            \textbf{\textit{Reverse}} & \textbf{\textit{Same}} \\
            \midrule
            \rowcolor[HTML]{E8F6F3}
            \textcolor{black}{\begin{tabular}[c]{@{}l@{}}
                \textit{Ideal Collapsed}\\
                \textit{Baseline}${}^{\dagger}$
            \end{tabular}} &
            \textcolor{black}{56.4${}^{\dagger}$} &
            \textcolor{black}{0.0} &
            \textcolor{black}{100.0${}^{\dagger}$} \\
            \midrule
            Marigold        & 57.4 & 15.3 & 89.8 \\
            \rowcolor[HTML]{E8F6F3}
            GeoWizard       & 59.5 & 17.6 & 91.9 \\
            ZoeDepth        & 68.8 & 45.4 & 86.8 \\
            \rowcolor[HTML]{E8F6F3}
            DPT             & 70.2 & 46.4 & 88.7 \\
            \midrule
            Depth Pro       & 66.3 & 31.1 & 93.5 \\
            \rowcolor[HTML]{E8F6F3}
            UniDepth-v2-L   & 61.3 & 13.6 & 93.7 \\
            UniK3D-L        & 58.9 & 19.2 & \textbf{93.9} \\
            \midrule
            DAv1-S          & 57.9 & 17.7 & 89.0 \\
            \rowcolor[HTML]{E8F6F3}
            DAv1-B          & 56.6 & 11.4 & 91.5 \\
            DAv1-L          & 57.1 & 10.9 & 92.8 \\
            \midrule
            \rowcolor[HTML]{E8F6F3}
            DAv2-O-S        & 63.0 & 36.6 & 83.5 \\
            DAv2-O-B        & 62.7 & 32.9 & 85.6 \\
            \rowcolor[HTML]{E8F6F3}
            DAv2-O-L        & 60.4 & 17.6 & 93.4 \\
            \midrule
            DAv2-I-S        & 60.9 & 27.7 & 86.5 \\
            \rowcolor[HTML]{E8F6F3}
            DAv2-I-B        & 63.7 & 28.1 & 91.2 \\
            DAv2-I-L        & 71.1 & 42.5 & 93.2 \\
            \midrule
            \rowcolor[HTML]{E8F6F3}
            DAv2-S          & 67.2 & 36.9 & 90.7 \\
            DAv2-B          & 73.3 & 48.2 & 92.7 \\
            \rowcolor[HTML]{E8F6F3}
            \textbf{DAv2-L} & \textbf{75.5} & \textbf{52.2} & {93.6} \\
            \hline \hline
            \end{tabular}%
        }
    \end{minipage}
    \hfill
    \begin{minipage}[t]{0.63\textwidth}
        \vspace{0mm}
        \captionsetup{type=table}
        \setlength{\belowcaptionskip}{0.2cm}
        \centering
        \renewcommand{\arraystretch}{1.1}
        \caption{\revdel{\textbf{Comparison with semantic priors.} Evaluated on DAv2-L.}\revadd{\textbf{Contextual comparison with semantic priors.} LVP uses DAv2-L; mask interpolation is built on background-biased DAv1-L to obtain a distal map before estimating transparent regions.} $\ddagger$~Using a transparency mask predictor~\cite{mei2020don}
        with mIoU~0.88 on \texttt{MD-3k}.}
        \resizebox{\textwidth}{!}{%
        \begin{tabular}{l|c|c|cc}
      \hline \hline
        \textbf{Method} & \textbf{Sem.} &
        \textbf{\textit{Overall}} &
        \textbf{\textit{Reverse}} & \textbf{\textit{Same}} \\
        \hline
        \rowcolor[HTML]{E8F6F3}
        \revdel{\textbf{LVP (Ours)}}\revadd{\textbf{LVP (DAv2-L)}} & No & 75.5 & 52.2 & 93.6 \\
        \midrule
        \revdel{+ Predicted Mask${}^{\ddagger}$}\revadd{Mask interpolation (pred., DAv1-L)${}^{\ddagger}$} & Yes & 75.8 & 55.2 & 91.6 \\
        \rowcolor[HTML]{E8F6F3}
        \revdel{+ GT Mask (Ideal Upper Bound)}\revadd{Mask interpolation (GT, DAv1-L; oracle)} & Yes & 82.5 & 69.2 & 92.7 \\
        \hline \hline
        \end{tabular}
        }
        \label{tab:comparison_LVP}\vspace{3mm}
        \captionsetup{type=figure}
        \includegraphics[width=\textwidth]{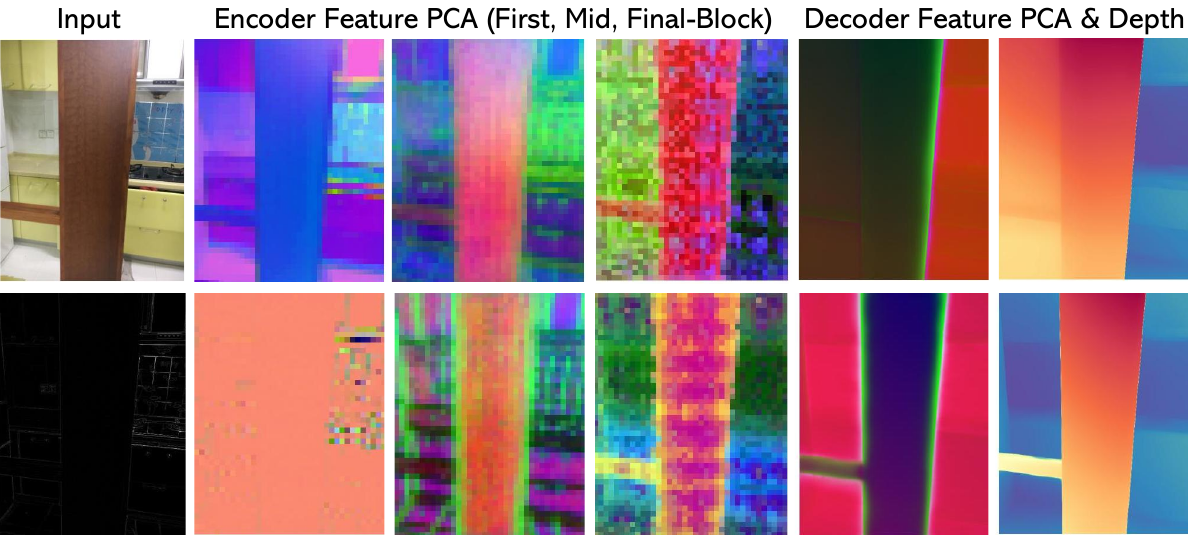}\vspace{-1mm}
        \caption{\textbf{Feature visualization.} PCA of DAv2-L encoder
        and decoder features. \revdel{RGB-input features (Top) lock onto the low-frequency reflection on the proximal glass surface. LVP-input features (Bottom) shift attention to the high-frequency edges of the background structure, consistent with spectral bias modulation.}\revadd{Under LVP input (Bottom), activations place greater emphasis on background high-frequency edges than under RGB input (Top). This qualitative, input-dependent feature highlighting is not evidence of discrete latent depth layers.}}
        \label{fig:embedding_visualization}
    \end{minipage}\vspace{-4mm}
\end{figure*}

\vspace{1mm}
\noindent\textbf{Quantitative analysis.}
\revdel{Table~\ref{tab:ml_sra_table} reports ML-SRA scores of two-layer depth maps produced by our LVP method across all models.}\revadd{Having diagnosed single-output preference, we next evaluate paired-output complementarity. Table~\ref{tab:ml_sra_table} reports ML-SRA for the RGB/LVP candidate pair across all models.}
To contextualize these results, we define an {\textit{Ideal Collapsed
Baseline}}: a hypothetical model that perfectly predicts the primary RGB
depth but na\"{i}vely duplicates it for the secondary layer. This
baseline achieves 100\% on the \textit{Same} subset but \textbf{0\%} on
the \textit{Reverse} subset because satisfying conflicting ordinal
constraints with a single depth map is impossible by construction.
Its weighted overall score of 56.4\% (derived from the benchmark partition: $1783/(1783+1378)$) therefore represents the strict ceiling for any duplicated single-map pair under ML-SRA.
Our method with DAv2-L achieves \textbf{75.5\%} overall and \textbf{52.2\%} on the \textit{Reverse} subset, representing a jump from 0\% to 52.2\% in precisely the cases where a duplicated single output cannot satisfy both relations. \revdel{This \textbf{+19.1\% absolute improvement} over the ideal single-hypothesis ceiling confirms that LVP successfully extracts distinct geometric information inaccessible to standard inference. The fact that LVP produces geometrically structured, physically meaningful outputs above chance on conflicting depth layers is consistent with the hypothesis that these models have learned frequency-conditioned depth priors during training.}\revadd{Therefore, the \textbf{+19.1-point} gain over the ideal single-hypothesis ceiling establishes the central empirical claim: the same frozen model, queried by RGB and LVP inputs, can jointly satisfy contradictory ordinal constraints beyond what any single depth map can achieve. }

\vspace{1mm}
\noindent\textbf{Model-family trend.}
Table~\ref{tab:ml_sra_table} also shows that LVP response is model-dependent, but not explained by a simple discriminative/generative split. DAv2 and DPT exhibit strong modulation, whereas DAv1 and diffusion-based estimators such as Marigold and GeoWizard respond weakly. This suggests that LVP effectiveness depends on the model's spectral receptivity, shaped by both architecture and training regime. We leave a direct causal analysis of this behavior to future work.

\vspace{1mm}
\noindent\textbf{Comparison with semantic priors.}
\revdel{In Table~\ref{tab:comparison_LVP}, we benchmark LVP against a semantics-guided pipeline that integrates a pre-trained segmentation model with the DAv1-L estimator. DAv1-L is known to be strongly background-biased, as characterized in Sec.~\ref{ssec:bias_diagnosis}. The pipeline estimates the nearer layer (transparent surface) by interpolating depth values from the boundaries of the predicted transparent mask. While this hybrid pipeline achieves 75.8\%, it relies on task-specific model combinations, strong planar interpolation assumptions, and an auxiliary segmentation network. In contrast, LVP achieves comparable performance (75.5\%) using \textbf{only the frozen depth model itself}, with no geometric constraints.}\revadd{As a complementary reference, Table~\ref{tab:comparison_LVP} compares against a semantics-assisted pipeline. LVP uses DAv2-L, whereas mask-assisted interpolation uses DAv1-L because its background bias supplies a distal map from which the transparent region is interpolated. The pipelines therefore differ in both backbone and prior. The predicted-mask pipeline reaches 75.8\%, and its GT-mask oracle reaches 82.5\%, but both require semantic localization and planar interpolation. In contrast, LVP reaches 75.5\% without an auxiliary semantic segmentation model on its stated backbone.}

\vspace{1mm}
\noindent\textbf{Latent feature visualization.}
\revdel{Figure~\ref{fig:embedding_visualization} visualizes the principal components of DAv2-L's feature maps under RGB and LVP inputs. Under standard RGB input, the encoder's features lock onto the low-frequency reflection on the glass surface. Under LVP input, these reflection-based activations are suppressed, and attention shifts to the high-frequency edges of the background structure. This is consistent with LVP operating as a spectral steering mechanism. We note that PCA of feature activations is a coarse diagnostic; different-looking inputs are expected to produce different feature representations. Whether this reflects access to a distinct latent depth representation or standard processing of an unusual input remains open.}\revadd{Finally, Figure~\ref{fig:embedding_visualization} provides a qualitative view of input-dependent feature highlighting: under LVP, activations emphasize high-frequency structures that differ from those emphasized by RGB. This feature-level change supports the spectral-modulation interpretation, while the quantitative claim of the paper rests on the ordinal behavior measured by \texttt{MD-3k}.}

\subsection{Ablation and Prompt Analysis}
\label{ssec:ablation}

\begin{figure*}[t]
    \centering
    \begin{minipage}{0.49\linewidth}
        \includegraphics[width=\textwidth]{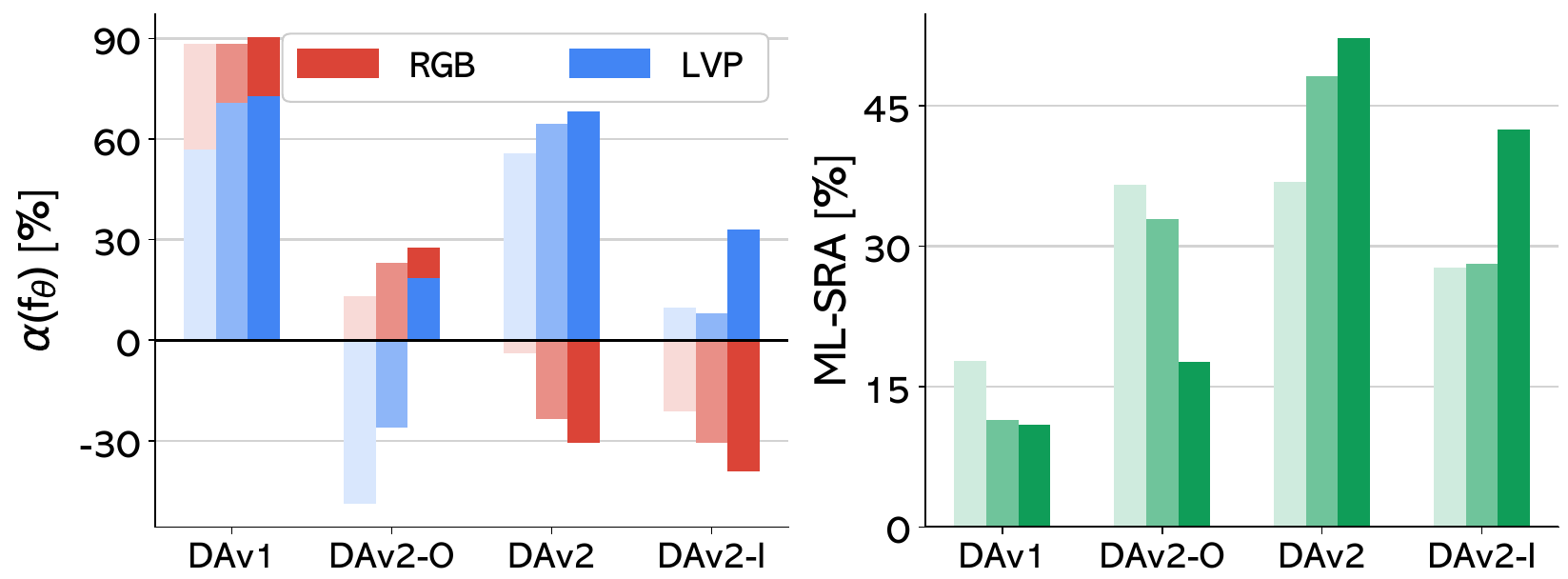}
        \centering
        \footnotesize (a) \textit{Reverse} subset of \texttt{MD-3k}
    \end{minipage}
    \hfill
    \begin{minipage}{0.49\linewidth}
        \includegraphics[width=\textwidth]{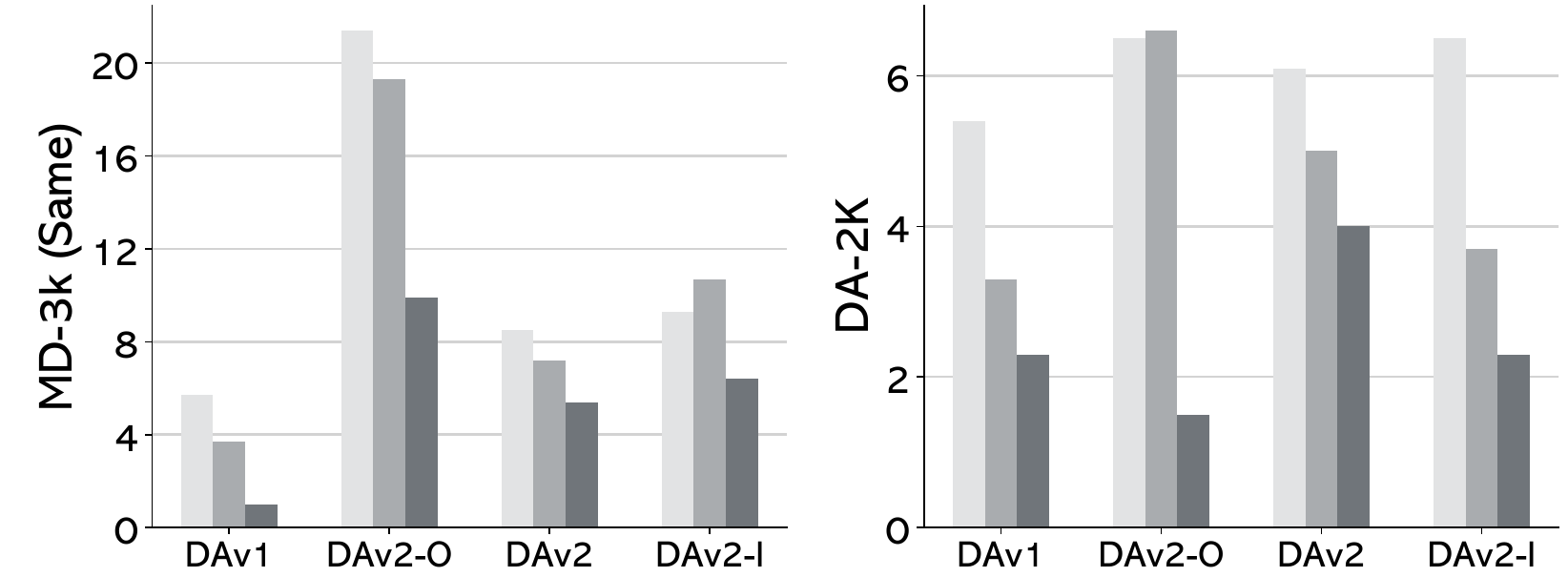}
        \centering
        \footnotesize (b) \textit{Same} subset of \texttt{MD-3k} and
        \texttt{DA-2K}
    \end{minipage}
    \setlength{\abovecaptionskip}{0.2cm}
    \caption{\textbf{Scaling analysis.} \revdel{(\textbf{a}) In ambiguous scenes with reverse multi-layer spatial relationship, scaling improves ML-SRA only when RGB and LVP biases diverge (e.g., DAv2). When biases converge, as in DAv2-O where both RGB and LVP favor the background, scaling leads to performance degradation as larger models become more committed to a single hypothesis. (\textbf{b}) In ambiguous scenes with same multi-layer spatial relationship and non-ambiguous scenes, larger models consistently generalize better to the LVP distribution shift, confirming that model capacity positively interacts with spectral prompting when cross-layer spatial relationships are consistent.}\revadd{(\textbf{a}) On the \textit{Reverse} subset of \texttt{MD-3k}, larger variants benefit most when RGB and LVP select different layers; when both inputs share the same layer bias, the candidate pair offers less complementarity. (\textbf{b}) On \textit{Same} subset of \texttt{MD-3k} and \texttt{DA-2K}, we plot the RGB/LVP ML-SRA gap in percentage points. Smaller bars mean that LVP stays closer to the RGB baseline when the ordinal relation is consistent.}}
    \label{fig:scaling_combined}
\end{figure*}

\noindent\textbf{Scaling behavior.}
Figure~\ref{fig:scaling_combined} further separates complementarity from stability. On the \textit{Reverse} subset (Fig.~\ref{fig:scaling_combined}a), scaling helps when RGB and LVP move toward different layers, as in DAv2. In contrast, when the two inputs retain the same layer bias  (most clearly in DAv2-O-L), the paired output has less room to satisfy conflicting relations. In the consistent-relation setting (Fig.~\ref{fig:scaling_combined}b), the y-axis is the RGB/LVP ML-SRA gap. As model size grows, this gap often shrinks on the \textit{Same} subset of \texttt{MD-3k} and \texttt{DA-2K}, indicating that LVP perturbs the RGB baseline less when the ordinal relation across layers is stable.

\begin{table}[t]
  \setlength{\abovecaptionskip}{0.1cm}
  \caption{\textbf{Laplacian (LVP) vs.\ Gaussian (GAU) prompts
  via ML-SRA [\%]}.
  }
  \vspace{-2mm}
  \renewcommand{\arraystretch}{1.0}
  \label{tab:ml_sra_subsets}
  \resizebox{\linewidth}{!}{%
    \begin{tabular}{l|cccccccccccccccc}
   \hline \hline
    \rotatebox{30}{\scriptsize Input} &
    \rotatebox{30}{\scriptsize Marigold} & \rotatebox{30}{\scriptsize GeoWizard} &
    \rotatebox{30}{\scriptsize ZoeDepth} & \rotatebox{30}{\scriptsize DPT} &
    \rotatebox{30}{\scriptsize DAv1-S} & \rotatebox{30}{\scriptsize DAv1-B} &
    \rotatebox{30}{\scriptsize DAv1-L} & \rotatebox{30}{\scriptsize DAv2-O-S} &
    \rotatebox{30}{\scriptsize DAv2-O-B} & \rotatebox{30}{\scriptsize DAv2-O-L} &
    \rotatebox{30}{\scriptsize DAv2-I-S} & \rotatebox{30}{\scriptsize DAv2-I-B} &
    \rotatebox{30}{\scriptsize DAv2-I-L} & \rotatebox{30}{\scriptsize DAv2-S} &
    \rotatebox{30}{\scriptsize DAv2-B} & \rotatebox{30}{\scriptsize DAv2-L} \\
    \midrule
    \multicolumn{17}{l}{\textbf{(a) \textit{Reverse} Subset of
    \texttt{MD-3k}}} \\
    \midrule
     \rowcolor[HTML]{E8F6F3} \textbf{LVP} &
    \textbf{15.3} & \textbf{17.6} & \textbf{45.4} & \textbf{46.4} &
    \textbf{17.7} & \textbf{11.4} & \textbf{10.9} & \textbf{36.6} &
    \textbf{32.9} & \textbf{17.6} & \textbf{27.7} & \textbf{28.1} &
    \textbf{42.5} & \textbf{36.9} & \textbf{48.2} & \textbf{52.2} \\
  
    \textbf{GAU} &
    6.2 & 6.0 & 0.7 & 0.3 & 0.7 & 0.3 & 0.4 & 5.7 & 2.6 & 0.6 &
    1.0 & 0.7 & 2.4 & 3.6 & 4.6 & 4.5 \\
    \midrule
    \multicolumn{17}{l}{\textbf{(b) \textit{Same} Subset of
    \texttt{MD-3k}}} \\
    \midrule
   \rowcolor[HTML]{E8F6F3}  \textbf{LVP} &
    89.8 & 91.9 & 86.8 & 88.7 & 89.0 & 91.5 & 92.8 & 83.5 & 85.6 &
    93.4 & 86.5 & 91.2 & 93.2 & 90.7 & 92.7 & 93.6 \\
   
    \textbf{GAU} &
    \textbf{94.2} & \textbf{95.3} & \textbf{94.6} & \textbf{94.3} &
    \textbf{95.7} & \textbf{96.1} & \textbf{96.5} & \textbf{92.6} &
    \textbf{94.6} & \textbf{96.4} & \textbf{95.3} & \textbf{96.6} &
    \textbf{96.9} & \textbf{97.8} & \textbf{98.1} & \textbf{98.2} \\
  \hline \hline
    \end{tabular}%
  }\vspace{-2mm}
\end{table}

\begin{figure*}[t]
    \centering
    \begin{minipage}[t]{0.54\linewidth}
        \vspace{2pt}
        \centering
        \setlength{\abovecaptionskip}{0.0cm}
        \includegraphics[width=\linewidth]{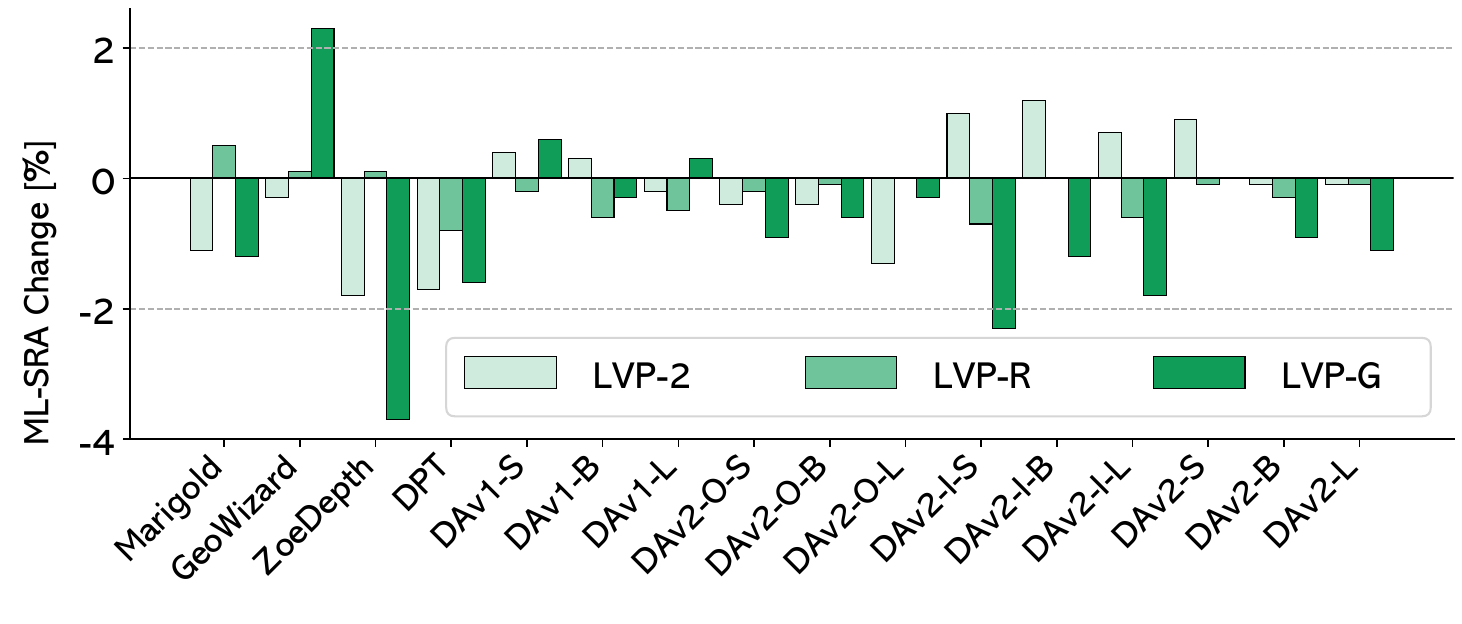}
        \caption{\textbf{Ablation of LVP design.} Relative change in
        ML-SRA [\%] compared to default LVP. Performance is robust to
        kernel variants (LVP-2: 8-neighbor), sign flip of Laplacian kernel (LVP-R), and
        grayscale input (LVP-G).}
        \label{fig:ablation_LVP}
    \end{minipage}
    \hfill
    \begin{minipage}[t]{0.44\linewidth}
        \vspace{0pt}
        \centering
        \captionsetup{type=table}
        \setlength{\abovecaptionskip}{0.1cm}
        \caption{\revdel{\textbf{LVP vs.\ Canny edge prompts.} ML-SRA [\%] on the \textit{Reverse} subset.}\revadd{\textbf{High-frequency prompt comparison.} ML-SRA [\%] values are grouped in \textit{Overall/Reverse/Same}.}}
        \label{tab:spectral_prompts}
        \vspace{1.6mm}
        \scriptsize
        \setlength{\tabcolsep}{2.7pt}
        \renewcommand{\arraystretch}{1.1}
        \begin{tabular}{l|cccc}
         \hline \hline
        \textbf{Model} & \textbf{LVP} & \revadd{\textbf{Sobel}} & \revadd{\textbf{Fourier}} & \revadd{\textbf{Wavelet}} \\
        \midrule
        \rowcolor[HTML]{E8F6F3}
        DAv2-S & \makecell{67.2\\36.9\\90.7} & \revadd{\makecell{69.8\\40.3\\92.6}} & \revadd{\makecell{66.9\\33.2\\93.0}} & \revadd{\makecell{68.7\\40.3\\90.6}} \\
        DAv2-B & \makecell{73.3\\48.2\\92.7} & \revadd{\makecell{72.2\\44.5\\93.7}} & \revadd{\makecell{72.4\\43.0\\\textbf{95.2}}} & \revadd{\makecell{73.1\\47.8\\92.6}} \\
        \rowcolor[HTML]{E8F6F3}
        DAv2-L & \makecell{\textbf{75.5}\\\textbf{52.2}\\{93.6}} & \revadd{\makecell{73.9\\47.2\\94.5}} & \revadd{\makecell{75.0\\49.1\\95.0}} & \revadd{\makecell{74.8\\50.4\\93.7}} \\
         \hline \hline
        \end{tabular}
        \ifdefined\trackedversion
        \vspace{1mm}\par\raggedright\scriptsize\color{gray}
        Deleted Canny-only comparison: DAv2-S/B/L Reverse ML-SRA was
        36.9/48.2/52.2 for LVP versus best Canny scores 24.2/39.8/49.4.
        \fi
    \end{minipage}
\end{figure*}

\vspace{1mm}
\noindent\textbf{Prompt design ablation.}
To isolate the role of the input transform, we compare LVP against a
low-pass Gaussian prompt in Table~\ref{tab:ml_sra_subsets}. On the
\textit{Reverse} subset, the Gaussian prompt fails across all models (near-zero
scores), as low-frequency preservation merely retains the primary depth
hypothesis. \revdel{LVP succeeds, confirming high-frequency perturbation as the critical driver of mode-switching.}\revadd{LVP succeeds, supporting high-frequency emphasis as an operational factor associated with preference modulation.} This trade-off is spectrally
asymmetric: on the \textit{Same} subset, the Gaussian outperforms LVP across all models, since low-frequency preservation suffices for unambiguous scenes
while high-frequency perturbation unnecessarily disrupts the primary
hypothesis. \revdel{LVP is thus selectively effective in the most geometrically ambiguous cases where layers exhibit conflicting depth orderings.}\revadd{This contrast separates two regimes: high-frequency prompting helps when the two layers impose conflicting orderings, whereas low-frequency preservation is better when one ordering is already sufficient.}
Performance is stable across kernel variations (4 vs.\ 8 neighbors),
sign flips, and grayscale conversion (Fig.~\ref{fig:ablation_LVP}),
\revdel{indicating that efficacy arises from the spectral operation itself rather than implementation details of the discretizations of the operator}\revadd{which is consistent with sensitivity to broad high-frequency emphasis rather than dependence on one Laplacian discretization}. 

\vspace{1mm}
\noindent\revdel{\textbf{Comparison to Canny edges.}
The original manuscript compared LVP only with binary Canny maps; LVP reached 52.2\% Reverse ML-SRA on DAv2-L versus 49.4\% for the best Canny setting.}

\vspace{1mm}
\noindent\textbf{Alternative spectral prompts.}
We then test whether the effect is specific to the Laplacian operator. Table~\ref{tab:spectral_prompts} evaluates Sobel, Fourier high-pass, and wavelet prompts. No operator dominates every model and subset: Sobel matches the best Reverse score for DAv2-S, while LVP is strongest for DAv2-B/L. The shared effect across continuous high-frequency operators supports the broader spectral-modulation finding; LVP remains a simple, parameter-free default rather than a uniquely privileged transform under this benchmark.

\subsection{Downstream Applications}
\label{sec:applications}

\begin{figure}[t]
    \centering
    \setlength{\abovecaptionskip}{0.2cm}
    \includegraphics[width=\linewidth]{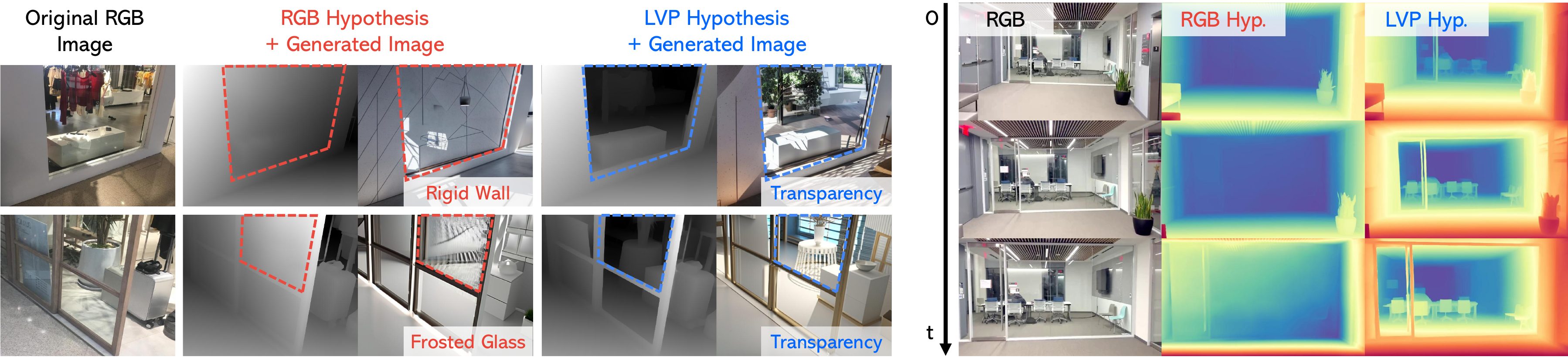}
   \caption{\textbf{Downstream illustrations.}
    Selected RGB/LVP-conditioned depth hypotheses provide alternative ControlNet conditions and frame-wise depth streams. }
    \label{fig:depth2rgb}\label{fig:video_depth}\vspace{-1mm}
\end{figure}

\noindent\textbf{Conditional generation and video hypotheses.}
In Fig.~\ref{fig:depth2rgb}, the RGB/LVP pair provides candidate depth conditions from the same frozen model rather than a final layered
reconstruction. In conditional generation, selected hypotheses can drive ControlNet~\cite{controlnet} renderings that emphasize different visible geometry while keeping the RGB scene fixed. In video, applying RGB and LVP frame by frame produces distinct depth streams, making layered ambiguity visible over time.

\begin{figure}[t]
    \setlength{\abovecaptionskip}{0.2cm}
    \includegraphics[width=\linewidth]{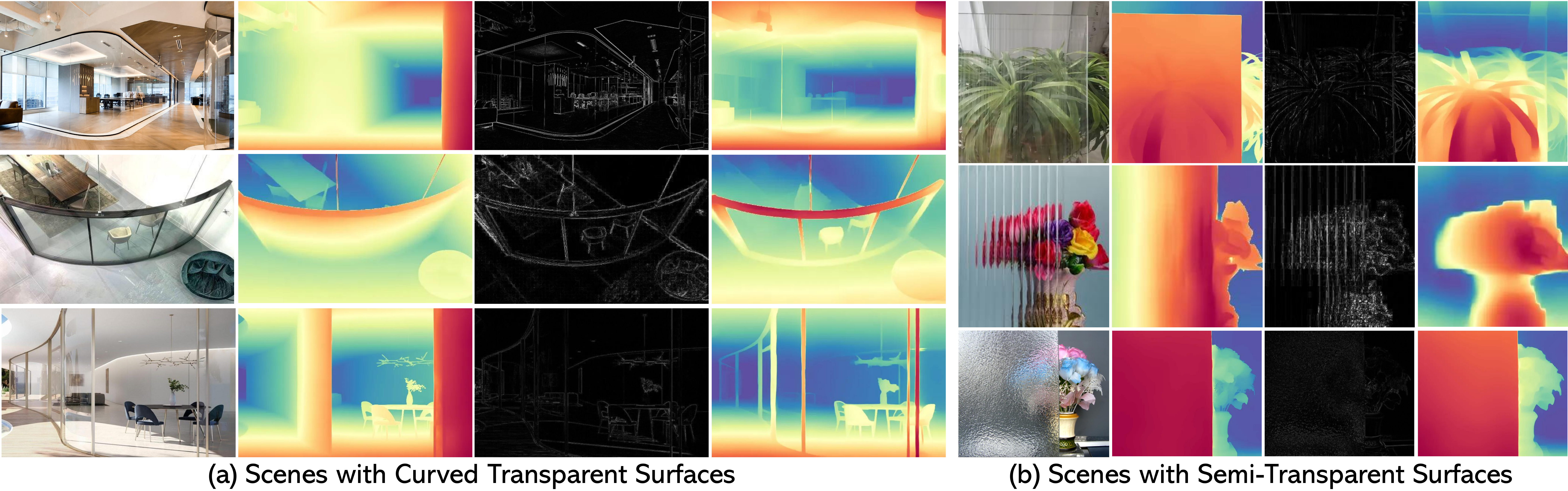}
    \caption{\textbf{Generalization and failure cases of LVP.}
    LVP can modulate depth-layer preference in challenging curved-glass scenes, but may fail on semi-transparent surfaces where foreground and background cues are frequency-entangled.}
    \label{fig:generalization_boundary}
\end{figure}\vspace{-1mm}

%% file: sec/5_conclusion.tex
\section{Discussion and Conclusions}
\label{sec:conclusion}
\revadd{Ambiguous layered scenes reveal a fundamental limitation of single-depth estimation: a single scalar target collapses multiple geometrically valid ray-wise interpretations into one dataset-dependent layer convention. \texttt{MD-3k} makes this convention explicit. Leading depth foundation models exhibit diverse RGB layer preferences, while LVP shows that some frozen models can be modulated to express complementary layer hypotheses without retraining. These findings suggest that standard RGB inference captures only one slice of a richer geometric posterior, and that future depth systems should represent, evaluate, and learn from multiple plausible scene depths rather than treating ambiguity as noise.}

\vspace{1mm}
\noindent\textbf{Limitations and future work.}
\revadd{As shown in Fig.~\ref{fig:generalization_boundary}, curved glass can preserve separable cues and produce a useful foreground shift, while textured transparent surfaces can entangle layer frequencies and fail. LVP is therefore model-dependent and should not be treated as a reliable layer extractor. \texttt{MD-3k} is sparsely-annotated and focused on transparent scenes, leaving dense validation, automatic layer selection, and broader ambiguous-scene benchmarks to future work.}

\section*{Acknowledgments}
The authors gratefully acknowledge Modal Labs for providing partial support through a generous academic compute grant.

%% file: sec/X_appendix.tex
\setcounter{table}{0}
\setcounter{figure}{0}
\setcounter{section}{0}
\renewcommand{\thetable}{\Alph{table}}
\renewcommand{\thefigure}{\Alph{figure}}
\renewcommand{\thesection}{\Alph{section}}

\section{More Quantitative Results}

\begin{table*}[t]
\centering
\scriptsize
\setlength\tabcolsep{1.8mm}
\renewcommand{\arraystretch}{1.3}
\caption{\textbf{ML-SRA on \texttt{MD-3k} with alternative high-frequency prompts.} Each cell reports \textit{Overall/Reverse/Same} [\%]. The effect generalizes beyond the Laplacian, but no operator dominates across all model families.}
\label{tab:supp_spectral_prompts}
\begin{tabular}{l|c|c|c|c}
\hline \hline
\textbf{Model} & \textbf{LVP} & \textbf{Sobel} & \textbf{Fourier high-pass} & \textbf{Wavelet} \\
\midrule
DAv2-O-S & 63.0/36.6/83.5 & 60.0/25.7/86.5 & 58.6/25.1/84.4 & 60.7/28.7/85.4 \\
DAv2-O-B & 62.7/32.9/85.6 & 60.3/24.2/88.2 & 58.8/23.7/86.0 & 62.2/30.6/86.6 \\
DAv2-O-L & 60.4/17.6/93.4 & 59.5/14.8/94.0 & 60.7/18.1/93.7 & 60.3/17.8/93.2 \\
\midrule
DAv2-I-S & 60.9/27.7/86.5 & 64.3/32.0/89.3 & 62.0/25.2/90.5 & 62.9/30.0/88.3 \\
DAv2-I-B & 63.7/28.1/91.2 & 64.1/27.0/92.8 & 65.2/29.7/92.6 & 64.5/29.5/91.5 \\
DAv2-I-L & 71.1/42.5/93.2 & 67.3/32.8/94.0 & 69.5/38.5/93.6 & 70.2/39.8/93.6 \\
\midrule
DAv2-S & 67.2/36.9/90.7 & 69.8/40.3/92.6 & 66.9/33.2/93.0 & 68.7/40.3/90.6 \\
DAv2-B & 73.3/48.2/92.7 & 72.2/44.5/93.7 & 72.4/43.0/\textbf{95.2} & 73.1/47.8/92.6 \\
DAv2-L & \textbf{75.5/52.2}/93.6 & 73.9/47.2/94.5 & 75.0/49.1/95.0 & 74.8/50.4/93.7 \\
\hline \hline
\end{tabular}
\end{table*}

\subsection{Alternative High-Frequency Prompts}
Table~\ref{tab:supp_spectral_prompts} compares LVP with Sobel, Fourier high-pass, and wavelet prompts on \texttt{MD-3k}. These alternatives also modulate the expressed layer preference, supporting a family of high-frequency diagnostics rather than a Laplacian-specific hidden-state claim. No operator is universally best: LVP gives the highest overall score for five of the nine DAv2 variants, including DAv2-B/L.

\begin{table*}[t]
\centering
\scriptsize
\setlength\tabcolsep{1.3mm}
\renewcommand{\arraystretch}{1.3}
\caption{\textbf{Zero-shot relative-depth performance on non-ambiguous datasets.} LVP is compared with standard RGB input using AbsRel ($\downarrow$) and $\delta_1/\delta_2/\delta_3$ ($\uparrow$). Losses are small on KITTI but substantial on ETH3D for several models; LVP is therefore an ambiguity probe rather than a universal RGB replacement.}
\label{tab:supp_non_ambiguous}
\resizebox{\textwidth}{!}{%
\begin{tabular}{l|c|cccc|cccc|cccc}
\hline \hline
\multirow{2}{*}{\textbf{Model}} & \multirow{2}{*}{\textbf{Input}} & \multicolumn{4}{c|}{\textbf{NYU-D}} & \multicolumn{4}{c|}{\textbf{KITTI}} & \multicolumn{4}{c}{\textbf{ETH3D}} \\
\cmidrule(lr){3-6}\cmidrule(lr){7-10}\cmidrule(lr){11-14}
& & AbsRel & $\delta_1$ & $\delta_2$ & $\delta_3$ & AbsRel & $\delta_1$ & $\delta_2$ & $\delta_3$ & AbsRel & $\delta_1$ & $\delta_2$ & $\delta_3$ \\
\midrule
\multirow{2}{*}{DAv1-S} & RGB & 0.140 & 83.42 & 94.65 & 97.12 & 0.361 & 41.46 & 78.41 & 87.53 & 0.137 & 83.68 & 94.70 & 97.66 \\
& LVP & 0.172 & 77.21 & 92.70 & 96.94 & 0.377 & 39.64 & 74.64 & 86.80 & 0.197 & 72.77 & 90.88 & 96.33 \\
\midrule
\multirow{2}{*}{DAv1-B} & RGB & 0.137 & 83.83 & 94.81 & 97.14 & 0.367 & 40.42 & 77.77 & 87.29 & 0.132 & 84.57 & 94.91 & 97.69 \\
& LVP & 0.154 & 80.67 & 93.91 & 97.13 & 0.373 & 40.11 & 75.60 & 86.85 & 0.185 & 74.61 & 91.74 & 96.73 \\
\midrule
\multirow{2}{*}{DAv1-L} & RGB & 0.137 & 83.82 & 94.83 & 97.15 & 0.364 & 40.96 & 78.13 & 87.42 & 0.129 & 85.05 & 95.02 & 97.75 \\
& LVP & 0.153 & 80.98 & 93.90 & 97.02 & 0.370 & 40.45 & 76.66 & 87.20 & 0.174 & 76.77 & 92.30 & 96.85 \\
\midrule
\multirow{2}{*}{DAv2-S} & RGB & 0.139 & 83.87 & 94.80 & 97.05 & 0.360 & 41.87 & 78.53 & 87.63 & 0.142 & 82.57 & 94.09 & 97.35 \\
& LVP & 0.164 & 78.45 & 93.28 & 97.12 & 0.376 & 39.93 & 74.87 & 86.99 & 0.190 & 74.18 & 91.36 & 96.54 \\
\midrule
\multirow{2}{*}{DAv2-B} & RGB & 0.140 & 83.85 & 94.83 & 97.01 & 0.364 & 41.33 & 77.92 & 87.33 & 0.137 & 83.53 & 94.35 & 97.43 \\
& LVP & 0.154 & 80.59 & 93.97 & 97.12 & 0.367 & 41.10 & 76.34 & 87.02 & 0.182 & 75.38 & 91.87 & 96.77 \\
\midrule
\multirow{2}{*}{DAv2-L} & RGB & 0.139 & 83.92 & 94.86 & 97.00 & 0.363 & 41.26 & 78.54 & 87.61 & 0.136 & 83.85 & 94.44 & 97.48 \\
& LVP & 0.152 & 81.04 & 94.01 & 97.04 & 0.367 & 41.35 & 76.77 & 87.10 & 0.177 & 76.17 & 92.09 & 96.81 \\
\hline \hline
\end{tabular}}
\end{table*}

{\revaddcolor
\subsection{Zero-Shot Dense Depth Fidelity on Non-Ambiguous Benchmarks}
Table~\ref{tab:supp_non_ambiguous} evaluates whether LVP preserves standard dense-depth fidelity on NYU-D~\cite{nyud}, KITTI~\cite{kitti}, and ETH3D~\cite{eth3d}. The results show a consistent trade-off: LVP increases sensitivity to an alternate depth hypothesis but usually reduces accuracy under conventional single-depth metrics. Across the 18 RGB/LVP comparisons in the table, AbsRel increases by 0.003--0.060; $\delta_1$ ranges from a 0.09-point gain to a 10.91-point drop, $\delta_2$ drops by 0.85--3.82 points, and $\delta_3$ ranges from a 0.11-point gain to a 1.33-point drop. The degradation is modest on KITTI and largest on ETH3D, confirming that LVP should be used as an ambiguity probe rather than a replacement for standard RGB inference on non-ambiguous benchmarks.
}

\subsection{Comparison between LVP and Canny Binary Edge Prompts}
\label{sec:supp_canny_prompt}

Table~\ref{tab:supp_canny_prompt} reports the comparison between LVP and Canny
binary edge prompts on the \textit{Reverse} subset of \texttt{MD-3k}. We focus
this controlled comparison on the non-finetuned DAv2-S/B/L model family, where
the main LVP effect is most pronounced. We evaluate four binary Canny edge
prompts with increasing hysteresis thresholds, Edge-1: $(50,150)$, Edge-2:
$(60,180)$, Edge-3: $(70,210)$, and Edge-4: $(80,240)$. LVP consistently
outperforms the best Canny variant across DAv2-S/B/L. This supports the
interpretation that high-frequency prompting is useful under conflicting layer
orderings, while also suggesting that the continuous Laplacian residual contains
richer structural information than a binarized edge map. We leave a broader
comparison across more generic models to future work.

\begin{table}[t]
\centering
\scriptsize
\setlength{\tabcolsep}{3.8mm}
\renewcommand{\arraystretch}{1.3}
\caption{\textbf{Comparison between LVP and Canny binary edge prompts on
\texttt{MD-3k} \textit{Reverse}.}
We report ML-SRA [\%] for the non-finetuned DAv2-S/B/L model family.
Edge-1--Edge-4 denote Canny binary edge prompts with increasing low/high
hysteresis thresholds: $(50,150)$, $(60,180)$, $(70,210)$, and $(80,240)$.}
\label{tab:supp_canny_prompt}
\begin{tabular}{l|c|cccc}
\hline \hline
\textbf{Model} & \textbf{LVP} & \textbf{Edge-1} & \textbf{Edge-2} & \textbf{Edge-3} & \textbf{Edge-4} \\
\midrule
DAv2-S & \textbf{36.9} & 24.2 & 18.8 & 17.3 & 16.0 \\
DAv2-B & \textbf{48.2} & 39.8 & 32.4 & 30.6 & 28.1 \\
DAv2-L & \textbf{52.2} & 49.4 & 39.8 & 37.3 & 35.9 \\
\hline \hline
\end{tabular}
\end{table}

\subsection{Depth-Layer Preference Values}
\label{sec:supp_depth_preference_values}

Table~\ref{tab:depth_preference_values} reports the numerical values underlying
the depth-layer preference visualization in the main paper. These
values are computed from the per-layer SRA on the \textit{Reverse} subset of
\texttt{MD-3k} as $\alpha=\mathrm{SRA}(2)-\mathrm{SRA}(1)$. Positive values
indicate that the model output agrees more with the visible background layer,
whereas negative values indicate stronger agreement with the transparent foreground
layer. The final column reports $\Delta\alpha=\alpha_{\mathrm{LVP}}-\alpha_{\mathrm{RGB}}$,
which summarizes how much LVP changes the expressed
layer preference of each frozen model.

\begin{table}[t]
\centering
\scriptsize
\setlength{\tabcolsep}{3.2mm}
\renewcommand{\arraystretch}{1.3}
\caption{\textbf{Depth-layer preference values.}
We compute $\alpha=\mathrm{SRA}(2)-\mathrm{SRA}(1)$ on the \textit{Reverse}
subset of \texttt{MD-3k}. Positive values indicate preference for the visible
background layer; negative values indicate preference for the transparent
foreground layer. $\Delta\alpha=\alpha_{\mathrm{LVP}}-\alpha_{\mathrm{RGB}}$.}
\label{tab:depth_preference_values}
\begin{tabular}{lccc}
\hline \hline
\textbf{Model} & \textbf{RGB $\alpha$} & \textbf{LVP $\alpha$} & \textbf{$\Delta\alpha$} \\
\midrule
Marigold        & $+20.8$ & $+15.4$ & $-5.4$ \\
GeoWizard       & $+35.4$ & $+19.4$ & $-16.0$ \\
ZoeDepth        & $+76.0$ & $-8.6$  & $-84.6$ \\
DPT             & $+79.6$ & $-10.4$ & $-90.0$ \\
\midrule
Depth Pro       & $-35.2$ & $+7.6$  & $+42.8$ \\
UniDepth-v2-L   & $+45.4$ & $+45.8$ & $+0.4$ \\
UniK3D-L        & $+56.4$ & $+39.4$ & $-17.0$ \\
\midrule
DAv1-S          & $+88.4$ & $+57.0$ & $-31.4$ \\
DAv1-B          & $+88.4$ & $+70.8$ & $-17.6$ \\
DAv1-L          & $+90.4$ & $+72.8$ & $-17.6$ \\
\midrule
DAv2-O-S        & $+13.2$ & $-48.8$ & $-62.0$ \\
DAv2-O-B        & $+23.0$ & $-26.2$ & $-49.2$ \\
DAv2-O-L        & $+27.6$ & $+18.4$ & $-9.2$ \\
\midrule
DAv2-I-S        & $-21.2$ & $+9.8$  & $+31.0$ \\
DAv2-I-B        & $-30.6$ & $+8.0$  & $+38.6$ \\
DAv2-I-L        & $-39.2$ & $+33.0$ & $+72.2$ \\
\midrule
DAv2-S          & $-4.0$  & $+55.8$ & $+59.8$ \\
DAv2-B          & $-23.4$ & $+64.6$ & $+88.0$ \\
DAv2-L          & $-30.6$ & $+68.2$ & $+98.8$ \\
\hline \hline
\end{tabular}
\end{table}

\clearpage

\section{More Qualitative Results}

\begin{figure}[t!] 
    \centering\setlength{\abovecaptionskip}{0.4cm}
    \includegraphics[width=0.76\textwidth]{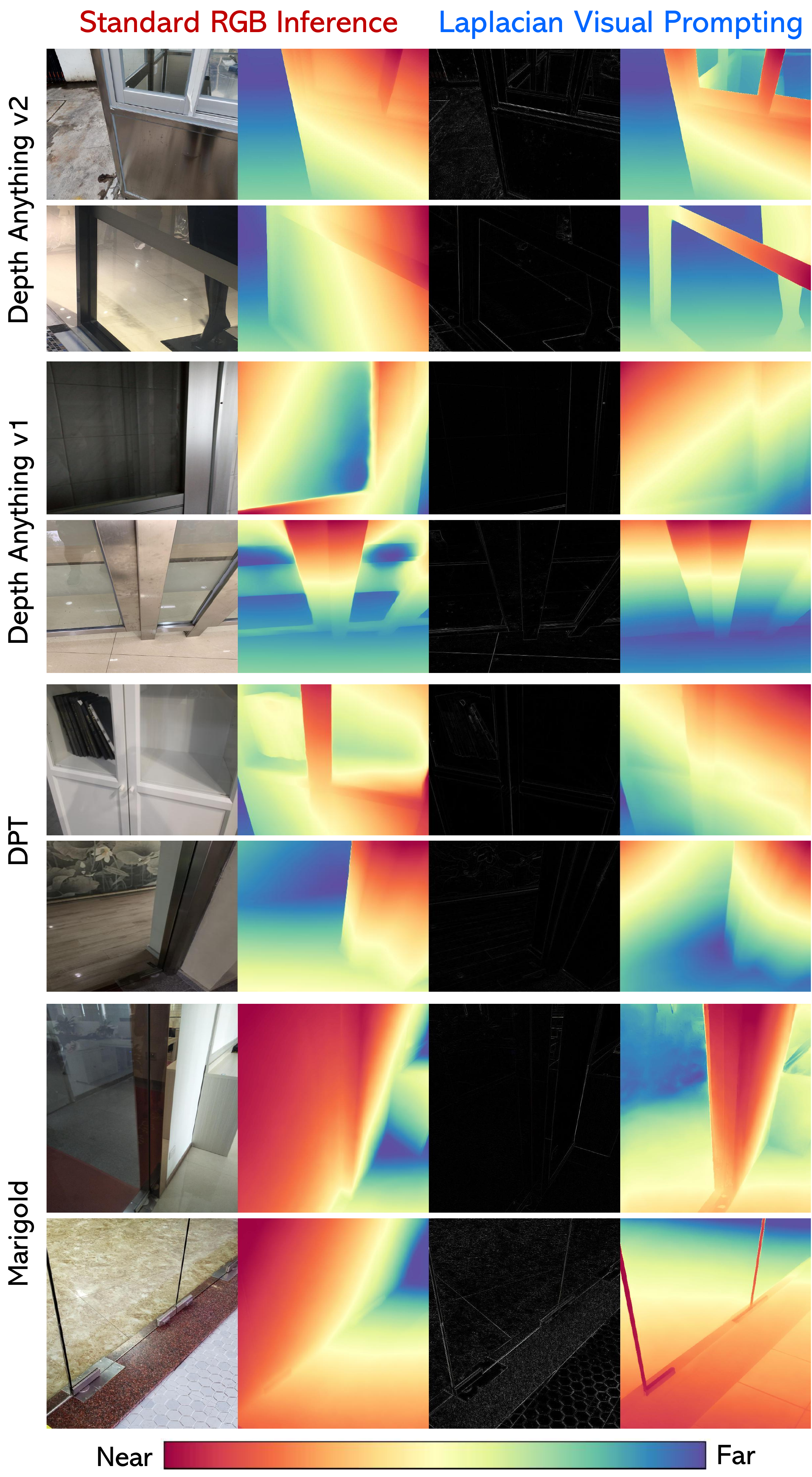}
\caption{\revdel{\textbf{Unlocking \textit{hidden depth} with Laplacian Visual Prompting on diverse models~\cite{depth_anything,yang2024depth,marigold,dpt}.}}\revadd{\textbf{Model-dependent output modulation with LVP~\cite{depth_anything,yang2024depth,marigold,dpt}.}} Each case shows the RGB input and its depth estimate, followed by the LVP input and corresponding output from the same frozen model.}
    \label{fig:hidden-depth-appendix}
\end{figure}

\begin{figure*}[t!]
\centering \setlength{\abovecaptionskip}{0.2cm}
\includegraphics[width=\textwidth]{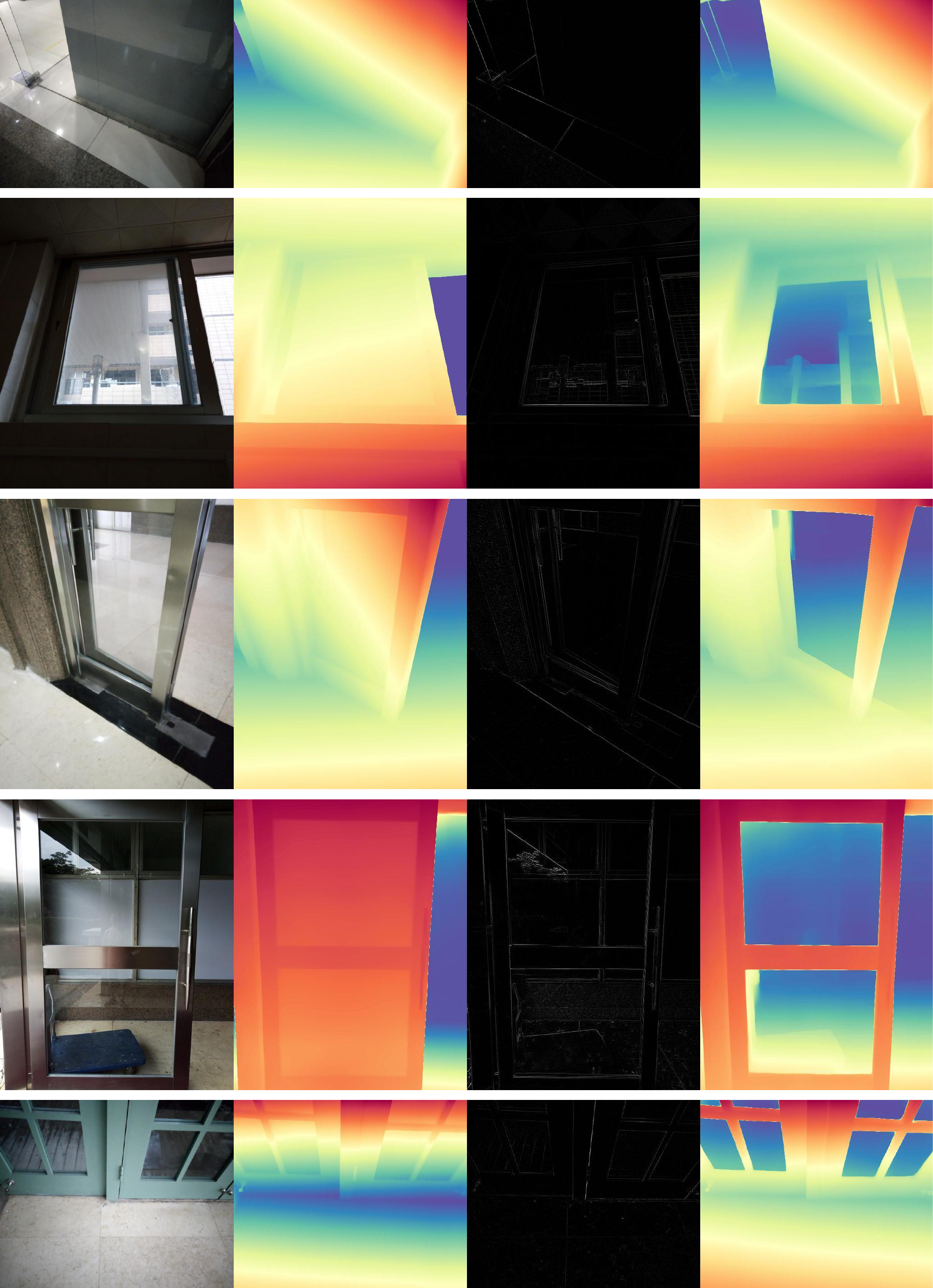}
\caption{\revdel{\textbf{LVP-empowered multi-layer depth.}}\revadd{\textbf{RGB/LVP output hypotheses.}} Each case includes an RGB image and its depth output, followed by the Laplacian input and its output.}
\label{fig:hidden_depth_2_1}
\end{figure*}

\begin{figure*}[t!]
\centering \setlength{\abovecaptionskip}{0.2cm}
\includegraphics[width=\textwidth]{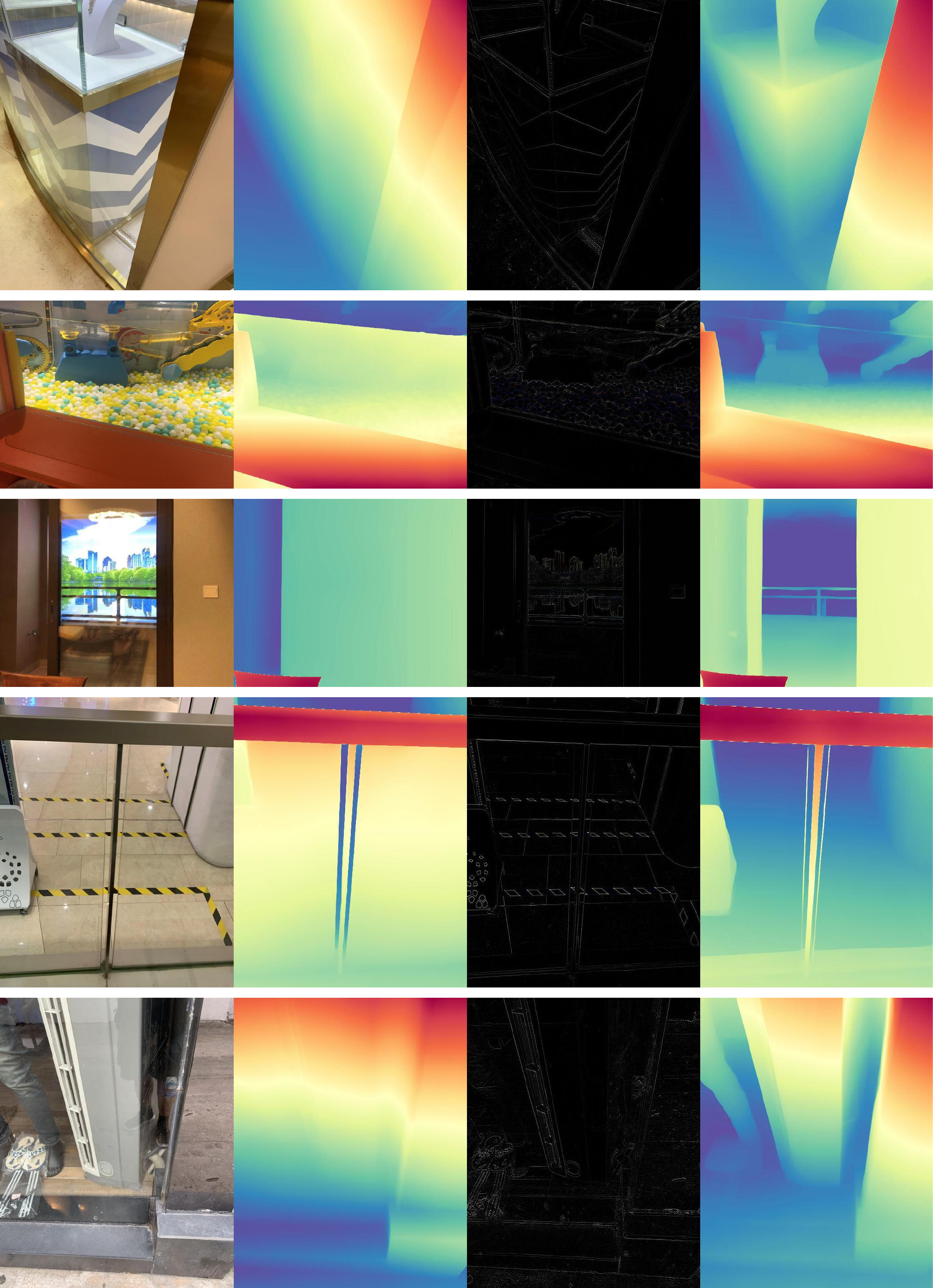}
\caption{\revdel{\textbf{LVP-empowered multi-layer depth.}}\revadd{\textbf{RGB/LVP output hypotheses.}} Each case includes an RGB image and its depth output, followed by the Laplacian input and its output.}
\label{fig:hidden_depth_2_2}
\end{figure*}

\begin{figure*}[t!]
\centering \setlength{\abovecaptionskip}{0.2cm}
\includegraphics[width=\textwidth]{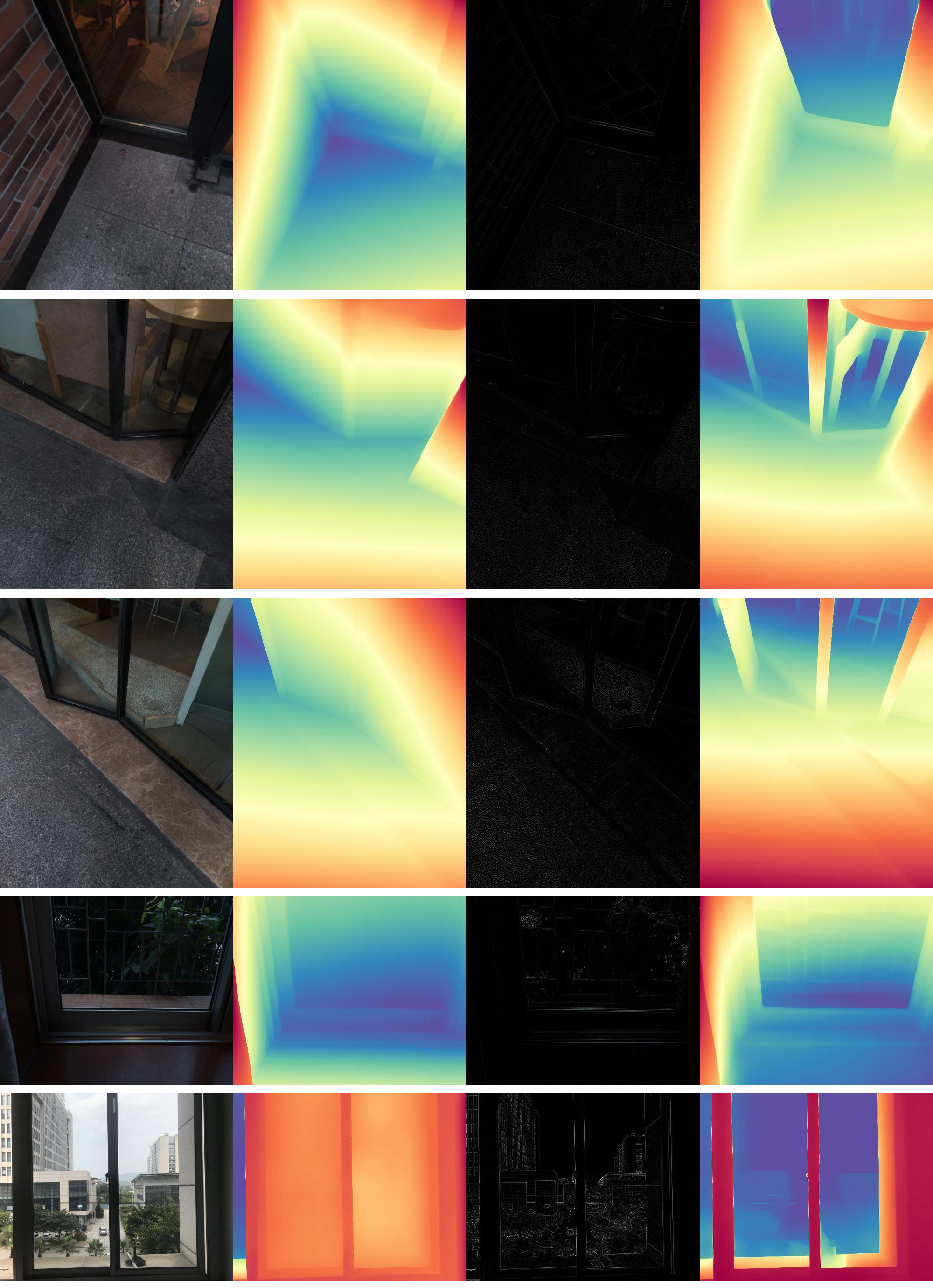}
\caption{\revdel{\textbf{LVP-empowered multi-layer depth.}}\revadd{\textbf{RGB/LVP output hypotheses.}} Each case includes an RGB image and its depth output, followed by the Laplacian input and its output.}
\label{fig:hidden_depth_2_3}
\end{figure*}

\begin{figure*}[t!]
\centering \setlength{\abovecaptionskip}{0.2cm}
\includegraphics[width=\textwidth]{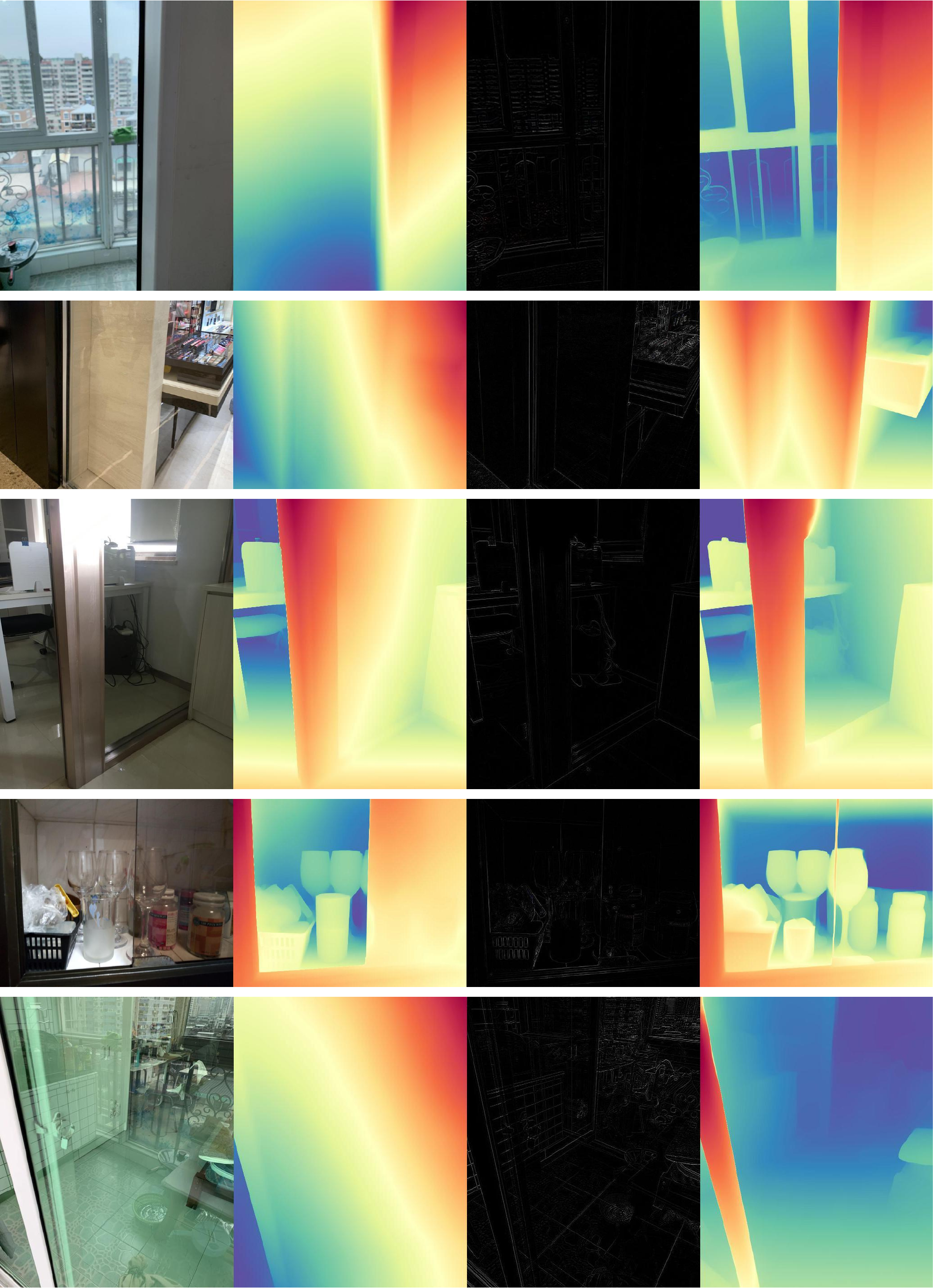}
\caption{\revdel{\textbf{LVP-empowered multi-layer depth.}}\revadd{\textbf{RGB/LVP output hypotheses.}} Each case includes an RGB image and its depth output, followed by the Laplacian input and its output.}
\label{fig:hidden_depth_2_4}
\end{figure*}

\begin{figure*}[t!]
\centering \setlength{\abovecaptionskip}{0.2cm}
\includegraphics[width=\textwidth]{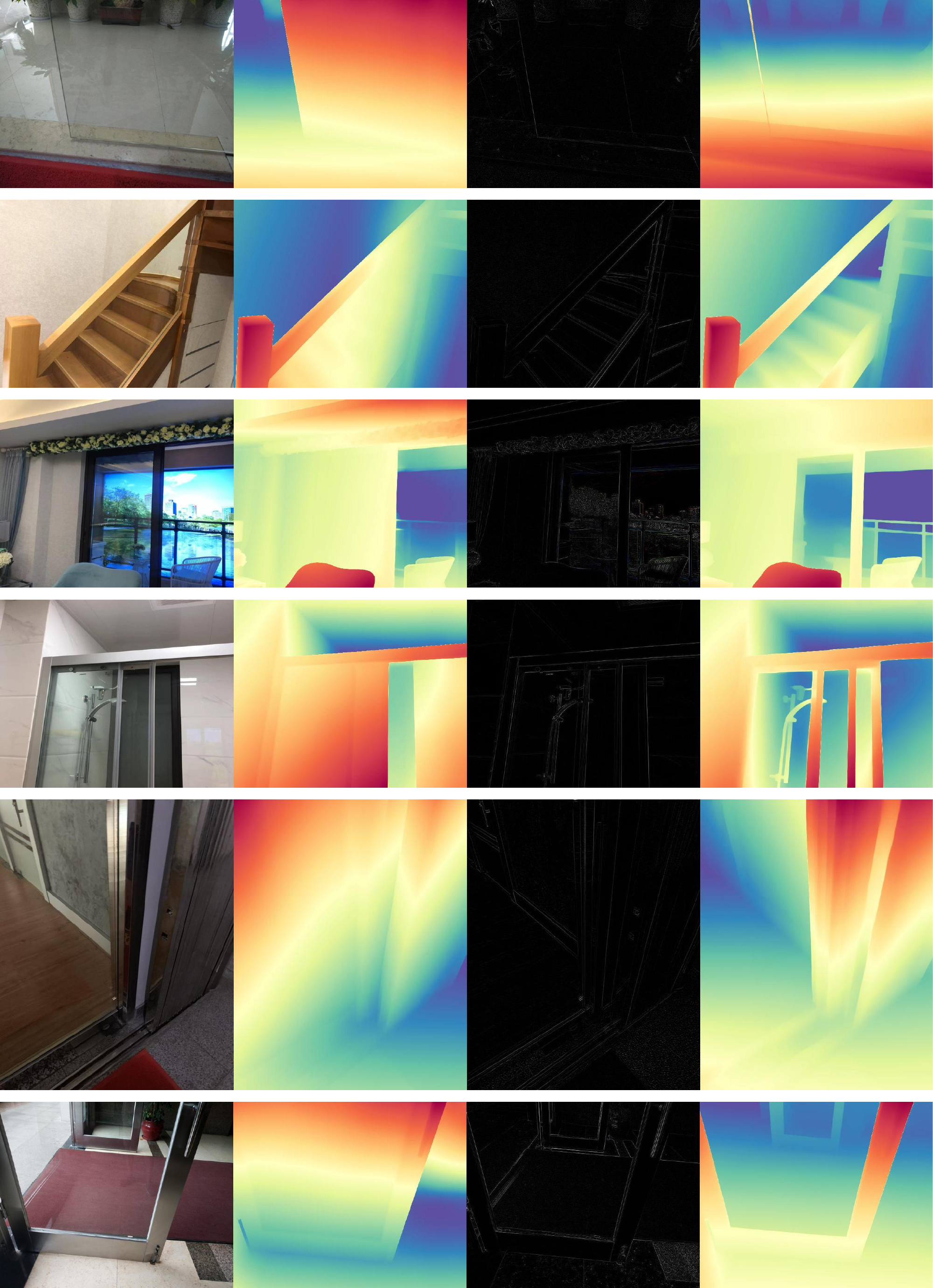}
\caption{\revdel{\textbf{LVP-empowered multi-layer depth.}}\revadd{\textbf{RGB/LVP output hypotheses.}} Each case includes an RGB image and its depth output, followed by the Laplacian input and its output.}
\label{fig:hidden_depth_2_5}
\end{figure*}

\begin{figure*}[t!]
\centering \setlength{\abovecaptionskip}{0.2cm}
\includegraphics[width=\textwidth]{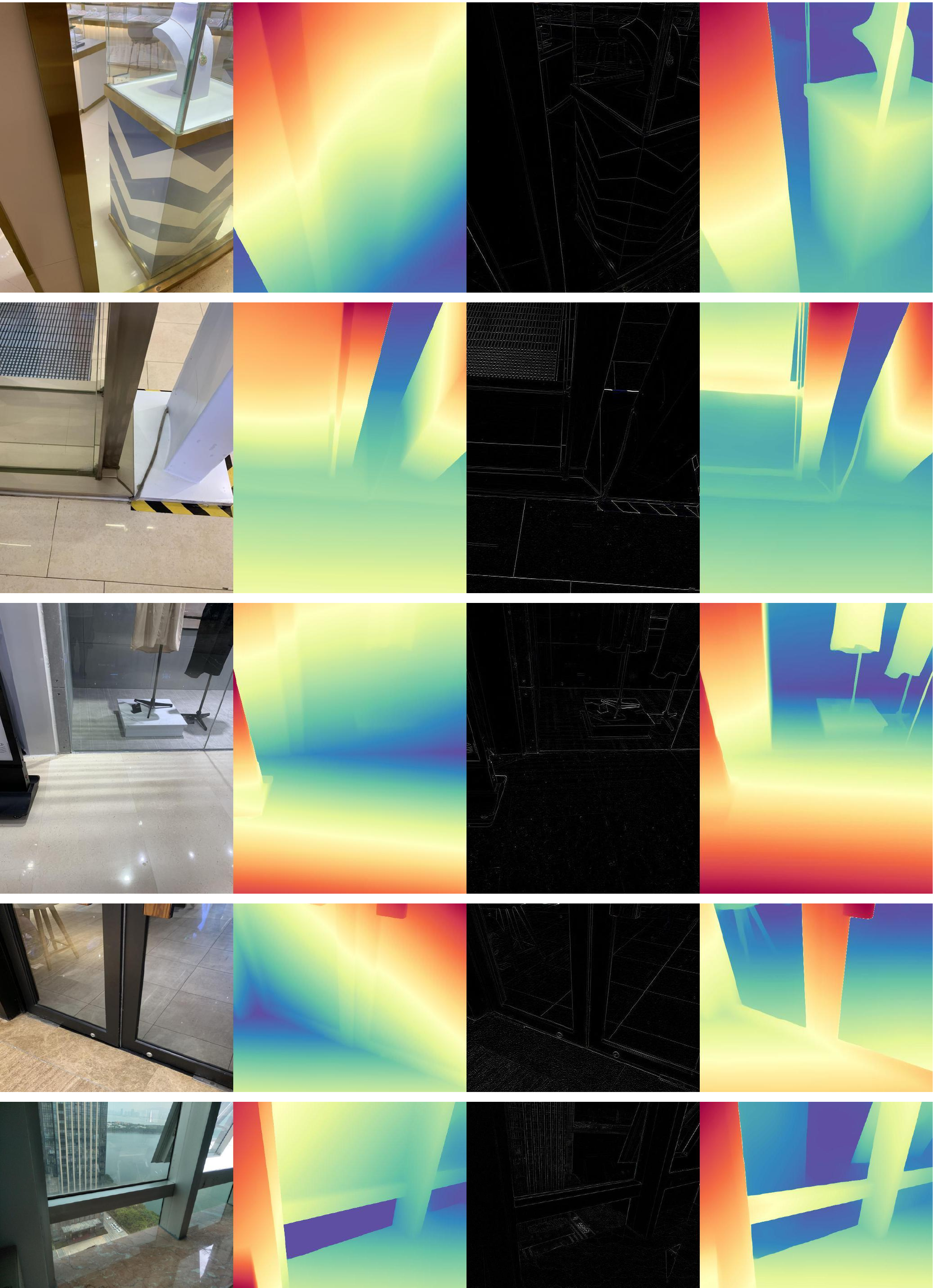}
\caption{\revdel{\textbf{LVP-empowered multi-layer depth.}}\revadd{\textbf{RGB/LVP output hypotheses.}} Each case includes an RGB image and its depth output, followed by the Laplacian input and its output.}
\label{fig:hidden_depth_2_6}\vspace{-2mm}
\end{figure*}

\noindent\revdel{\textbf{Multi-layer depth decoupling with Laplacian Visual Prompting.} In addition to demonstrating the effectiveness of LVP in revealing hidden depth layers in various models (Fig.~\ref{fig:hidden-depth-appendix}), we further demonstrate its capabilities using the best-performing baseline model, \textit{i.e.}, the Depth Anything v2-Large (DAv2-L) model, in Figures~\ref{fig:hidden_depth_2_1} through~\ref{fig:hidden_depth_2_6}. LVP effectively elicits alternative depth hypotheses, particularly in scenes with transparency and occlusion. Although conventional depth estimations often fail to capture the layered nature of these scenes, collapsing multiple depths into a single layer, Laplacian-prompted depth maps reveal previously hidden depth layers, clearly delineating transparent surfaces and occluded objects behind the transparent surfaces.}\revadd{\textbf{Additional qualitative examples of depth-layer modulation.} Figure~\ref{fig:hidden-depth-appendix} and Figs.~\ref{fig:hidden_depth_2_1}--\ref{fig:hidden_depth_2_6} show RGB- and LVP-conditioned outputs, with DAv2-L used for the extended examples. In receptive models, LVP often changes the output ordering in transparent regions while preserving recognizable scene structure. These examples illustrate the behavioral effect; they do not establish that two discrete depth maps are stored internally.}

\begin{figure*}[t!] 
    \centering\setlength{\abovecaptionskip}{0.2cm}
    \includegraphics[width=\textwidth]{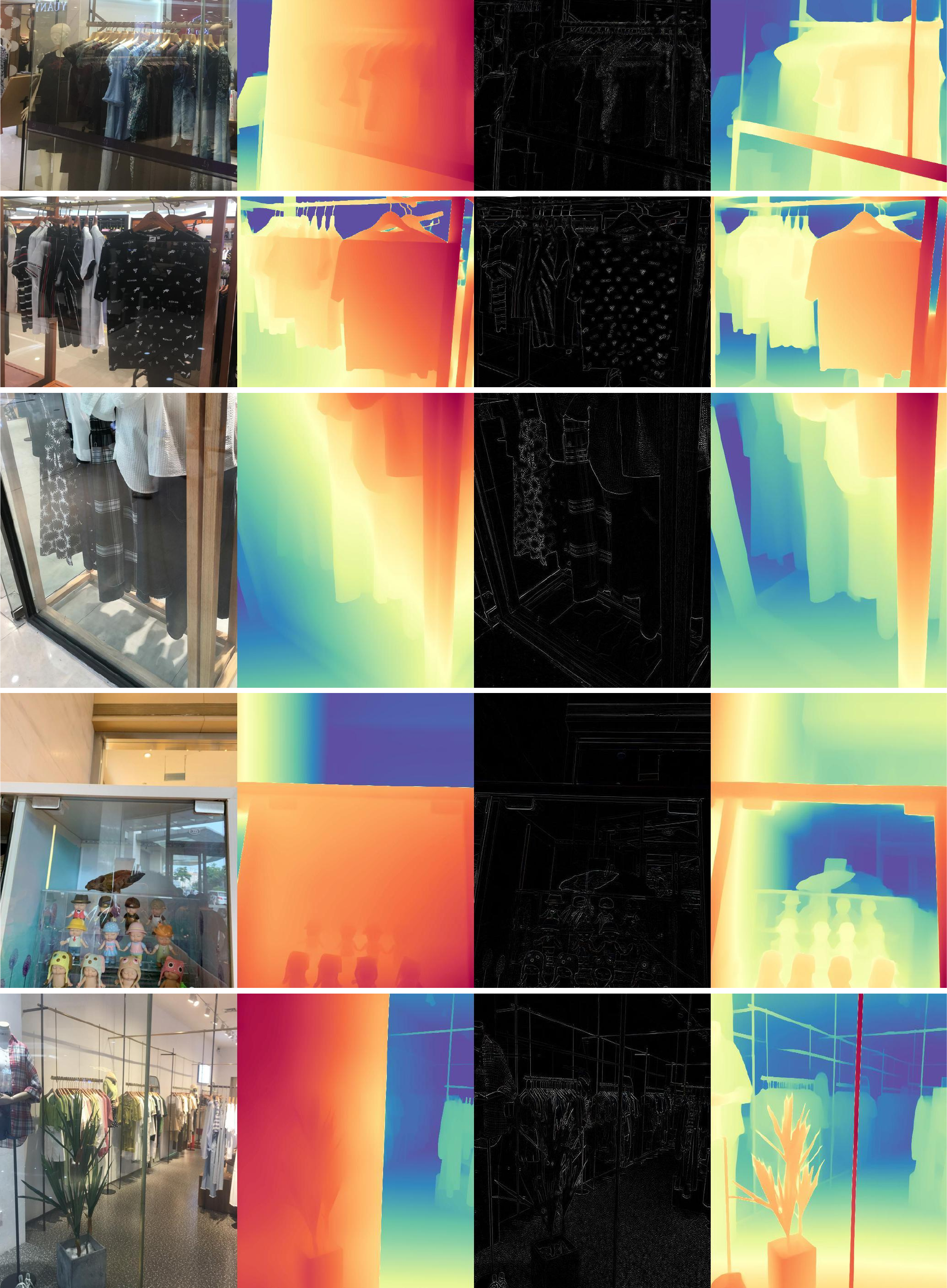}
\caption{\revdel{\textbf{{Failure cases} of multi-layer depth estimation via Laplacian Visual Prompting.}}\revadd{\textbf{Failure cases of LVP-conditioned output modulation.}} Each case shows the RGB input and output followed by the Laplacian input and output.}
    \label{fig:failure_cases}\vspace{-2mm}
\end{figure*}

\vspace{1mm}
\noindent\revdel{\textbf{Failure cases in multi-layer depth decoupling with Laplacian Visual Prompting.} Despite significant improvements in multi-layer depth estimation, LVP is not immune to failure cases, especially due to its training-free nature. We present these failures in Fig.~\ref{fig:failure_cases}. One type occurs when the initial single-layer depth prediction from the RGB input is already incorrect. In these instances, LVP struggles to correct the depth bias, resulting in inaccurate multi-layer depth predictions. This is particularly problematic when the RGB model's depth map contains errors in ambiguous regions, limiting LVP's ability to produce accurate alternative depth hypotheses. The second type of failure happens when LVP predicts a depth layer similar to the RGB output, failing to decouple distinct depth layers. This occurs when LVP cannot extract sufficient high-frequency information from the Laplacian mask to differentiate between near and far surfaces, especially in scenes with low depth contrast or complex occlusions.}\revadd{\textbf{Failure cases.} Figure~\ref{fig:failure_cases} shows two common boundaries: LVP cannot repair an already incorrect RGB hypothesis, and it may return nearly the same ordering as RGB when spectral separation is weak. The latter occurs in low-contrast, highly textured, or complexly occluded scenes, where high-frequency emphasis does not isolate a complementary structural cue.}

\begin{figure*}[t!]
\centering  \setlength{\abovecaptionskip}{0.2cm}
\includegraphics[width=\textwidth]{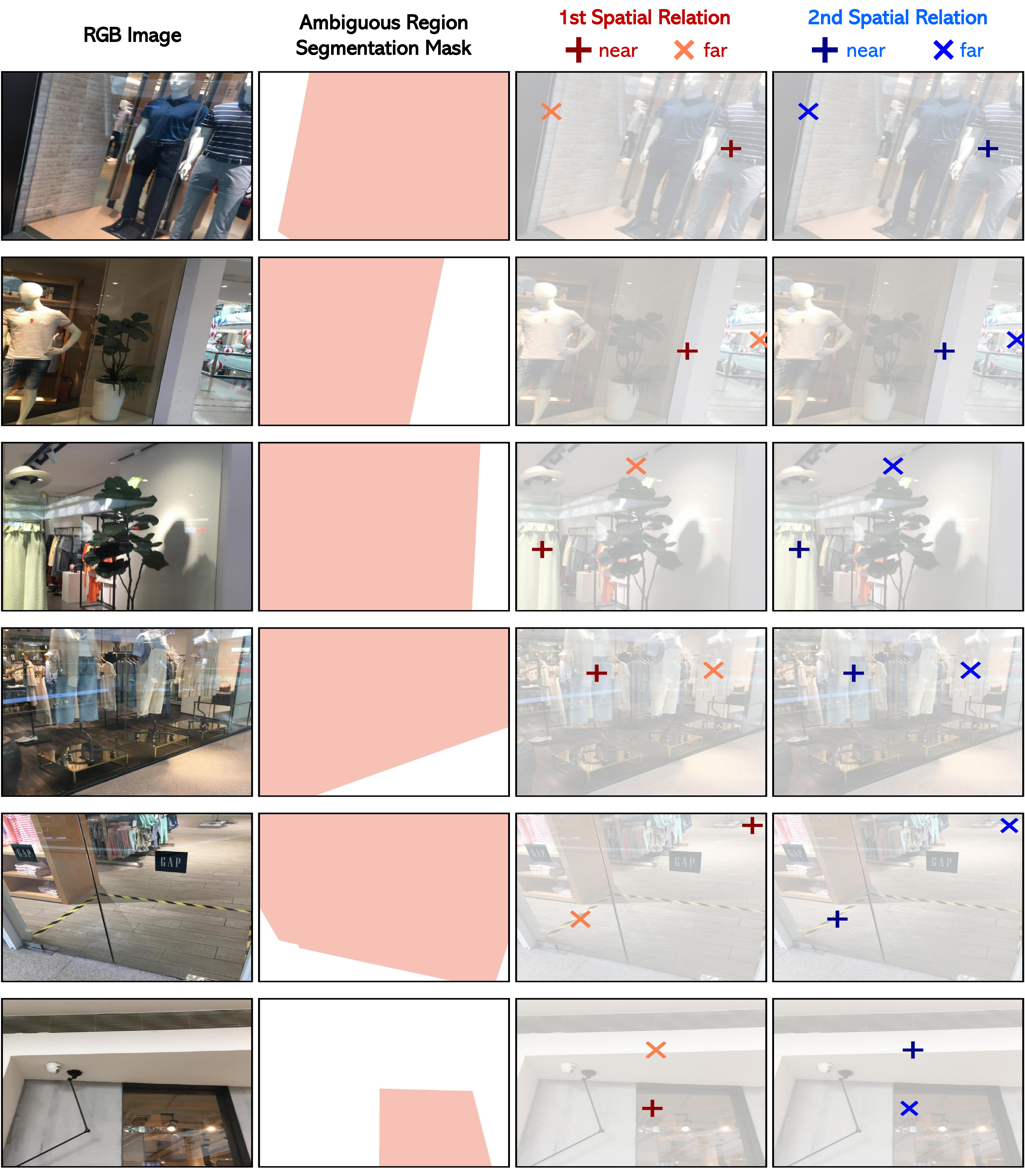}
 \caption{\textbf{MD-3k benchmark for evaluating multi-layer spatial relationships.}  Example images with annotated sparse point pairs are shown, illustrating ambiguous regions and relative depth relationships.  The first and second spatial relation columns show ground truth annotations for near/far relationships between layers, using red and blue markers, respectively.}
\label{fig:md3k_case1}\vspace{-2mm}
\end{figure*}

\vspace{1mm}
\noindent\textbf{Additional \texttt{MD-3k} benchmark samples.} Fig.~\ref{fig:md3k_case1} presents additional examples from MD-3k, our benchmark for evaluating multi-layer spatial
relationships. These examples highlight the diverse and challenging scenarios within \texttt{MD-3k}, including varying levels of depth ambiguity and transparency. By providing a broader range of scenes, we aim to assess how well models can disambiguate depth layers in multi-layered environments, particularly in real-world images that reflect the complexities and nuances of natural scenes.

\begin{figure*}[t!]
\centering \setlength{\abovecaptionskip}{0.2cm}
\includegraphics[width=\textwidth]{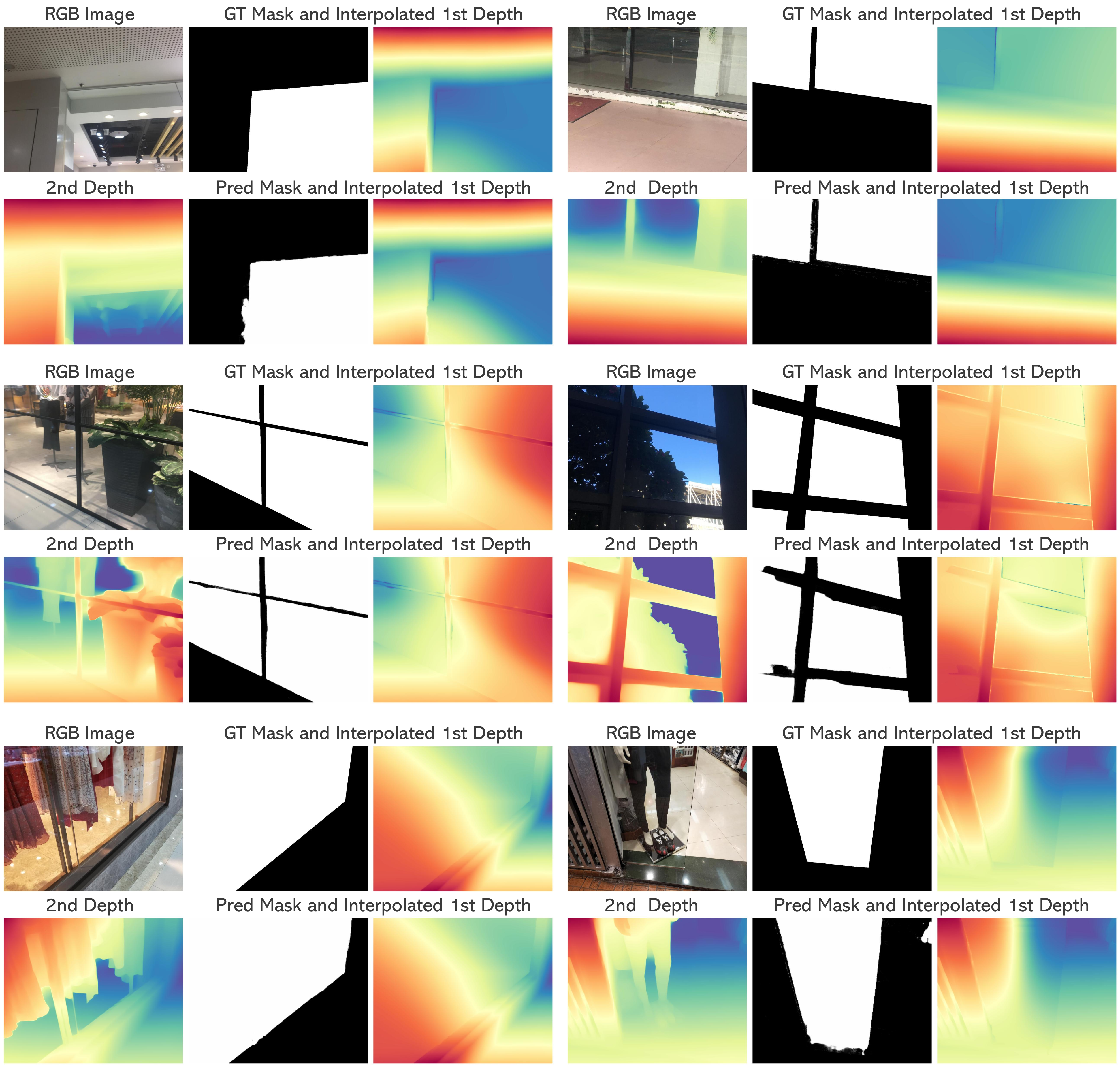}
\caption{\textbf{Multi-layer depth with extra semantic prior ({successful} {cases}).} GT-mask interpolation is shown for reference; predicted-mask interpolation shows the deployable semantic-prior variant}
\label{fig:depth_seg_good}
\end{figure*}

\begin{figure*}[t!]
\centering \setlength{\abovecaptionskip}{0.2cm}
\includegraphics[width=\textwidth]{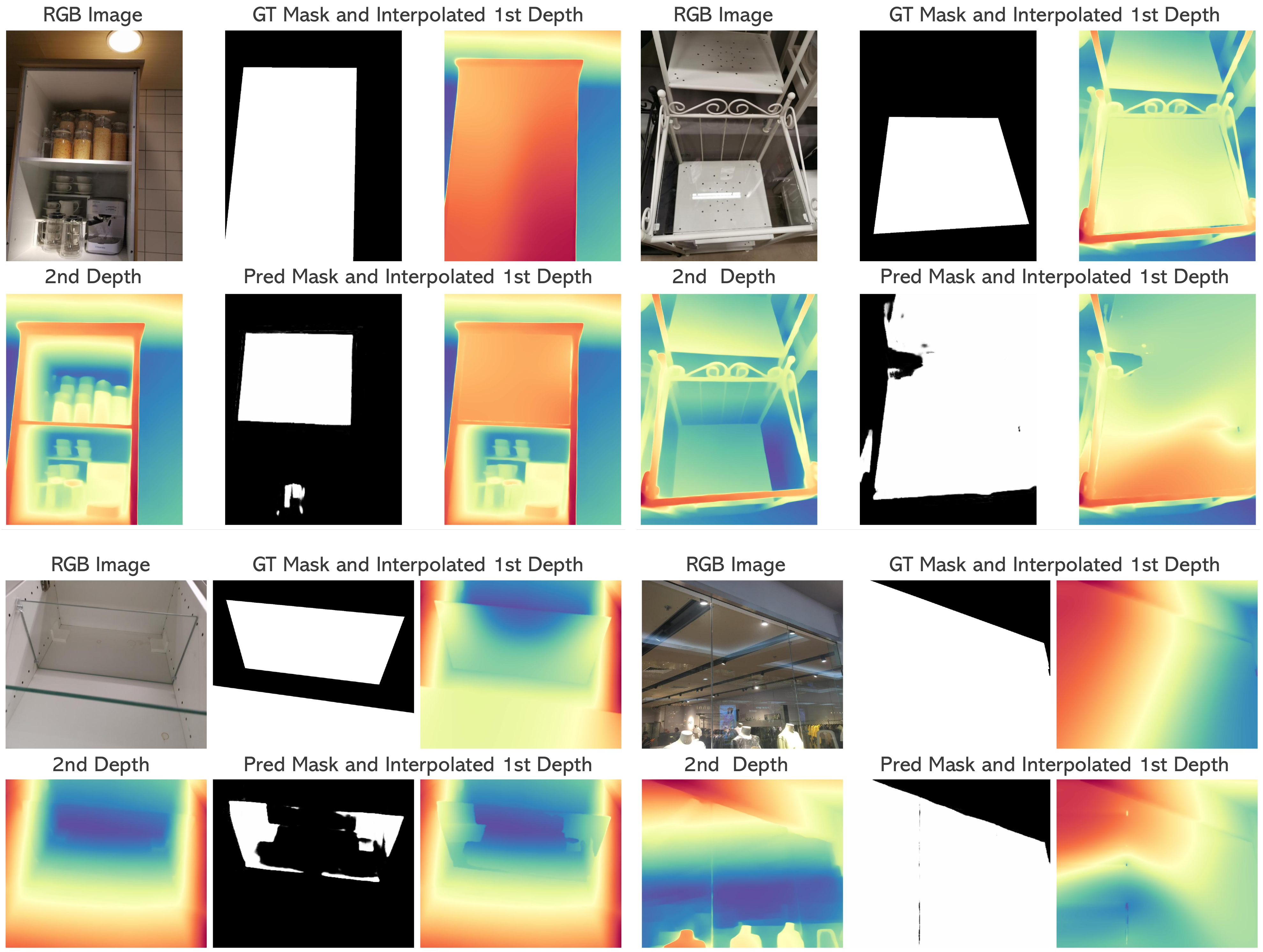}
\caption{\textbf{Multi-layer depth with extra semantic prior ({failure} {cases}).} GT-mask interpolation is shown for reference; predicted-mask interpolation shows the deployable semantic-prior variant}
\label{fig:depth_seg_bad}
\end{figure*}

\clearpage
\section{Implementation Details}

\subsection{Multi-layer Depth via Semantic Prior for Comparison}
\label{sec:segmentation}

Our semantics-guided approach to multi-layer depth estimation integrates monocular depth predictions with semantic segmentation. We use DAv1-L~\cite{depth_anything} for initial single-layer depth estimation. As noted in the main paper, DAv1-L tends to predict greater depths in ambiguous regions. Building on this bias, we estimate the nearer depth layer, typically corresponding to transparent surfaces, by interpolating depth values from the boundaries of transparent regions. This process is guided by a segmentation mask of the transparent surface and informed by DAv1-L's depth estimates outside the ambiguous regions.

Figures~\ref{fig:depth_seg_good} and~\ref{fig:depth_seg_bad} present qualitative results, showcasing both successful and failure cases. While this hybrid approach, combining DAv1-L’s depth bias with semantic segmentation, achieves higher quantitative precision for multi-layer depth estimation than our training-free LVP method, we emphasize the importance of developing foundation models that can directly handle multi-layer depth estimation, rather than relying on task-specific model combinations.

\subsection{Depth-Conditioned Image Generation}

\revdel{Our image generation pipeline uses depth maps and text prompts to guide scene synthesis. Depth maps are estimated using our proposed method, ensuring geometric accuracy, while text prompts modulate the visual attributes of the generated images.}\revadd{Our image-generation illustration uses selected RGB/LVP depth outputs and text prompts as ControlNet conditions. The depth maps  control geometry while text prompts control appearance. The examples test conditioning diversity, not metric geometric accuracy.}

We employ two distinct text prompts to control the scene's appearance:

\begin{itemize}
    \item {a bright, well-lit photograph of an interior space with natural daylight, clear windows, balanced lighting, accurate geometry and structure, photorealistic, vibrant colors, modern interior design, clean and airy space}
\end{itemize}

\begin{figure*}[t!]
\centering \setlength{\abovecaptionskip}{0.2cm}
\includegraphics[width=\textwidth]{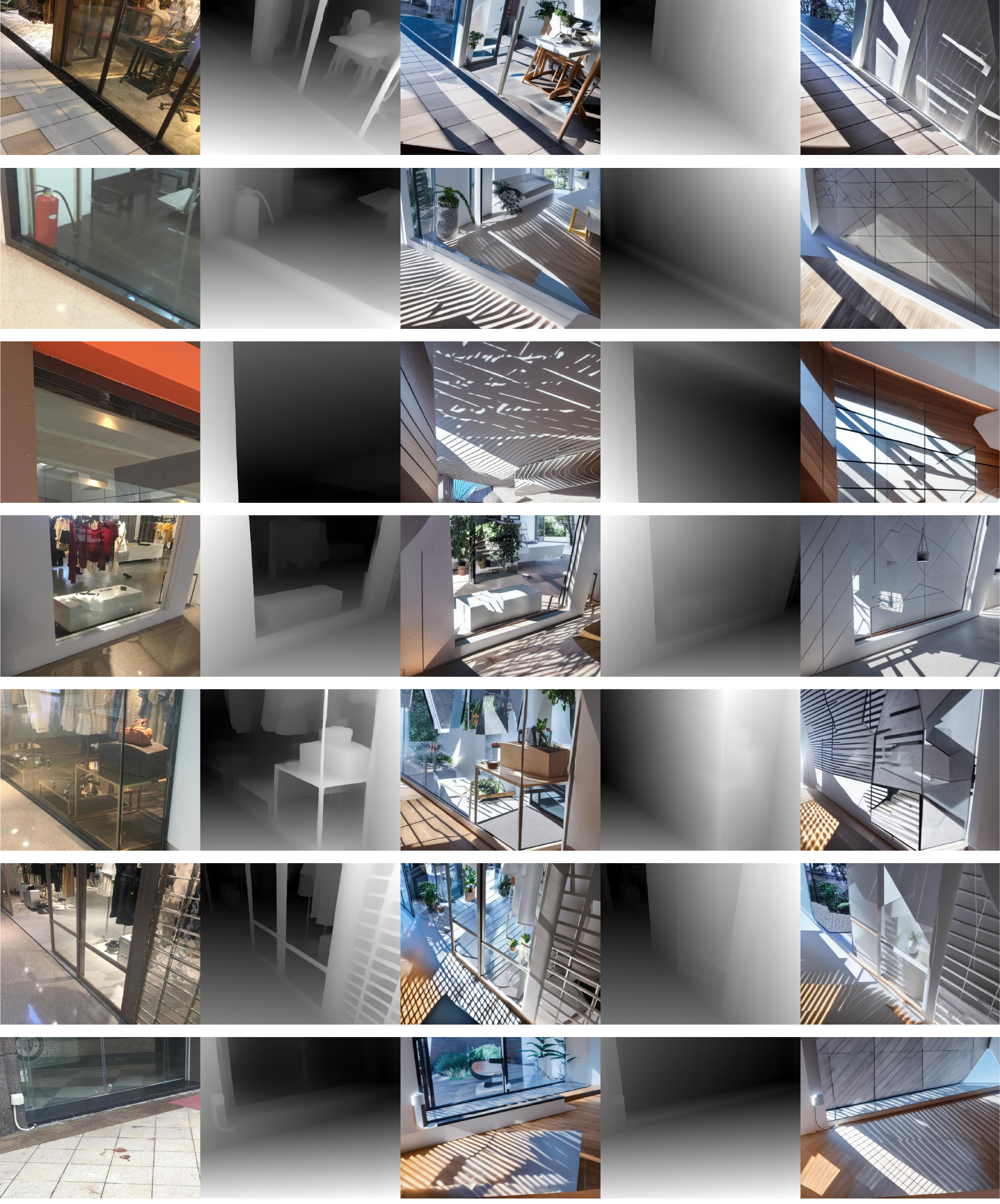}
\caption{\textbf{Multi-hypothesis spatial understanding supports flexible geometry-conditioned visual generation.} From left to right: original RGB image, depth from Laplacian Visual Prompting with its corresponding generated RGB image, and depth from the original RGB image with its generated RGB counterpart. }
\label{fig:visual_gen_appendix_1}
\end{figure*}

\revdel{The resulting RGB images, synthesized from multi-layer depth representations derived via LVP, are illustrated in Fig.~\ref{fig:visual_gen_appendix_1}.}\revadd{Images conditioned on one selected output hypothesis are shown in Fig.~\ref{fig:visual_gen_appendix_1}.}

\begin{itemize}
    \item {a bright, well-lit photograph of an interior space, accurate geometry and structure, photorealistic, modern interior design, clean and airy space}
\end{itemize}

\begin{figure*}[t!]
\centering \setlength{\abovecaptionskip}{0.2cm}
\includegraphics[width=\textwidth]{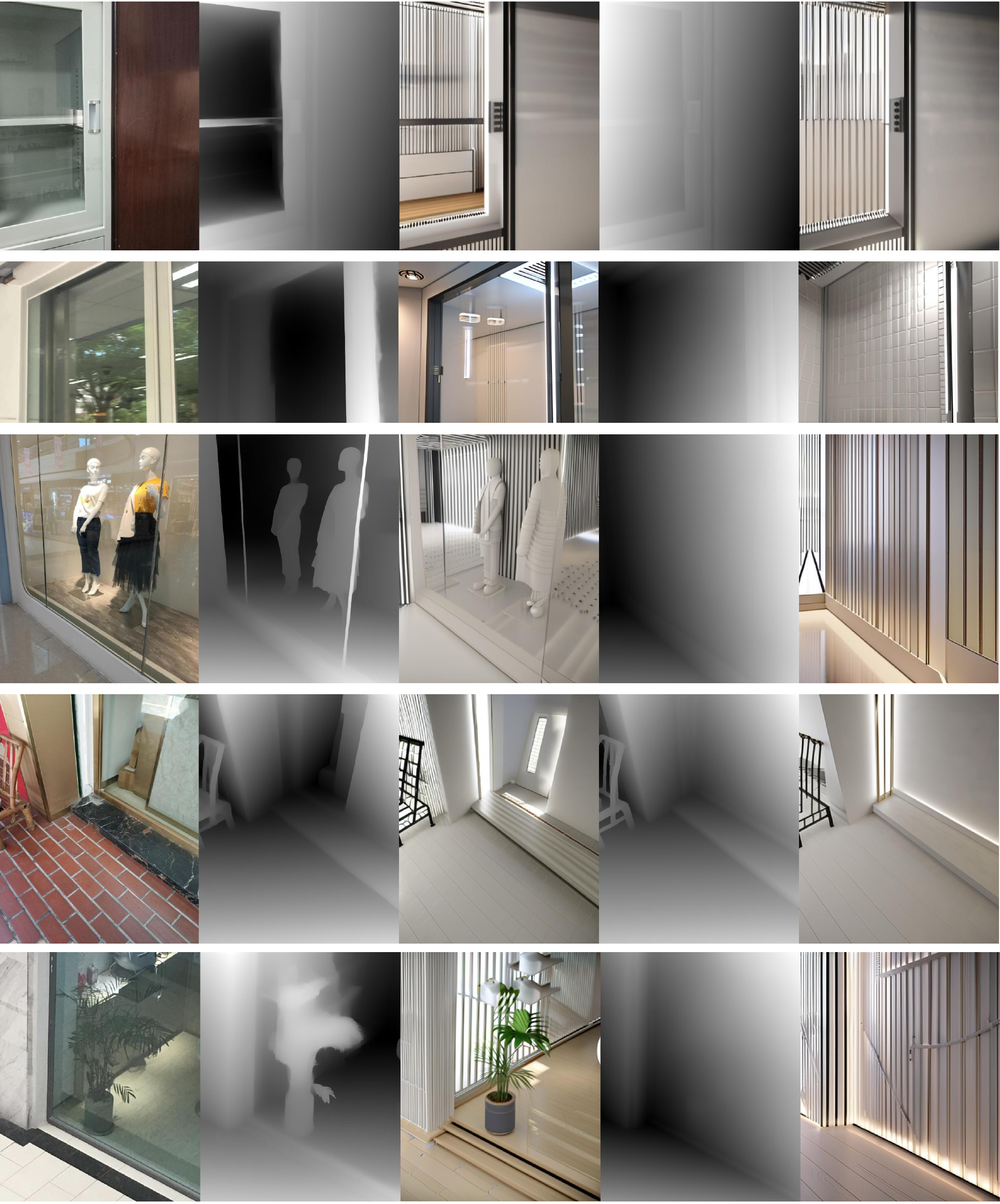}
\caption{\textbf{Multi-hypothesis spatial understanding supports flexible geometry-conditioned visual generation.} From left to right: original RGB image, depth from Laplacian Visual Prompting with its corresponding generated RGB image, and depth from the original RGB image with its generated RGB counterpart. }
\label{fig:visual_gen_appendix_2}
\end{figure*}

\revdel{The corresponding RGB images, generated using multi-layer depth representations from LVP, are presented in Fig.~\ref{fig:visual_gen_appendix_2}.}\revadd{Images conditioned on the alternative output hypothesis are shown in Fig.~\ref{fig:visual_gen_appendix_2}.}

We use
these prompts to encourage photorealistic synthesis with stable geometry and lighting, complementing depth information. More details on the diffusion models used for depth-conditioned visual generation are available in the \textit{Diffusers} library\footnote{\url{https://huggingface.co/docs/diffusers/en/index}}.

\subsection{Output Assignment and Controllability}
\label{sec:supp_output_assignment}

For ML-SRA evaluation, RGB and LVP outputs are treated as an unordered
candidate pair. Since the two outputs are not named as ``foreground'' or
``background'' by the model, we assign them to the two annotated layers using a
single benchmark-level permutation for each model. This assignment is fixed
across all images and is not selected per instance.

Table~\ref{tab:mlsra_assignment_protocol} summarizes the assignment protocol.
First, we compute the RGB model's single-output layer preference from the
per-layer SRA values. If RGB agrees more with the transparent foreground layer,
RGB is assigned to layer 1 and LVP is assigned to layer 2. If RGB agrees more
with the visible background layer, RGB is assigned to layer 2 and LVP is assigned
to layer 1. ML-SRA is then computed under this fixed assignment for every image
in the benchmark.

\begin{table}[t]
\centering
\small
\setlength{\tabcolsep}{3.5mm}
\renewcommand{\arraystretch}{1.12}
\caption{\textbf{Benchmark-level output assignment protocol for ML-SRA.}
The RGB/LVP-to-layer assignment is selected once per model from benchmark-level
RGB preference and is then fixed for all images.}
\label{tab:mlsra_assignment_protocol}
\begin{tabular}{p{0.13\linewidth}p{0.78\linewidth}}
\hline \hline
\textbf{Step} & \textbf{Operation} \\
\midrule
1 &
For a frozen model, compute per-layer RGB accuracies
$\mathrm{SRA}_{\mathrm{RGB}}(1)$ and $\mathrm{SRA}_{\mathrm{RGB}}(2)$ on
\texttt{MD-3k}. \\
\midrule
2 &
Compute the RGB depth-layer preference
$\alpha_{\mathrm{RGB}}=\mathrm{SRA}_{\mathrm{RGB}}(2)
-\mathrm{SRA}_{\mathrm{RGB}}(1)$. \\
\midrule
3a &
If $\alpha_{\mathrm{RGB}}<0$, assign RGB to the transparent foreground layer
and LVP to the visible background layer:
$\pi^\star(1)=\mathrm{RGB}$ and $\pi^\star(2)=\mathrm{LVP}$. \\
\midrule
3b &
If $\alpha_{\mathrm{RGB}}>0$, assign RGB to the visible background layer
and LVP to the transparent foreground layer:
$\pi^\star(1)=\mathrm{LVP}$ and $\pi^\star(2)=\mathrm{RGB}$. \\
\midrule
4 &
Evaluate the fixed pair using
\[
\mathrm{ML\mbox{-}SRA}
=
\frac{1}{|\mathcal P|}
\sum_{m=1}^{M}
\mathbb{I}\!\left(
\hat{\mathcal D}_{\pi^\star(1)} \equiv y_m^{(1)}
\;\wedge\;
\hat{\mathcal D}_{\pi^\star(2)} \equiv y_m^{(2)}
\right).
\]
\\
\hline \hline
\end{tabular}
\end{table}

This protocol measures candidate-pair complementarity after dataset-level label
matching. It does not implement a user-controlled layer switch and does not use a
per-image oracle. In deployment, selecting which output should be used as a
desired physical layer would require labeled calibration or an external semantic,
material, or uncertainty signal.

\clearpage

\input{sec/datasheet}

\clearpage
\section{Broader Impact}\label{sec:broader_impact}

\revdel{This work transcends monocular depth estimation, tackling the fundamental challenge of depth ambiguity that limits spatial understanding across AI. By resolving this core limitation, we pave the way for more reliable and versatile AI in complex 3D environments, impacting diverse fields reliant on robust perception.}

\revdel{Our central contribution, Laplacian Visual Prompting (LVP), introduces a training-free spectral technique. LVP empowers models to explicitly disentangle depth ambiguity and generate multi-hypothesis predictions, unlocking latent spatial knowledge. This broadly applicable spectral prompting paradigm extends beyond depth, offering a transformative tool for visual model adaptation and interpretation across tasks grappling with ambiguity and layered representations.}

\revdel{The \texttt{\texttt{MD-3k}} benchmark, the first of its kind with multi-layer spatial relationship labels, provides a critical platform for rigorous evaluation of multi-layer spatial understanding. \texttt{\texttt{MD-3k}} enables fine-grained analysis of depth disentanglement, pushing beyond single-layer metrics and driving progress in foundational spatial intelligence for computer vision, robotics, and AI.}

\revdel{Ultimately, LVP demonstrates spectral prompting as a powerful mechanism to unlock zero-shot multi-layer spatial understanding from existing models. This paradigm shift towards spectral prompting and multi-hypothesis spatial foundation models opens transformative avenues for interpretable, adaptable, and robust AI. By conquering depth ambiguity, this research delivers essential tools and insights for building foundation models capable of truly understanding real-world 3D scenes.}

\revadd{This work provides a benchmark and a lightweight diagnostic for studying transparent-scene ambiguity in frozen depth models. Potential benefits include clearer auditing of model-specific layer bias and new research on ambiguity-aware perception. The method should not be interpreted as a complete or safety-certified multi-layer estimator: its response is model-dependent, it can degrade standard depth accuracy, and ordinal benchmark success does not guarantee metric correctness.}

\section{Availability and Maintenance}
\label{sec:availablility-maintenance}

\revdel{To accelerate progress in multi-layer spatial understanding and spectral prompting, all code and datasets from this study will be publicly released.}\revadd{Our code and benchmark are publicly available at the \href{https://github.com/Xiaohao-Xu/Ambiguity-in-Space}{\texttt{Ambiguity-in-Space}} GitHub repository.} \revdel{This repository provides:}\revadd{The release is organized around:}

\begin{itemize}
    \item \textbf{Laplacian Visual Prompting (LVP) code}. \revdel{Ready-to-use implementation for spectral prompting.}\revadd{Implementation of the fixed Laplacian transform and model-inference examples.}
    \item \textbf{\texttt{MD-3k} benchmark}. \revdel{The first dataset for rigorous evaluation for depth-layer preference and multi-layer spatial relationship.}\revadd{Ordinal annotations and instructions for retrieving the underlying GDD images.}
    \item \textbf{Evaluation suite and baselines}. \revdel{Code to reproduce and extend our experimental results.}\revadd{Scripts for SRA/ML-SRA evaluation and the reported baseline outputs.}
    \item \textbf{Reproduction guide}. \revdel{Step-by-step instructions for full experiment replication.}\revadd{Documented data layout, checkpoints, and commands for the released experiments.}
\end{itemize}

\revdel{We are committed to the sustained accessibility and usability of these resources, empowering the community to build upon this foundation and drive future innovations in ambiguity-free spatial AI.}\revadd{We intend to maintain versioned releases and clear documentation, supporting reproducible study of ambiguity-aware depth estimation.} 

\section{License}\label{sec:license}
\revdel{The \texttt{\texttt{MD-3k}} benchmark and the Laplacian Visual Prompting code are released under the Apache License 2.0.}\revadd{Our annotations and evaluation code are released under the Apache License 2.0; the underlying GDD images remain subject to GDD's separate terms.} 

\section{LLM Usage Statement}
\revdel{We used LLM to proofread the text and improve sentence clarity.}\revadd{A large language model was used for language editing and sentence-level clarity; the authors remain responsible for all scientific content, analyses, and claims.}

\clearpage
\section{Public Model and Code Resources Used}
\label{sec:public-assets}

We acknowledge the following key public model and code resources.

\begin{itemize}

\item Depth-Anything-v2 \footnote{\url{https://github.com/DepthAnything/Depth-Anything-V2}} \dotfill Apache-2.0+CC-BY-NC-4.0
\item Depth-Anything \footnote{\url{https://github.com/LiheYoung/Depth-Anything}.} \dotfill Apache-2.0

\item DPT \footnote{\url{https://github.com/isl-org/DPT}.} \dotfill MIT

\item ZoeDepth \footnote{\url{https://github.com/isl-org/ZoeDepth}.} \dotfill MIT

\item Marigold \footnote{\url{https://github.com/prs-eth/Marigold}.} \dotfill Apache-2.0

\item GeoWizard \footnote{\url{https://github.com/fuxiao0719/GeoWizard}.} \dotfill CC BY 4.0

\item Diffusers \footnote{\url{https://github.com/huggingface/diffusers/tree/main}} \dotfill Apache-2.0

\item Video Depth Anything \footnote{\url{https://github.com/DepthAnything/Video-Depth-Anything}} \dotfill Apache-2.0

\end{itemize}

%% file: sec/datasheet.tex
\section{{Datasheet for {\texttt{MD-3k}} Benchmark}}
\label{sec:datasheet}

We document the necessary information about the proposed dataset and benchmark following the guidelines of Gebru \textit{et al}.~\cite{datasheet}.

\subsection{{Motivation}}\label{subsec:datasheet-motivation}

\begin{figure}[t!]
    \centering
    \setlength{\abovecaptionskip}{0.3cm}
    \includegraphics[width=\textwidth]{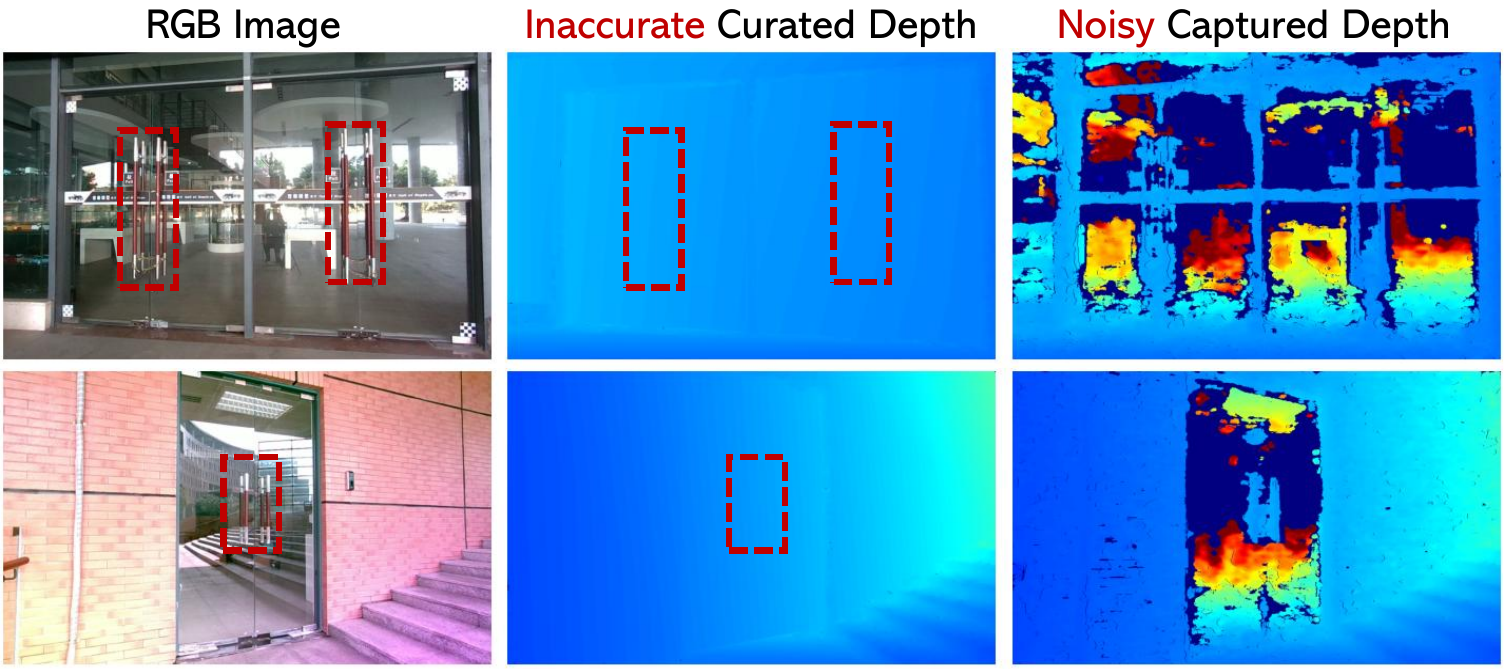}
\caption{\textbf{Limitation of real-world datasets for transparent scenes.} Noise and inaccuracies in depth data from existing datasets~\cite{liang2023monocular} for complex, ambiguous scenes. These arise from limitations in both sensor acquisition and human annotation.}
    \label{fig:challenge_existing_dataset}
\end{figure}

\begin{enumerate}[label=Q\arabic*]
\item \textbf{For what purpose was the dataset created?} Was there a specific task in mind? Was there a specific gap that needed to be filled? Please provide a description.

\begin{itemize}
\item Fig.~\ref{fig:challenge_existing_dataset} highlights the limitations of existing datasets for ambiguous transparent scenes.  They often contain noisy raw depth from sensors (due to physical limitations) and inaccurate curated depth (due to human error). These challenges motivate our creation of the \texttt{MD-3k} benchmark.

\item Our benchmark was created to evaluate multi-layer spatial perception, specifically focusing on the challenge of depth disentanglement in ambiguous 3D scenes. Existing depth datasets lack multi-layer spatial relationship labels, hindering fine-grained analysis in regions with transparency and spatial ambiguity. 
\end{itemize}

\item \textbf{Who created the dataset (e.g., which team, research group) and on behalf of which entity (e.g., company, institution, organization)?}

\begin{itemize}
\item This benchmark is established by the authors of this paper.
\end{itemize}

\item \textbf{Who funded the creation of the dataset?} If there is an associated grant, please provide the name of the grantor and the grant name and number.

\begin{itemize}
\item N/A.
\end{itemize}

\item \textbf{Any other comments?}

\begin{itemize}
\item No.
\end{itemize}

\end{enumerate}

\subsection{{Composition}}\label{subsec:datasheet-composition}

\begin{enumerate}[label=Q\arabic*]

\item \textbf{What do the instances that comprise the dataset represent (e.g., documents, photos, people, countries)?} \textit{Are there multiple types of instances (e.g., movies, users, and ratings; people and interactions between them; nodes and edges)? Please provide a description.}

\begin{itemize}
\item The instances in \texttt{MD-3k} represent high-resolution RGB images of indoor and outdoor scenes containing ambiguous regions, particularly those involving transparent objects. Each instance is associated with segmentation masks highlighting ambiguous regions and pairwise spatial relationship labels for sparse points within these regions.
\end{itemize}

\item \textbf{How many instances are there in total (of each type, if appropriate)?}

\begin{itemize}
\item The \texttt{MD-3k} benchmark comprises 3,161 high-resolution RGB images. Each image contains annotations of spatial relationships for pairs of sparse points in ambiguous regions, totaling 3,161 annotated pairs. For each pair, two layer-specific ordinal labels are provided: one for the transparent foreground and one for the
visible background behind it.
\end{itemize}

\item \textbf{Does the dataset contain all possible instances or is it a sample (not necessarily random) of instances from a larger set?} \textit{If the dataset is a sample, what is the larger set? Is the sample representative of the larger set (e.g., geographic coverage)? If so, please describe how this representativeness was validated/verified. If it is not representative of the larger set, please describe why not (e.g., to cover a more diverse range of instances, because instances were withheld or unavailable).}

\begin{itemize}
\item The dataset is a carefully selected sample from the GDD segmentation dataset~\cite{mei2020don}. The larger set is the entire GDD dataset.  The sample is not random but specifically chosen to include scenes rich in ambiguous regions, particularly those with transparent objects, to address the benchmark's focus on multi-layer spatial understanding in such challenging scenarios.
\end{itemize}

\item \textbf{What data does each instance consist of?} \textit{“Raw” data (e.g., unprocessed text or images) or features? In either case, please provide a description.}

\begin{itemize}
\item Each instance consists of:
\begin{itemize}
    \item \textbf{RGB image:} High-resolution (720p) RGB image in PNG format.
    \item \textbf{Segmentation masks:} Binary masks highlighting ambiguous regions within the RGB image, in PNG format.
    \item \textbf{Spatial relationship labels:} Pairwise spatial relationship labels for sparse points in ambiguous regions, provided in JSON format. Each pair has two layer-specific ordinal labels: one for the transparent foreground and one for the visible background, indicating near/far ordering of the two points in each layer.
\end{itemize}
This data is considered `raw' in the sense that it is primarily image data and annotations, not pre-extracted features.
\end{itemize}

\item \textbf{Is there a label or target associated with each instance?} \textit{If so, please provide a description.}

\begin{itemize}
\item Yes, the primary labels are the \textbf{pairwise spatial relationship labels}. For each annotated pair of sparse points in an ambiguous region of an RGB image, there are two labels indicating the spatial relationship (depth order) between the points in two layers. The labels are near/far ordinal relations for each layer.
\end{itemize}

\item \textbf{Is any information missing from individual instances?} \textit{If so, please provide a description, explaining why this information is missing (e.g., because it was unavailable). This does not include intentionally removed information, but might include, e.g., redacted text.}

\begin{itemize}
\item No.
\end{itemize}

\item \textbf{Are relationships between individual instances made explicit (e.g., users' movie ratings, social network links)?} \textit{If so, please describe how these relationships are made explicit.}

\begin{itemize}
\item The instances are related by their source dataset, GDD~\cite{mei2020don}. All images are selected from GDD and share the characteristics of scenes within that dataset.  Furthermore, images are implicitly related by the common theme of containing ambiguous regions and transparent objects, as this was the selection criterion.
\end{itemize}

\item \textbf{Are there recommended data splits (e.g., training, development, testing)?} \textit{If so, please provide a description of these splits, explaining the rationale behind them.}

\begin{itemize}
\item No, we do not provide predefined data splits. Users are free to define their own splits based on their specific research needs. We mainly treat it as an exploratory diagnostic benchmark.
\end{itemize}

\item \textbf{Are there any errors, sources of noise, or redundancies in the dataset?} \textit{If so, please provide a description.}

\begin{itemize}
\item We have implemented a rigorous annotation pipeline, including multi-round verification by expert annotators, to minimize errors and noise in the spatial relationship labels. However, as with any human annotation, there might be minor inconsistencies or subjective interpretations. We believe the overall quality of the annotations is high due to the careful curation process. Redundancies are not intentionally introduced.
\end{itemize}

\item \textbf{Is the dataset self-contained, or does it link to or otherwise rely on external resources (e.g., websites, tweets, other datasets)?} \textit{If it links to or relies on external resources, a) are there guarantees that they will exist, and remain constant, over time; b) are there official archival versions of the complete dataset (i.e., including the external resources as they existed at the time the dataset was created); c) are there any restrictions (e.g., licenses, fees) associated with any of the external resources that might apply to a future user? Please provide descriptions of all external resources and any restrictions associated with them, as well as links or other access points, as appropriate.}

\begin{itemize}
\item The \texttt{MD-3k} benchmark is distributed as a self-contained dataset of annotations, segmentation masks, and image lists. It relies on the images from the GDD dataset~\cite{mei2020don} as the underlying visual data. Users will need to obtain the GDD dataset separately to use \texttt{MD-3k} fully.
    \begin{itemize}
        \item a) We cannot guarantee the long-term availability of the GDD dataset. However, GDD is a publicly available dataset for research purposes.
        \item b) We do not provide archival versions of the GDD dataset. Users should refer to the original GDD dataset sources for archival information.
        \item c) Users should adhere to the licensing terms of the GDD dataset, which are separate from the \texttt{MD-3k} benchmark license. Please refer to the GDD dataset documentation for details on licenses and restrictions.
    \end{itemize}
\end{itemize}

\item \textbf{Does the dataset contain data that might be considered confidential (e.g., data that is protected by legal privilege or by doctor–patient confidentiality, data that includes the content of individuals’ non-public communications)?} \textit{If so, please provide a description.}

\begin{itemize}
\item No, the \texttt{MD-3k} benchmark utilizes images from the publicly available GDD dataset, which does not contain confidential information.
\end{itemize}

\item \textbf{Does the dataset contain data that, if viewed directly, might be offensive, insulting, threatening, or might otherwise cause anxiety?} \textit{If so, please describe why.}

\begin{itemize}
\item No. The images in the \texttt{MD-3k} benchmark depict common indoor and outdoor scenes and do not contain offensive, insulting, threatening, or anxiety-inducing content to the best of our knowledge.
\end{itemize}

\item \textbf{Does the dataset relate to people?} \textit{If not, you may skip the remaining questions in this section.}

\begin{itemize}
\item No. 
\end{itemize}

\item \textbf{Does the dataset identify any subpopulations (e.g., by age, gender)?}

\begin{itemize}
\item N/A.
\end{itemize}

\item \textbf{Is it possible to identify individuals (i.e., one or more natural persons), either directly or indirectly (i.e., in combination with other data) from the dataset?} \textit{If so, please describe how.}

\begin{itemize}
\item N/A.
\end{itemize}

\item \textbf{Does the dataset contain data that might be considered sensitive in any way (e.g., data that reveals racial or ethnic origins, sexual orientations, religious beliefs, political opinions or union memberships, or locations; financial or health data; biometric or genetic data; forms of government identification, such as social security numbers; criminal history)?} \textit{If so, please provide a description.}

\begin{itemize}
\item No.
\end{itemize}

\item \textbf{Any other comments?}

\begin{itemize}
\item No.
\end{itemize}

\end{enumerate}
\subsection{{Collection Process}}\label{subsec:datasheet-collection}

\begin{enumerate}[label=Q\arabic*]
\item \textbf{How was the data associated with each instance acquired?} \textit{Was the data directly observable (e.g., raw text, movie ratings), reported by subjects (e.g., survey responses), or indirectly inferred/derived from other data (e.g., part-of-speech tags, model-based guesses for age or language)? If data was reported by subjects or indirectly inferred/derived from other data, was the data validated/verified? If so, please describe how.}

\begin{itemize}
\item The RGB images were directly observable, sourced from the GDD segmentation dataset~\cite{mei2020don}. The segmentation masks and spatial relationship labels were indirectly derived through expert human annotation. Expert annotators manually identified ambiguous regions and provided pairwise spatial relationship labels. The annotations were validated through a multi-round verification process involving multiple annotators to ensure consistency and accuracy.
\end{itemize}

\item \textbf{What mechanisms or procedures were used to collect the data (e.g., hardware apparatus or sensor, manual human curation, software program, software API)?} \textit{How were these mechanisms or procedures validated?}

\begin{itemize}
\item The data collection process primarily involved \textbf{manual human curation}. Expert annotators used in-house annotation tools to:
    \begin{itemize}
        \item Visually inspect RGB images from the GDD dataset.
        \item Identify and segment ambiguous regions, particularly those involving transparent objects, creating segmentation masks.
        \item Select sparse point pairs within these ambiguous regions.
        \item Determine and assign near/far ordinal relations for each sparse point pair in both the transparent-foreground and visible-background layers.
    \end{itemize}
The annotation procedure was validated through a multi-round verification process. Different annotators reviewed and cross-validated annotations to resolve discrepancies and ensure consistency and accuracy of the labels.
\end{itemize}

\item \textbf{If the dataset is a sample from a larger set, what was the sampling strategy (e.g., deterministic, probabilistic with specific sampling probabilities)?}

\begin{itemize}
\item The sampling strategy was \textbf{deterministic and targeted}. Images were selected from the GDD dataset based on a specific criterion: the presence of ambiguous regions, especially those featuring transparent objects. 
\end{itemize}

\item \textbf{Who was involved in the data collection process (e.g., students, crowdworkers, contractors) and how were they compensated (e.g., how much were crowdworkers paid)?}

\begin{itemize}
\item The annotators were authors with expertise in computer vision and image annotation; no crowdworker compensation was involved.
\end{itemize}

\item \textbf{Over what timeframe was the data collected? Does this timeframe match the creation timeframe of the data associated with the instances (e.g., recent crawl of old news articles)?} \textit{If not, please describe the timeframe in which the data associated with the instances was created.}

\begin{itemize}
\item The data annotation and collection process took place between Dec 2025 and Jan 2026. This timeframe represents the creation timeframe of the spatial relationship labels associated with the images from the source GDD dataset. The GDD dataset itself was created prior to this timeframe.
\end{itemize}

\item \textbf{Were any ethical review processes conducted (e.g., by an institutional review board)?} \textit{If so, please provide a description of these review processes, including the outcomes, as well as a link or other access point to any supporting documentation.}

\begin{itemize}
\item Ethical review processes were not formally conducted by an institutional review board specifically for the creation of \texttt{MD-3k}. However, the benchmark utilizes publicly available images from the GDD dataset, which is intended for research purposes. 
\end{itemize}

\item \textbf{Does the dataset relate to people?} \textit{If not, you may skip the remaining questions in this section.}

\begin{itemize}
\item No.
\end{itemize}

\item \textbf{Did you collect the data from the individuals in question directly, or obtain it via third parties or other sources (e.g., websites)?}

\begin{itemize}
\item N/A.
\end{itemize}

\item \textbf{Were the individuals in question notified about the data collection?} \textit{If so, please describe (or show with screenshots or other information) how notice was provided, and provide a link or other access point to, or otherwise reproduce, the exact language of the notification itself.}

\begin{itemize}
\item N/A.
\end{itemize}

\item \textbf{Did the individuals in question consent to the collection and use of their data?} \textit{If so, please describe (or show with screenshots or other information) how consent was requested and provided, and provide a link or other access point to, or otherwise reproduce, the exact language to which the individuals consented.}

\begin{itemize}
\item N/A.
\end{itemize}

\item \textbf{If consent was obtained, were the consenting individuals provided with a mechanism to revoke their consent in the future or for certain uses?} \textit{If so, please provide a description, as well as a link or other access point to the mechanism (if appropriate).}

\begin{itemize}
\item N/A.
\end{itemize}

\item \textbf{Has an analysis of the potential impact of the dataset and its use on data subjects (e.g., a data protection impact analysis) been conducted?} \textit{If so, please provide a description of this analysis, including the outcomes, as well as a link or other access point to any supporting documentation.}

\begin{itemize}
\item N/A.
\end{itemize}

\item \textbf{Any other comments?}

\begin{itemize}
\item No.
\end{itemize}

\end{enumerate}

\subsection{{Preprocessing, Cleaning, and/or Labeling}}\label{subsec:datasheet-preprocess}


\begin{enumerate}[label=Q\arabic*]

\item \textbf{Was any preprocessing/cleaning/labeling of the data done (e.g., discretization or bucketing, tokenization, part-of-speech tagging, SIFT feature extraction, removal of instances, processing of missing values)?} \textit{If so, please provide a description. If not, you may skip the remainder of the questions in this section.}

\begin{itemize}
\item Yes, \textbf{labeling} was performed. Expert annotators manually labeled spatial relationships for pairs of sparse points in ambiguous regions. This labeling process is the core contribution of the \texttt{MD-3k} benchmark. No other preprocessing or cleaning of the RGB images from the GDD dataset was performed.
\end{itemize}

\item \textbf{Was the “raw” data saved in addition to the preprocessed, cleaned, or labeled data (e.g., to support unanticipated future uses)?} \textit{If so, please provide a link or other access point to the “raw” data.}

\begin{itemize}
\item N/A. The `raw' data in this context would be the original RGB images from the GDD dataset. We are distributing the segmentation masks and spatial relationship labels, which are the `labeled' data. The `raw' RGB images are available from the original GDD dataset~\cite{mei2020don}.
\end{itemize}

\item \textbf{Is the software used to preprocess/clean/label the instances available?} \textit{If so, please provide a link or other access point.}

\begin{itemize}
\item The annotation tools used for labeling are not publicly released at this time. However, we  provide detailed descriptions of the annotation process and data format to facilitate the expansion of the benchmark.
\end{itemize}

\item \textbf{Any other comments?}

\begin{itemize}
\item No.
\end{itemize}

\end{enumerate}

\subsection{{Uses}}\label{subsec:datasheet-uses}
%

\begin{enumerate}[label=Q\arabic*]

\item \textbf{Has the dataset been used for any tasks already?} \textit{If so, please provide a description.}

\begin{itemize}
\item Yes. In this paper, \texttt{MD-3k} is used to evaluate depth-layer preference and RGB/LVP candidate-pair complementarity in frozen monocular depth models. We are not aware of external uses yet.
\end{itemize}

\item \textbf{Is there a repository that links to any or all papers or systems that use the dataset?} \textit{If so, please provide a link or other access point.}

\begin{itemize}
\item We will maintain a repository that links to papers and systems that utilize the \texttt{MD-3k} benchmark as they become available.
\end{itemize}

\item \textbf{What (other) tasks could the dataset be used for?}

\begin{itemize}
\item The primary intended use of \texttt{MD-3k} is for evaluating models for \textbf{multi-layer spatial understanding} and \textbf{depth disentanglement} in ambiguous scenes. Specifically, it can be used to:
    \begin{itemize}
        \item Evaluate the performance of depth estimation models in regions with transparency and complex spatial arrangements.
        \item Benchmark algorithms designed for understanding layered scene representations.
        \item Analyze the ability of models to reason about relative depth ordering in multi-layer contexts.
        \item Develop and test novel approaches for handling spatial ambiguity in 3D scene understanding.
    \end{itemize}
\end{itemize}

\item \textbf{Is there anything about the composition of the dataset or the way it was collected and preprocessed/cleaned/labeled that might impact future uses?} \textit{For example, is there anything that a future user might need to know to avoid uses that could result in unfair treatment of individuals or groups (e.g., stereotyping, quality of service issues) or other undesirable harms (e.g., financial harms, legal risks)? If so, please provide a description. Is there anything a future user could do to mitigate these undesirable harms?}

\begin{itemize}
\item The \texttt{MD-3k} benchmark is focused on ambiguous scenes, particularly those with transparent objects. Users should be aware that the dataset is specifically designed to challenge models in these scenarios.  It might not be representative of general scenes without ambiguity.  Future users should consider this focus when applying the benchmark and interpreting results.  As the dataset does not relate to people or sensitive attributes, the risk of unfair treatment or other harms is considered low. However, responsible and ethical use of the benchmark is always encouraged.
\end{itemize}

\item \textbf{Are there tasks for which the dataset should not be used?} \textit{If so, please provide a description.}

\begin{itemize}
\item We are not aware of any specific tasks for which \texttt{MD-3k} should not be used. However, its primary focus is on multi-layer spatial understanding in ambiguous regions. Using it for tasks completely unrelated to spatial reasoning or depth perception might not be appropriate.

\end{itemize}
\item \textbf{Any other comments?}

\begin{itemize}
\item No.
\end{itemize}

\end{enumerate}

\subsection{{Distribution and License}}\label{subsec:datasheet-distribution}

\begin{enumerate}[label=Q\arabic*]

\item \textbf{Will the dataset be distributed to third parties outside of the entity (e.g.,
 company, institution, organization) on behalf of which the dataset was created?} \textit{If so, please provide a description.}

\begin{itemize}
\item Yes, the \texttt{MD-3k} benchmark will be publicly available for research purposes.
\end{itemize}

\item \textbf{How will the dataset be distributed (e.g., tarball on website, API, GitHub)?} \textit{Does the dataset have a digital object identifier (DOI)?}

\begin{itemize}
\item The benchmark is distributed through GitHub; no DOI has been assigned at this time.
\end{itemize}

\item \textbf{When will the dataset be distributed?}

\begin{itemize}
\item Our dataset is already released publicly.
\end{itemize}

\item \textbf{Will the dataset be distributed under a copyright or other intellectual property (IP) license, and/or under applicable terms of use (ToU)?} \textit{If so, please describe this license and/or ToU, and provide a link or other access point to, or otherwise reproduce, any relevant licensing terms or ToU, as well as any fees associated with these restrictions.}

\begin{itemize}
\item The \texttt{MD-3k} benchmark, including annotations and code, is released under the \textbf{Apache-2.0 license}. This is an open-source license that allows for free use, modification, and distribution for research and commercial purposes, with proper attribution. 
\end{itemize}

\item \textbf{Have any third parties imposed IP-based or other restrictions on the data associated with the instances?} \textit{If so, please describe these restrictions, and provide a link or other access point to, or otherwise reproduce, any relevant licensing terms, as well as any fees associated with these restrictions.}

\begin{itemize}
\item The \texttt{MD-3k} benchmark relies on RGB images from the GDD dataset~\cite{mei2020don}. Users of \texttt{MD-3k} should also comply with the licensing terms of the GDD dataset, which are separate from the Apache-2.0 license of our benchmark. We recommend users refer to the GDD dataset documentation for details on their specific licensing terms and any potential restrictions. We are not aware of any IP-based or other restrictions imposed by third parties directly on our annotations and benchmark data, other than the underlying GDD images.
\end{itemize}

\item \textbf{Do any export controls or other regulatory restrictions apply to the dataset or to individual instances?} \textit{If so, please describe these restrictions, and provide a link or other access point to, or otherwise reproduce, any supporting documentation.}

\begin{itemize}
\item N/A.
\end{itemize}

\item \textbf{Any other comments?}

\begin{itemize}
\item No.
\end{itemize}

\end{enumerate}

\subsection{{Maintenance}}\label{subsec:datasheet-mainteanance}


\begin{enumerate}[label=Q\arabic*]

\item \textbf{Who will be supporting/hosting/maintaining the dataset?}

\begin{itemize}
\item The authors will be responsible for supporting, hosting, and maintaining the \texttt{MD-3k} benchmark.
\end{itemize}

\item \textbf{How can the owner/curator/manager of the dataset be contacted (e.g., email address)?}

\begin{itemize}
\item Users can contact the maintainers through the GitHub issue tracker and the contact email listed in the repository.
\end{itemize}

\item \textbf{Is there an erratum?} \textit{If so, please provide a link or other access point.}

\begin{itemize}
\item No.
\end{itemize}

\item \textbf{Will the dataset be updated (e.g., to correct labeling errors, add new instances, delete instances)?} \textit{If so, please describe how often, by whom, and how updates will be communicated to users (e.g., mailing list, GitHub)?}

\begin{itemize}
\item Yes, we may update the \texttt{MD-3k} benchmark and will highlight that in the dataset repo if so. 
\end{itemize}

\item \textbf{If the dataset relates to people, are there applicable limits on the retention of the data associated with the instances (e.g., were individuals in question told that their data would be retained for a fixed period of time and then deleted)?} \textit{If so, please describe these limits and explain how they will be enforced.}

\begin{itemize}
\item N/A.
\end{itemize}

\item \textbf{Will older versions of the dataset continue to be supported?} \textit{If so, please describe how. If not, please describe how its obsolescence will be communicated to users.}

\begin{itemize}
\item We intend to host and maintain all versions of the \texttt{MD-3k} benchmark in our GitHub repository. This will allow users to access and utilize specific versions of the benchmark for reproducibility and comparison purposes. If a version becomes obsolete, it will be clearly marked as such in the repository, but will remain accessible.
\end{itemize}

\item \textbf{If others want to extend/augment/build on/contribute to the dataset, is there a mechanism for them to do so?} \textit{If so, please provide a description. Will these contributions be validated/verified? If so, please describe how. If not, why not? Is there a process for communicating/distributing these contributions to other users? If so, please provide a description.}

\begin{itemize}
\item We welcome contributions from the community to extend, augment, or build upon the \texttt{MD-3k} benchmark. Users can contribute by:
    \begin{itemize}
        \item Reporting issues or suggesting improvements via the issue tracker in the benchmark repository.
        \item Submitting pull requests with code contributions (e.g., evaluation scripts, new baselines).
        \item Proposing new annotations or extensions to the dataset by contacting the authors through the contact method provided in the repository.
    \end{itemize}
\end{itemize}

\item \textbf{Any other comments?}

\begin{itemize}
\item No.
\end{itemize}

\end{enumerate}

%% file: main.bib
@String(CVPR= {IEEE Conf. Comput. Vis. Pattern Recog.})

@String(ICCV= {Int. Conf. Comput. Vis.})

@String(ECCV= {Eur. Conf. Comput. Vis.})

@String(ICASSP=	{ICASSP})

@String(ICLR = {Int. Conf. Learn. Represent.})

@String(AAAI = {AAAI})

@String(CVPR  = {CVPR})

@String(ICCV  = {ICCV})

@String(ECCV  = {ECCV})

@String(ICLR  = {ICLR})

@inproceedings{mei2020don,
  title={Don't hit me! glass detection in real-world scenes},
  author={Mei, Haiyang and Yang, Xin and Wang, Yang and Liu, Yuanyuan and He, Shengfeng and Zhang, Qiang and Wei, Xiaopeng and Lau, Rynson WH},
  booktitle={CVPR},
  pages={3687--3696},
  year={2020}
}

@article{yang2024depth,
  title={Depth Anything V2},
  author={Yang, Lihe and Kang, Bingyi and Huang, Zilong and Zhao, Zhen and Xu, Xiaogang and Feng, Jiashi and Zhao, Hengshuang},
  journal={NeurIPS},
  year={2024}
}

@inproceedings{depth_anything,
  title={Depth anything: Unleashing the power of large-scale unlabeled data},
  author={Yang, Lihe and Kang, Bingyi and Huang, Zilong and Xu, Xiaogang and Feng, Jiashi and Zhao, Hengshuang},
  booktitle={CVPR},
  year={2024}
}

@inproceedings{marigold,
  title={Repurposing diffusion-based image generators for monocular depth estimation},
  author={Ke, Bingxin and Obukhov, Anton and Huang, Shengyu and Metzger, Nando and Daudt, Rodrigo Caye and Schindler, Konrad},
  booktitle={CVPR},
  year={2024}
}

@article{liang2023monocular,
  title={Monocular depth estimation for glass walls with context: a new dataset and method},
  author={Liang, Yuan and Deng, Bailin and Liu, Wenxi and Qin, Jing and He, Shengfeng},
  journal={TPAMI},
  year={2023},
  publisher={IEEE}
}

@inproceedings{controlnet,
  title={Adding conditional control to text-to-image diffusion models},
  author={Zhang, Lvmin and Rao, Anyi and Agrawala, Maneesh},
  booktitle={ICCV},
  year={2023}
}

@article{Bahng_2022_NeurIPS,
         title={Exploring Visual Prompts for Adapting Large-Scale Models}, 
         author={Hyojin Bahng and Ali Jahanian and Swami Sankaranarayanan and Phillip Isola},
         journal={arXiv preprint arXiv:2203.17274},
         year={2022}
}

@article{midas,
  title={Towards robust monocular depth estimation: Mixing datasets for zero-shot cross-dataset transfer},
  author={Ranftl, Ren{\'e} and Lasinger, Katrin and Hafner, David and Schindler, Konrad and Koltun, Vladlen},
  journal={TPAMI},
  year={2022},
}

@article{midasv31,
  title={MiDaS v3. 1--A Model Zoo for Robust Monocular Relative Depth Estimation},
  author={Birkl, Reiner and Wofk, Diana and M{\"u}ller, Matthias},
  journal={arXiv:2307.14460},
  year={2023}
}

@inproceedings{metric3d,
  title={Metric3D: Towards zero-shot metric 3d prediction from a single image},
  author={Yin, Wei and Zhang, Chi and Chen, Hao and Cai, Zhipeng and Yu, Gang and Wang, Kaixuan and Chen, Xiaozhi and Shen, Chunhua},
  booktitle={ICCV},
  year={2023}
}

@article{geowizard,
  title={GeoWizard: Unleashing the Diffusion Priors for 3D Geometry Estimation from a Single Image},
  author={Fu, Xiao and Yin, Wei and Hu, Mu and Wang, Kaixuan and Ma, Yuexin and Tan, Ping and Shen, Shaojie and Lin, Dahua and Long, Xiaoxiao},
  journal={ECCV},
  year={2024}
}

@article{zoedepth,
  title={Zoedepth: Zero-shot transfer by combining relative and metric depth},
  author={Bhat, Shariq Farooq and Birkl, Reiner and Wofk, Diana and Wonka, Peter and M{\"u}ller, Matthias},
  journal={arXiv:2302.12288},
  year={2023}
}

@inproceedings{sd,
  title={High-resolution image synthesis with latent diffusion models},
  author={Rombach, Robin and Blattmann, Andreas and Lorenz, Dominik and Esser, Patrick and Ommer, Bj{\"o}rn},
  booktitle={CVPR},
  year={2022}
}

@article{kitti,
  title={Vision meets robotics: The kitti dataset},
  author={Geiger, Andreas and Lenz, Philip and Stiller, Christoph and Urtasun, Raquel},
  journal={IJRR},
  year={2013}
}

@inproceedings{bai2024sequential,
  title={Sequential modeling enables scalable learning for large vision models},
  author={Bai, Yutong and Geng, Xinyang and Mangalam, Karttikeya and Bar, Amir and Yuille, Alan L and Darrell, Trevor and Malik, Jitendra and Efros, Alexei A},
  booktitle={CVPR},
  pages={22861--22872},
  year={2024}
}

@inproceedings{nyud,
  title={Indoor segmentation and support inference from rgbd images},
  author={Silberman, Nathan and Hoiem, Derek and Kohli, Pushmeet and Fergus, Rob},
  booktitle={ECCV},
  year={2012}
}

@inproceedings{hypersim,
  title={Hypersim: A photorealistic synthetic dataset for holistic indoor scene understanding},
  author={Roberts, Mike and Ramapuram, Jason and Ranjan, Anurag and Kumar, Atulit and Bautista, Miguel Angel and Paczan, Nathan and Webb, Russ and Susskind, Joshua M},
  booktitle={ICCV},
  year={2021}
}

@inproceedings{diw,
  title={Single-image depth perception in the wild},
  author={Chen, Weifeng and Fu, Zhao and Yang, Dawei and Deng, Jia},
  booktitle={NeurIPS},
  year={2016}
}

@inproceedings{eigen2014depth,
  title={Depth map prediction from a single image using a multi-scale deep network},
  author={Eigen, David and Puhrsch, Christian and Fergus, Rob},
  booktitle={NeurIPS},
  year={2014}
}

@inproceedings{adabins,
  title={Adabins: Depth estimation using adaptive bins},
  author={Bhat, Shariq Farooq and Alhashim, Ibraheem and Wonka, Peter},
  booktitle={CVPR},
  year={2021}
}

@inproceedings{dpt,
  title={Vision transformers for dense prediction},
  author={Ranftl, Ren{\'e} and Bochkovskiy, Alexey and Koltun, Vladlen},
  booktitle={ICCV},
  year={2021}
}

@inproceedings{eth3d,
  title={A multi-view stereo benchmark with high-resolution images and multi-camera videos},
  author={Schops, Thomas and Schonberger, Johannes L and Galliani, Silvano and Sattler, Torsten and Schindler, Konrad and Pollefeys, Marc and Geiger, Andreas},
  booktitle={CVPR},
  year={2017}
}

@STRING{ijrr    = "Int. J. Robot. Research" }

@article{datasheet,
author = {Gebru, Timnit and Morgenstern, Jamie and Vecchione, Briana and Vaughan, Jennifer Wortman and Wallach, Hanna and Daum\'{e} III, Hal and Crawford, Kate},
title = {Datasheets for datasets},
year = {2021},
issue_date = {December 2021},
publisher = {Association for Computing Machinery},
address = {New York, NY, USA},
volume = {64},
number = {12},
journal = {Commun. ACM},
month = {nov},
pages = {86–92},
numpages = {7}
}

@inproceedings{neekhara2022cross,
  title={Cross-modal Adversarial Reprogramming},
  author={Neekhara, Paarth and Hussain, Shehzeen and Du, Jinglong and Dubnov, Shlomo and Koushanfar, Farinaz and McAuley, Julian},
  booktitle={WACV},
  pages={2427--2435},
  year={2022}
}

@article{brown2020language,
  title={Language models are few-shot learners},
  author={Brown, Tom and Mann, Benjamin and Ryder, Nick and Subbiah, Melanie and Kaplan, Jared D and Dhariwal, Prafulla and Neelakantan, Arvind and Shyam, Pranav and Sastry, Girish and Askell, Amanda and others},
  journal={NeurIPS},
  volume={33},
  pages={1877--1901},
  year={2020}
}

@inproceedings{chen2022visual,
  title={Visual prompting for adversarial robustness},
  author={Chen, Aochuan and Lorenz, Peter and Yao, Yuguang and Chen, Pin-Yu and Liu, Sijia},
  booktitle={ICASSP},
  pages={1--5},
  year={2023},
  organization={IEEE}
}

@article{tsai2020transfer,
  title={Transfer Learning without Knowing: Reprogramming Black-box Machine Learning Models with Scarce Data and Limited Resources},
  author={Tsai, Yun-Yun and Chen, Pin-Yu and Ho, Tsung-Yi},
  journal={ICML},
  year={2020}
}

@inproceedings{chen2021adversarial,
  title={Adversarial Reprogramming of Pretrained Neural Networks for Fraud Detection},
  author={Chen, Lingwei and Fan, Yujie and Ye, Yanfang},
  booktitle={CIKM},
  pages={2935--2939},
  year={2021}
}

@inproceedings{singha2023ad,
 title = {{AD-CLIP}: Adapting Domains in Prompt Space Using {CLIP}},
 author = {Singha, Mainak and Pal, Harsh and Jha, Ankit and Banerjee, Biplab},
 booktitle = {ICCV Workshops},
 pages = {4355--4364},
 year = {2023}
}

@inproceedings{depthfm,
 title = {{DepthFM}: Fast Monocular Depth Estimation with Flow Matching},
 author = {Gui, Ming and Schusterbauer, Johannes and Prestel, Ulrich and Ma, Pingchuan and Kotovenko, Dmytro and Grebenkova, Olga and Baumann, Stefan Andreas
and Hu, Vincent Tao and Ommer, Bj{\"o}rn},
 booktitle = {AAAI},
 year = {2025}
}

@inproceedings{wasim2023vita,
  title={Vita-clip: Video and text adaptive clip via multimodal prompting},
  author={Wasim, Syed Talal and Naseer, Muzammal and Khan, Salman and Khan, Fahad Shahbaz and Shah, Mubarak},
  booktitle={CVPR},
  pages={23034--23044},
  year={2023}
}

@inproceedings{wang2024vilt,
  title={ViLT-CLIP: Video and language tuning CLIP with multimodal prompt learning and scenario-guided optimization},
  author={Wang, Hao and Liu, Fang and Jiao, Licheng and Wang, Jiahao and Hao, Zehua and Li, Shuo and Li, Lingling and Chen, Puhua and Liu, Xu},
  booktitle={AAAI},
  year={2024}
}

@article{khattak2022maple,
  title={MaPLe: Multi-modal Prompt Learning},
  author={Khattak, Muhammad Uzair and Rasheed, Hanoona and Maaz, Muhammad and Khan, Salman and Khan, Fahad Shahbaz},
  journal={CVPR},
  year={2023}
}

@inproceedings{
bochkovskii2024depth,
title={Depth Pro: Sharp Monocular Metric Depth in Less Than a Second},
author={Alexey Bochkovskiy and Ama{\"e}l Delaunoy and Hugo Germain and Marcel Santos and Yichao Zhou and Stephan Richter and Vladlen Koltun},
booktitle={ICLR},
year={2025}
}

@inproceedings{piccinelli2025unik3d,
  title={UniK3D: Universal Camera Monocular 3D Estimation},
  author={Piccinelli, Luigi and Sakaridis, Christos and Segu, Mattia and Yang, Yung-Hsu and Li, Siyuan and Abbeloos, Wim and Van Gool, Luc},
  booktitle={CVPR},
  pages={1028--1039},
  year={2025}
}

@article{piccinelli2025unidepthv2,
  title={Unidepthv2: Universal monocular metric depth estimation made simpler},
  author={Piccinelli, Luigi and Sakaridis, Christos and Yang, Yung-Hsu and Segu, Mattia and Li, Siyuan and Abbeloos, Wim and Van Gool, Luc},
  journal={arXiv preprint arXiv:2502.20110},
  year={2025}
}

@inproceedings{sajjan2020cleargrasp,
  title     = {{C}lear{G}rasp: 3{D} Shape Estimation of Transparent Objects for Manipulation},
  author    = {Sajjan, Shreeyak and Moore, Matthew and Pan, Mike and Nagaraja, Ganesh and Lee, Johnny and Zeng, Andy and Song, Shuran},
  booktitle = {ICRA},
  pages     = {3634--3642},
  year      = {2020}
}

@inproceedings{zhu2021rgbd,
  title     = {{RGB-D} Local Implicit Function for Depth Completion of Transparent Objects},
  author    = {Zhu, Luyang and Mousavian, Arsalan and Xiang, Yu and Mazhar, Hammad and van Eenbergen, Jozef and Debnath, Shoubhik and Fox, Dieter},
  booktitle = {CVPR},
  pages     = {12725--12734},
  year      = {2021}
}

@article{fang2022transcg,
  author={Fang, Hongjie and Fang, Hao-Shu and Xu, Sheng and Lu, Cewu},
  journal={IEEE Robotics and Automation Letters}, 
  title={TransCG: A Large-Scale Real-World Dataset for Transparent Object Depth Completion and a Grasping Baseline}, 
  year={2022},
  volume={7},
  number={3},
  pages={7383-7390}}

@inproceedings{chen2023tode,
  title     = {{T}{O}{D}{E}-{T}rans: Transparent Object Depth Estimation with Transformer},
  author    = {Chen, Kang and Wang, Shaochen and Xia, Beihao and Li, Dongxu and Kan, Zhen and Li, Bin},
  booktitle = {ICRA},
  pages     = {4880--4886},
  year      = {2023}
}

@article{wen2025layereddepth,
  title   = {Seeing and Seeing Through the Glass: Real and Synthetic Data for Multi-Layer Depth Estimation},
  author  = {Wen, Hongyu and Zuo, Yiming and Subramanian, Venkat and Chen, Patrick and Deng, Jia},
  journal = {ICCV},
  year    = {2025}
}
